# Machine learning for the diagnosis of Parkinson's disease: A systematic review

Jie Mei[a*], Christian Desrosiers[b] and Johannes Frasnelli[a,c]

a. Department of Anatomy, Université du Québec à Trois-Rivières (UQTR), 3351, boul. des Forges, C.P. 500, Trois-Rivières, QC G9A 5H7, Canada
b. Department of Software and IT Engineering, École de technologie supérieure, 1100 Notre-Dame St W, Montreal, QC H3C 1K3, Canada
c. Centre de Recherche de l'Hôpital du Sacré-Coeur de Montréal, Centre intégré universitaire de santé et de services sociaux du Nord-de-l'Île-de-Montréal (CIUSSS du Nord-de-l'Île-de-Montréal), 5400 Boul Gouin O, Montréal, QC H4J 1C5, Canada

**Keywords:** Parkinson's Disease; Machine Learning; Diagnosis; Differential Diagnosis

[*] Correspondence to: Jie Mei, Department of Anatomy, Université du Québec à Trois-Rivières, 3351, boul. des Forges, C.P. 500, Trois-Rivières, QC G9A 5H7 Canada; Email: jie.mei@uqtr.ca

## Abstract
Diagnosis of Parkinson's disease (PD) is commonly based on medical observations and assessment of clinical signs, including the characterization of a variety of motor symptoms. However, traditional diagnostic approaches may suffer from subjectivity as they rely on the evaluation of movements that are sometimes subtle to human eyes and therefore difficult to classify, leading to possible misclassification. In the meantime, early non-motor symptoms of PD may be mild and can be caused by many other conditions. Therefore, these symptoms are often overlooked, making diagnosis of PD at an early stage challenging. To address these difficulties and to refine the diagnosis and assessment procedures of PD, machine learning methods have been implemented for the classification of PD and healthy controls or patients with similar clinical presentations (e.g., movement disorders or other Parkinsonian syndromes). To provide a comprehensive overview of data modalities and machine learning methods that have been used in the diagnosis and differential diagnosis of PD, in this study, we conducted a systematic literature review of studies published until February 14, 2020, using the PubMed and IEEE *Xplore* databases. A total of 209 studies were included, extracted for relevant information and presented in this systematic review, with an investigation of their aims, sources of data, types of data, machine learning methods and associated outcomes. These studies demonstrate a high potential for adaptation of machine learning methods and novel biomarkers in clinical decision making, leading to increasingly systematic, informed diagnosis of PD.

## 1. Introduction
Parkinson's disease (PD) is one of the most common neurodegenerative diseases with a prevalence rate of 1% in the population above 60 years old, affecting 1 - 2 per 1000 people[1]. The estimated global population affected by PD has more than doubled from 1990 to 2016 (from 2.5 million to 6.1 million), which is a result of increased number of elderly people and age-standardized prevalence rates[2]. PD is a progressive neurological disorder associated with motor and non-motor features[3] which comprises multiple aspects of movements, including planning, initiation and execution[4]. During its development, movement-related symptoms such as tremor, rigidity and difficulties in initiation can be observed, prior to cognitive and behavioral deficits[5]. PD severely affects patients' quality of life (QoL), social functions and family relationships, and places heavy economic burdens at individual and society levels[6-8].

The diagnosis of PD is traditionally based on motor symptoms. Despite the establishment of cardinal signs of PD in clinical assessments, most of the rating scales used in the evaluation of disease severity have not been fully evaluated and validated[3]. Although non-motor symptoms (e.g., cognitive and behavioral abnormalities, sleep disorders, sensory abnormalities such as olfactory dysfunction) are present in many patients prior to the onset of PD[3, 9], they lack specificity, are complicated to assess and/or yield variability from patient to patient[10]. Therefore, non-motor symptoms do not yet allow for diagnosis of PD independently[11], although some have been used as supportive diagnostic criteria[12].

In recent years, machine learning has become a key player in the healthcare sector. As its name implies, machine learning allows for a computer program to learn and extract meaningful representation from data with minimal human supervision. For the diagnosis of PD, machine learning models have been applied to a multitude of data modalities, including movement data (e.g., handwriting[13, 14] or gait[15-17]), neuroimaging[18-20], voice[21, 22], cerebrospinal fluid[23, 24] (CSF), cardiac scintigraphy[25], serum[26] and optical coherence tomography (OCT)[27]. Machine learning also allows for combining different modalities, such as magnetic resonance imaging (MRI) and single-photon emission computed tomography (SPECT) data[28, 29], in the diagnosis of PD. By using machine learning approaches, we may therefore identify relevant features that are not traditionally used in the clinical diagnosis of PD, and rely on these alternative measures to detect PD in preclinical stages or atypical forms.

In recent years, the number of publications on the application of machine learning to the diagnosis of PD has increased. Although previous studies have systematically and thoroughly reviewed the use of machine learning in the diagnosis and assessment of PD, they were limited to the analysis of motor symptoms, kinematics and wearable sensor data[30-32], or studies published between 2015 and 2016[33]. In this study, we aim to (a) comprehensively summarize all published studies that applied machine learning models to the diagnosis of PD for an exhaustive overview of data sources, data types, machine learning models and associated outcomes, (b) assess and compare the feasibility and efficiency of different machine learning methods in the diagnosis of PD, and (c) provide machine learning practitioners interested in the diagnosis of PD with an overview of previously used models and data modalities and the associated outcomes. As a result, the application of machine learning to clinical and non-clinical data of different modalities has often led to high diagnostic accuracies in human participants, therefore may encourage the adaptation of machine learning algorithms and novel biomarkers in clinical settings to assist more accurate and informed decision making.

# 2. Methods

## 2.1 Search strategy

A systematic literature search was conducted on the PubMed (https://pubmed.ncbi.nlm.nih.gov) and IEEE *Xplore* (https://ieeexplore.ieee.org/search/advanced/command) databases on February 14, 2020 for all returned results. Boolean search strings used are shown in Table 1. All retrieved studies were systematically identified, screened and extracted for relevant information following the Preferred Reporting Items for Systematic Reviews and Meta-Analyses (PRISMA) guidelines[34].

| Database | Boolean search string |
|---|---|
| PubMed | ("Parkinson Disease"[Mesh] OR Parkinson*) **AND**<br>("Machine Learning"[Mesh] OR machine learn* OR machine-learn* OR deep learn* OR deep-learn*) **AND**<br>(human OR patient) **AND**<br>("Diagnosis"[Mesh] OR diagnos* OR detect* OR classif* OR identif*)<br>**NOT** review[Publication Type] |
| IEEE *Xplore* | (Parkinson*) **AND**<br>(machine learn* OR machine-learn* OR deep learn* OR deep-learn*) AND (human OR patient) **AND**<br>(diagnosis OR diagnose OR diagnosing OR detection OR detect OR detecting OR classification OR classify OR classifying OR identification OR identify OR identifying) |

**Table 1.** Boolean search strings used for the retrieval of relevant publications on PubMed and IEEE *Xplore* databases.

## 2.2 Inclusion and exclusion criteria

Studies that satisfy one or more of the following criteria and used machine learning methods were included: (1) classification of PD from healthy controls (HC), (2) classification of PD from Parkinsonism (e.g., progressive supranuclear palsy (PSP) and multiple system atrophy (MSA)), (3) classification of PD from other movement disorders (e.g., essential tremor (ET)).

Studies falling into one or more of the following categories were excluded: (1) studies related to diseases other than PD (e.g., differential diagnosis of PSP, MSA and other atypical Parkinsonian disorders), (2) studies not related to the diagnosis of PD (e.g., subtyping or severity assessment, analysis of behavior, disease progression, treatment outcome prediction, identification and localization of brain structures or parameter optimization during surgery), (3) studies related to the diagnosis of PD, but performed analysis and assessed model performance using individual events (e.g., activity-, gait-, genome-level classification), (4) classification of PD from non-Parkinsonism (e.g., Alzheimer's disease), (5) study did not use metrics that measure classification performance, (6) study used organisms other than human (e.g., Caenorhabditis elegans, mice or rats), (7) study did not provide sufficient or accurate descriptions of machine learning methods, datasets or subjects used (e.g., does not provide sample size, or incorrectly described the dataset(s) used), (8) not original journal article or conference proceedings papers (e.g., review and viewpoint paper), and (9) in languages other than English.

## 2.3 Data extraction

The following information is included in the data extraction table: (1) objectives, (2) type of diagnosis (diagnosis, differential diagnosis, sub-typing), (3) data source, (4) data type, (5) number of subjects, (6) machine learning method(s), (7) associated outcomes, (8) year and (9) reference.

For studies published online first and archived in another year, "year of publication" was defined as the year during which the study was published online. If this information was unavailable, the year in which the article was copyrighted was regarded as the year of publication. For studies that introduced novel models and used existing models merely for comparison, information related to the novel models was extracted. Classification of PD and scans without evidence for dopaminergic deficit (SWEDD) was treated as subtyping[35].

## 2.4 Model evaluation

In the present study, accuracy was used to compare performance of machine learning models. For each data type, we summarized the type of machine learning models that led to the per-study highest accuracy. However, in some studies, only one machine learning model was tested. Therefore, we define "model associated with the per-study highest accuracy" as (a) the only model implemented and used in a study or (b) the model that achieved the highest accuracy or that was highlighted in studies that used multiple models. Results are expressed as mean (SD).

For studies reported both training and testing/validation accuracy, testing or validation accuracy was considered. For studies that reported both validation and test accuracy, test accuracy was considered. For studies with more than one dataset or classification problem (e.g., HC vs PD and HC vs idiopathic hyposmia vs PD), accuracy was averaged across datasets or classification problems. For studies that reported classification accuracy for each group of subjects individually, accuracy was averaged across groups. For studies reported a range of accuracies or accuracies given by different cross validation methods or feature combinations, the highest accuracies were considered. In studies that compared HC with diseases other than PD or PD with diseases other than Parkinsonism, diagnosis of diseases other than PD or Parkinsonism (e.g., amyotrophic lateral sclerosis) was not considered. Accuracy of severity assessment was not considered.

# 3. Results

## 3.1 Systematic review

Based on the search criteria, we retrieved 427 (PubMed) and 215 (IEEEXplore) search results, leading to a total of 642 publications. After removing duplicates, we screened 593 publications for titles and abstracts, following which we excluded 313 and examined 280 full text articles. Overall, we included 209 research articles for data extraction (Figure 1; see Supplementary Materials for a full list of included studies). All articles were published from the year 2009 onwards, and an increase in the number of papers published per year was observed (Figure 2).

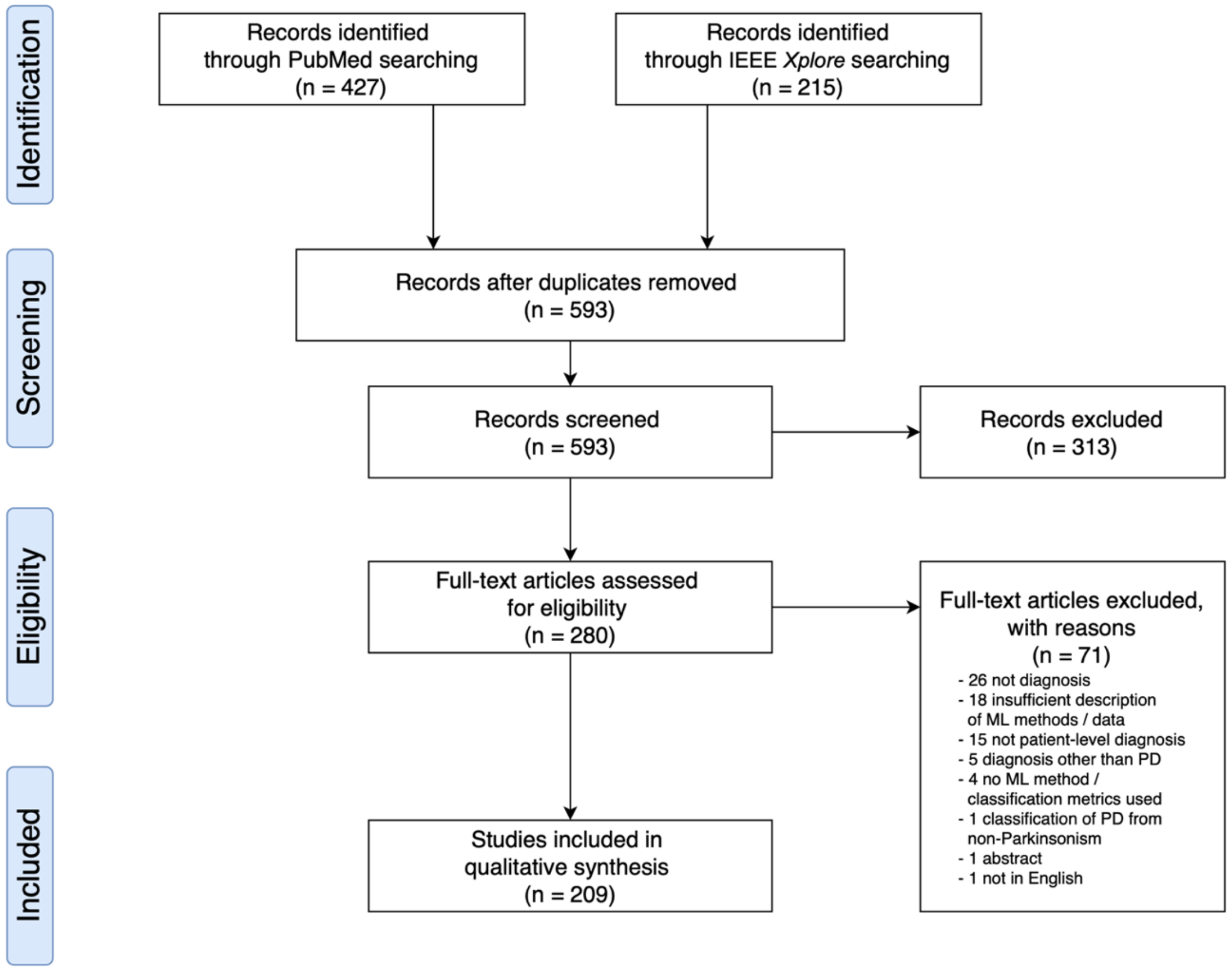

**Figure 1.** PRISMA Flow Diagram of Literature Search and Selection Process showing the number of studies identified, screened, extracted and included in the systematic review.

## 3.2 Study objectives

In included studies, although “diagnosis of PD” was used as the search criteria, machine learning had been applied for diagnosis (PD vs HC), differential diagnosis (idiopathic PD vs atypical Parkinsonism) and sub-typing (differentiation of sub-types of PD) purposes. Most studies focused on diagnosis (n = 168, 80.4%) or differential diagnosis (n = 20, 9.6%). Fourteen studies performed both diagnosis and differential diagnosis (6.7%), 5 studies (2.4%) diagnosed and subtyped PD, 2 studies (1.0%) included diagnosis, differential diagnosis and subtyping.

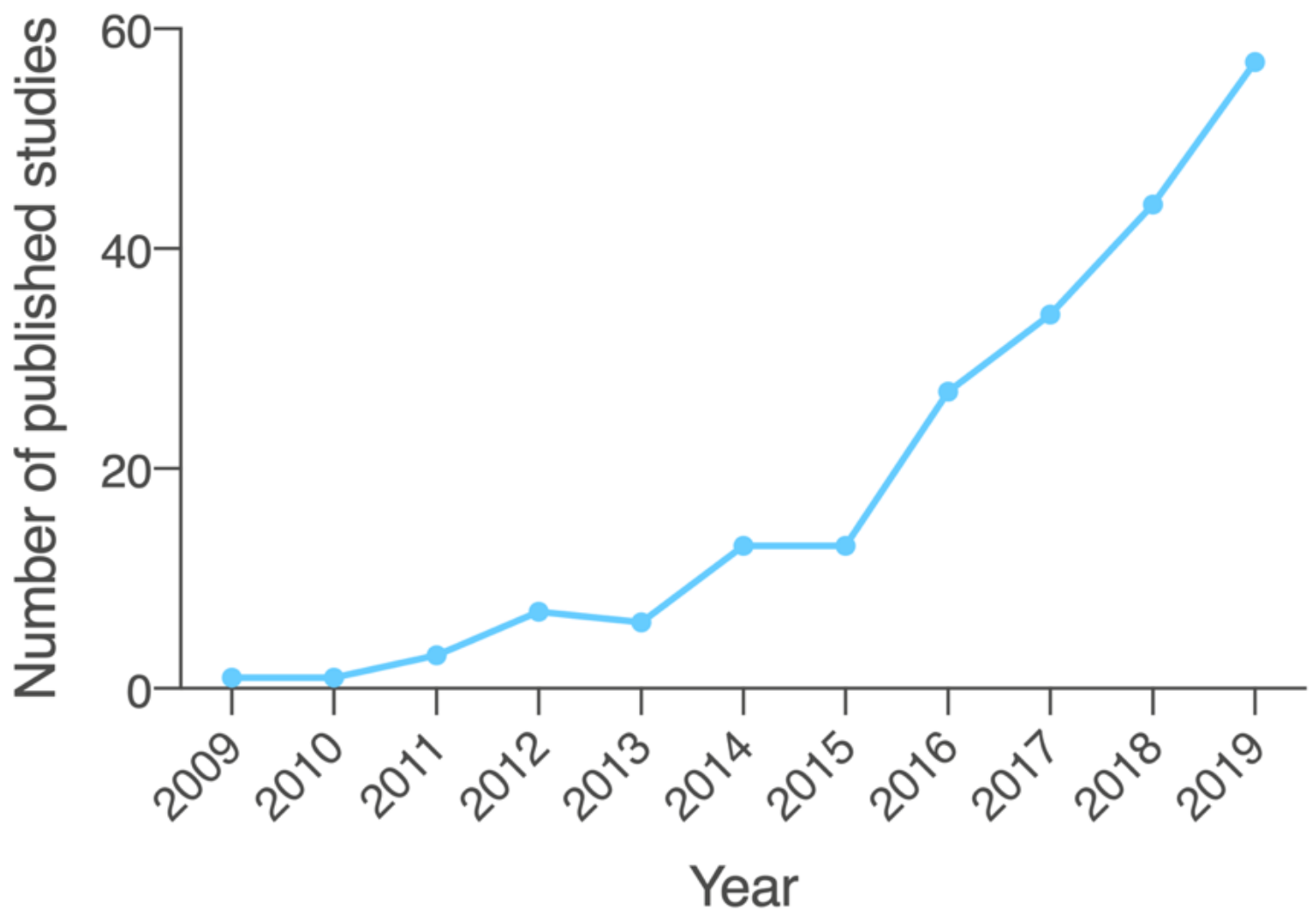

**Figure 2.** Number of included studies published each year since 2009 on machine learning applied to PD diagnosis. Studies published in the year 2020 were excluded.

### 3.3 Data source and sample size

In 93 out of 209 studies (43.1%), original data were collected from human participants. In 108 studies (51.7%), data used were from public repositories and databases, including University of California at Irvine (UCI) Machine Learning Repository[36] (n = 44), Parkinson's Progression Markers Initiative[37] (PPMI; n = 33), PhysioNet[38] (n = 15), HandPD dataset[39] (n = 6), mPower database[40] (n = 4), and 6 other databases[41-46] (Table 2).

In 3 studies, data from public repositories were combined with data from local databases or participants[19, 47, 48]. In the remaining studies, data were sourced[16] from another study[49], collected at another institution[20], obtained from the authors’ institutional database[27], collected postmortem[23] or commercially sourced[26].

The 209 studies had an average sample size of 184.6 (289.3), with a smallest sample size of 10[50] and a largest sample size of 2,289[51] (Figure 3, a, b, c). For studies that recruited human participants (n = 93), data from an average of 118.0 (142.9) participants were collected (range: 10 to 920; Figure 3, d). For other studies (n = 116), an average sample size of 238.1 (358.5) was reported (range: 30 to 2,289; Figure 3, d).

| Data source/Database | Number of studies | Percentage |
|---|---|---|
| independent recruitment of human participants | 93 | 43.06% |
| UCI Machine Learning Repository | 44 | 20.37% |
| PPMI database | 33 | 15.28% |
| PhysioNet | 15 | 6.94% |
| HandPD dataset | 6 | 2.78% |
| mPower database | 4 | 1.85% |
| other databases<br>(1 PACS, 1 PaHaW, 1 PC-GITA database, 1 PDMultiMC database, 1 Neurovoz corpus, 1 The NTUA Parkinson Dataset) | 6 | 2.78% |
| collected postmortem | 1 | 0.46% |
| commercially sourced | 1 | 0.46% |
| acquired at another institution | 1 | 0.46% |
| from another study | 1 | 0.46% |
| from the author's institutional database | 1 | 0.46% |
| others<br>(1 PPMI + Sheffield Teaching Hospitals NHS Foundation Trust;<br>1 PPMI + Seoul National University Hospital cohort;<br>1 UCI + collected from participants) | 3 | 1.39% |

**Table 2.** Source of data of the included studies. PACS: Picture Archiving and Communication System; PaHaW: Parkinson's Disease Handwriting Database.

### 3.4 Machine learning methods applied to the diagnosis of PD

We divided 448 machine learning models from the 209 studies into 8 categories: (1) support vector machine (SVM) and variants (n = 132 from 130 studies), (2) neural networks (n = 76 from 62 studies), (3) ensemble learning (n = 82 from 57 studies), (4) nearest neighbor and variants (n = 33 from 33 studies), (5) regression (n = 31 from 31 studies), (6) decision tree (n = 28 from 27 studies), (7) naïve Bayes (n = 26, from 26 studies) and (8) discriminant analysis (n = 12 from 12 studies). A small percentage of models used did not fall into any of the categories (n = 28, used in 24 studies).

On average, 2.14 machine learning models per study were applied to the diagnosis of PD. One study may have used more than one category of models. For a full description of data types used to train each type of machine learning models and the associated outcomes, see Supplementary Materials and Supplementary Figure 1.

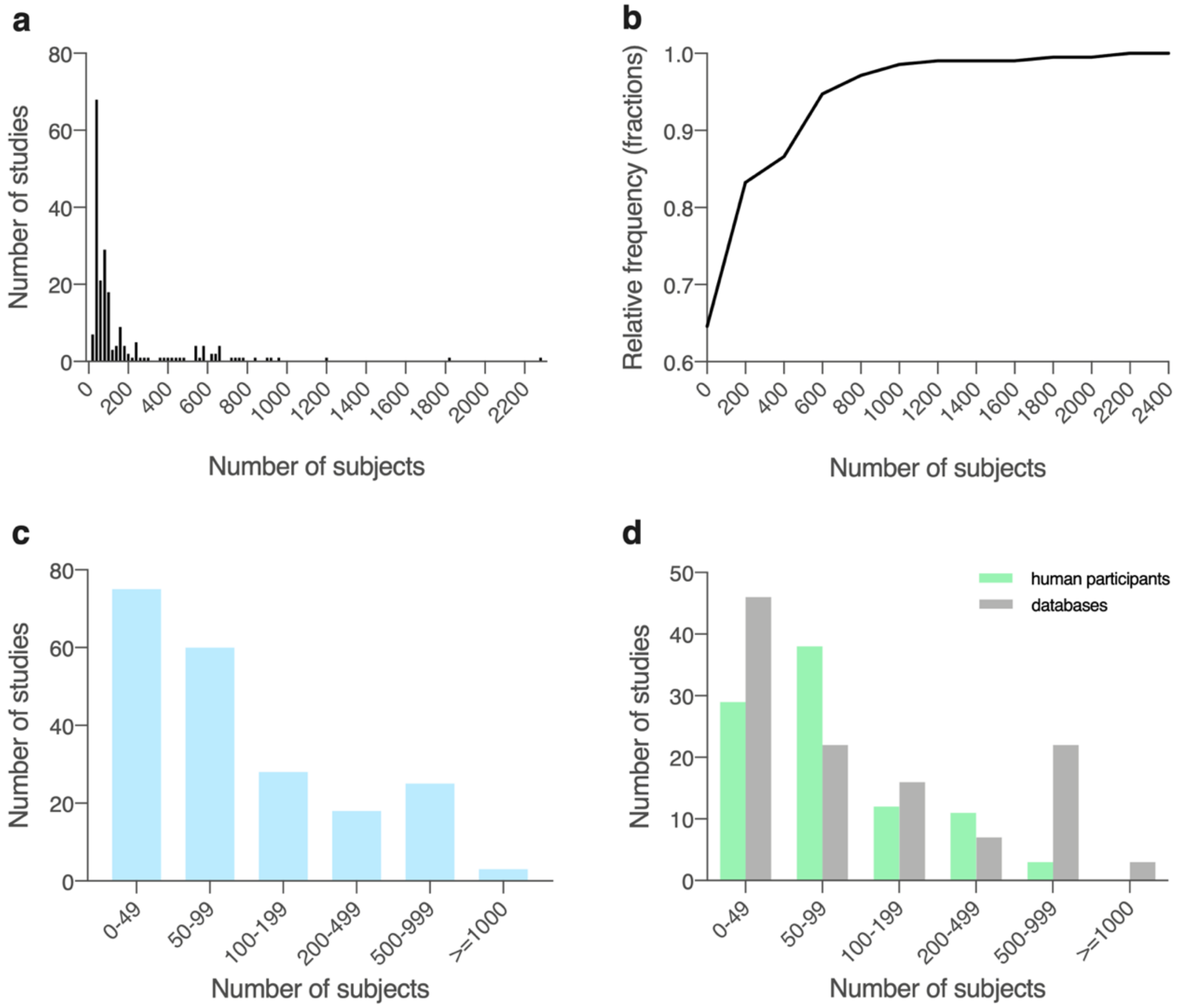

**Figure 3.** Sample size of the included studies. (a) Histogram depicting the frequency of the sample sizes studied. (b) Cumulative relative frequency graph. (c) Histogram depicting the frequency of a sample size of 0-50, 50-100, 100-200, 200-500, 500-100 and over 1000. (d) Same as c but plotted for studies using recruited human participants and studies using published databases individually.

### 3.5 Performance metrics

Various metrics have been used to assess the performance of machine learning models (Table 3; Supplementary Table 1). The most common metric was accuracy (n = 174, 83.3%), which was used individually (n = 55) or in combination with other metrics (n = 119) in model evaluation. Among the 174 studies that used accuracy, some have combined accuracy with sensitivity (i.e., recall) and specificity (n = 42), or with sensitivity, specificity and AUC (n = 16), or with recall (i.e., sensitivity), precision and F1 score (n = 7) for a more systematic understanding of model performance. A total of 35 studies (16.7%) used metrics other than accuracy. In these studies, the most used performance metrics were AUC (n = 19), sensitivity (n = 17) and specificity (n = 14), and the three were often applied together (n = 9) with or without other metrics.

| Performance metric | Definition | Number of studies |
|---|---|---|
| Accuracy | $\frac{TP+TN}{TP+TN+FP+FN}$ | 174 |
| Sensitivity (recall) | $\frac{TP}{TP+FN}$ | 110 |
| Specificity (TNR) | $\frac{TN}{TN+FP}$ | 94 |
| AUC | the two-dimensional area under the Receiver Operating Characteristic (ROC) curve | 60 |
| MCC | $\frac{TP\times TN-FP\times FN}{\sqrt{(TP+FP)(TP+FN)(TN+FP)(TN+FN)}}$ | 9 |
| Precision (PPV) | $\frac{TP}{TP+FP}$ | 31 |
| NPV | $\frac{TN}{TN+FN}$ | 8 |
| F1 score | $2\times\frac{precision\times recall}{precision+recall}$ | 25 |
| others (7 kappa; 4 error rate; 3 EER; 1 MSE; 1 LOR; 1 confusion matrix; 1 cross validation score; 1 YI; 1 FPR; 1 FNR; 1 G-mean; 1 PE; 5 combination of metrics) | N/A | 28 |

**Table 3.** Performance metrics used in the evaluation of machine learning models. TNR: true negative rate; AUC: Area under the ROC Curve; MCC: Matthews correlation coefficient; PPV: positive predictive value; NPV: negative predictive value; EER: equal error rate; MSE: mean squared error; LOR: log odds ratio; YI: Youden's Index; FPR: false positive rate; FNR: false negative rate; PE: probability excess.

## 3.6 Data types and associated outcomes

Out of 209 studies, 122 (58.4%) applied machine learning methods to movement-related data, i.e., voice recordings (n = 55, 26.3%), movement or gait data (n = 51, 24.4%) or handwriting (n = 16, 7.7%). Imaging modalities analyzed including MRI (n = 36, 17.2%), SPECT (n = 14, 6.7%) and positron emission tomography (PET; n = 4, 1.9%). Five studies analyzed CSF samples (2.4%). In 18 studies (8.6%), a combination of different types of data was used.

Ten studies (4.8%) used data that do not belong to any categories mentioned above (Supplementary Table 1), such as single nucleotide polymorphisms[52] (SNPs), electromyography[50] (EMG), OCT[27], cardiac scintigraphy[25], Patient Questionnaire of Movement Disorder Society Unified Parkinson's Disease Rating Scale (MDS-UPDRS)[53], whole-blood gene expression profiles[54], transcranial sonography[55] (TCS), eye movements[56], electroencephalography[57] (EEG) and serum samples[26].

### *3.6.1 Voice recordings (n = 55)*

The 49 studies that used accuracy to evaluate machine learning models achieved an average accuracy of 90.9 (8.6) %, ranging from 70.0%[58, 59] to 100.0%[60-63]. In 3 studies, the highest accuracy was achieved by two types of machine learning models individually, namely regression or SVM[58], neural network or SVM[63], and ensemble learning or SVM[64] (Figure 4, a, b). The per-study highest accuracy was achieved with SVM in 23 studies (39.7%), with neural network in 16 studies (27.6%), with ensemble learning in 7 studies (12.1%), with nearest neighbor in 3 studies (5.2%) and with regression in 2 studies (3.4%). Models that do not belong to any given categories led to the per-study highest accuracy in 7 studies (12.1%; Figure 4, c, d).

Voice recordings from the UCI machine learning repository were used in 42 studies (Table 4). Among the 42 studies, 39 used accuracy to evaluate classification performance and the average accuracy was 92.0 (9.0) %. The lowest accuracy was 70.0% and the highest accuracy was 100.0%. Eight out of 9 studies that collected voice recordings from human participants used accuracy as the performance metric, and the average, lowest and highest accuracies were 87.7 (6.8) %, 77.5% and 98.6%, respectively. The 4 remaining studies used data from the Neurovoz corpus (n = 1), mPower database (n = 1), PC-GITA database (n = 1), or data from both the UCI machine learning repository and human participants (n = 1). Two out of these 4 studies used accuracy to evaluate model performance and reported an accuracy of 81.6% and 91.7%.

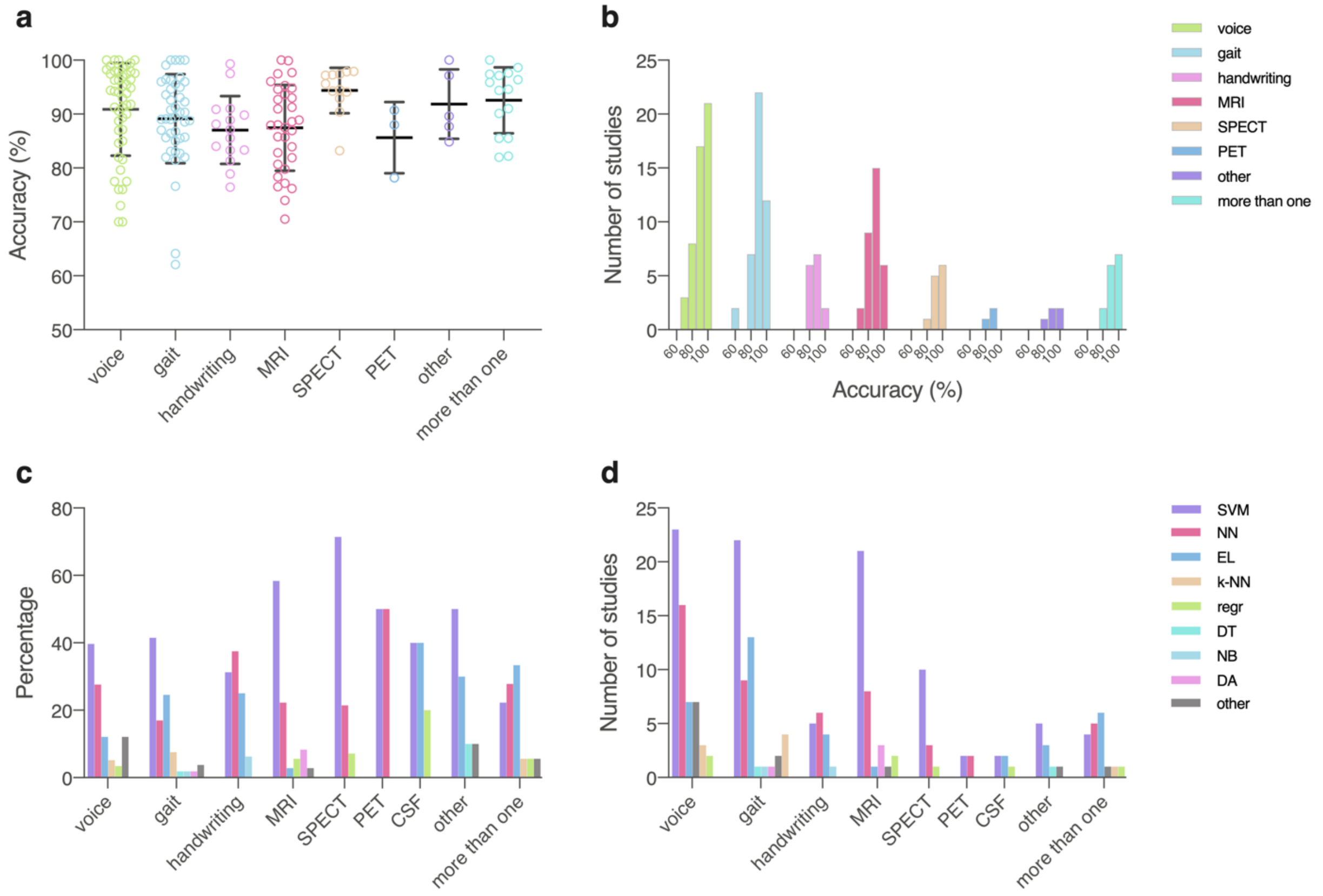

**Figure 4.** Data type, machine learning models applied and associated accuracy. (a) Accuracy achieved in individual studies and average accuracy for each data type. Error bar: standard deviation. (b) Distribution of accuracy per data type. (c) Distribution of machine learning models applied per data type. (d) Same as c but plotted as number of studies. MRI: magnetic resonance

imaging; SPECT: single-photon emission computed tomography; PET: positron emission tomography; CSF: cerebrospinal fluid; SVM: support vector machine; NN: neural network; EL: ensemble learning; k-NN: nearest neighbor; regr: regression; DT: decision tree; NB: naïve Bayes; DA: discriminant analysis; other: data/models that do not belong to any of the given categories.

| **Objectives** | **Type of diagnosis** | **Source of data** | **Number of subjects (n)** | **Machine learning method(s)** | **Outcomes** | **Year** | **Reference** |
|---|---|---|---|---|---|---|---|
| Classification of PD from HC | Diagnosis | UCI machine learning repository | 31; 8 HC + 23 PD | Fuzzy neural system with 10-fold cross validation | testing accuracy = 100% | 2016 | Abiyev and Abizade |
| Classification of PD from HC | Diagnosis | UCI machine learning repository | 31; 8 HC + 23 PD | RPART, C4.5, PART, Bagging CART, random forest, Boosted C5.0, SVM | SVM:<br>accuracy = 97.57%<br>sensitivity = 0.9756<br>specificity = 0.9987<br>NPV = 0.9995 | 2019 | Aich et al |
| Classification of PD from HC | Diagnosis | UCI machine learning repository | 31; 8 HC + 23 PD | DBN of 2 RBMs | testing accuracy = 94% | 2016 | Al-Fatlawi et al |
| Classification of PD from HC | Diagnosis | UCI machine learning repository | 31; 8 HC + 23 PD | EFMM-OneR with 10-fold cross validation or 5-fold cross validation | accuracy = 94.21% | 2019 | Al-Sayaydeha and Mohammad |
| Classification of PD from HC | Diagnosis | UCI machine learning repository | 40; 20 HC + 20 PD | linear regression, LDA, Gaussian naïve Bayes, decision tree, KNN, SVM-linear, SVM-RBF with leave-one-subject-out cross validation | logistic regression or SVM-linear accuracy = 70% | 2019 | Ali et al |
| Classification of PD from HC | Diagnosis | UCI machine learning repository | 40; 20 HC + 20 PD | LDA-NN-GA with leave-one-subject-out cross validation | Training:<br>accuracy = 95%<br>sensitivity = 95%<br><br>Test:<br>accuracy = 100%<br>sensitivity = 100% | 2019 | Ali et al |
| Classification of PD from HC | Diagnosis | UCI machine learning repository | 31; 8 HC + 23 PD | NNge with AdaBoost with 10-fold cross validation | accuracy = 96.30% | 2018 | Alqahtani et al |
| Classification of PD from HC | Diagnosis | UCI machine learning repository | 31; 8 HC + 23 PD | logistic regression, KNN, naïve Bayes, SVM, decision tree, random forest, DNN with 10-fold cross validation (deep NN) | KNN accuracy = 95.513% | 2018 | Anand et al |
| Classification of PD from HC | Diagnosis | UCI machine learning repository | 31; 8 HC + 23 PD | MLP with a train-validation-test ratio of 50:20:30 | training accuracy = 97.86%<br>test accuracy = 92.96%<br>MSE = 0.03552 | 2012 | Bakar et al |
| Classification of PD from HC | Diagnosis | UCI machine learning repository | 31 (8 HC + 23 PD) for dataset 1 and 68 (20 HC + 48 PD) for dataset 2 | FKNN, SVM, KELM with 10-fold cross validation | FKNN accuracy = 97.89% | 2018 | Cai et al |
| Classification of PD from HC | Diagnosis | UCI machine learning repository | 40; 20 HC + 20 PD | SVM, logistic regression, ET, gradient boosting, random forest with train-test split ratio = 80:20 | logistic regression accuracy = 76.03% | 2019 | Celik and Omurca |
| Classification of PD from HC | Diagnosis | UCI machine learning repository | 40; 20 HC + 20 PD | MLP, GRNN with a training-test ratio of 50:50 | GRNN:<br>error rate = 0.0995 (spread parameter = 195.1189)<br>error rate = 0.0958 (spread parameter = 1.2)<br>error rate = 0.0928 (spread parameter = 364.8) | 2016 | Cimen and Bolat |
| Classification of PD from HC | Diagnosis | UCI machine learning repository | 31; 8 HC + 23 PD | ECFA-SVM with 10-fold cross validation | accuracy = 97.95%<br>sensitivity = 97.90%<br>precision = 97.90%<br>F-measure = 97.90%<br>specificity = 96.50%<br>AUC = 97.20% | 2017 | Dash et al |
| Classification of PD from HC | Diagnosis | UCI machine learning repository | 40; 20 HC + 20 PD | fuzzy classifier with 10-fold cross validation, leave-one-out cross validation or a train-test ratio of 70:30 | accuracy = 100% | 2019 | Dastjerd et al |
| Classification of PD from HC | Diagnosis | UCI machine learning repository | 31; 8 HC + 23 PD | averaged perceptron, BPM, boosted decision tree, decision forests, decision jungle, locally deep SVM, logistic regression, NN, SVM with 10-fold cross-validation | boosted decision trees:<br>accuracy = 0.912105<br>precision = 0.935714<br>F-score = 0.942368<br>AUC = 0.966293 | 2017 | Dinesh and He |
| Classification of PD from HC | Diagnosis | UCI machine learning repository | 50; 8 HC + 42 PD | KNN, SVM, ELM with a train-validation ratio of 70:30 | SVM:<br>accuracy = 96.43%<br>MCC = 0.77 | 2017 | Erdogdu Sakar et al |
| Classification of PD from HC | Diagnosis | UCI machine learning repository | 252; 64 HC + 188 PD | CNN with leave-one-person-out cross validation | accuracy = 0.869<br>F-Measure = 0.917<br>MCC = 0.632 | 2019 | Gunduz |
| Classification of PD from HC | Diagnosis | UCI machine learning repository | 31; 8 HC + 23 PD | SVM, logistic regression, KNN, DNN with a train-test ratio of 70:30 | DNN:<br>accuracy = 98%<br>specificity = 95%<br>sensitivity = 99% | 2018 | Haq et al |
| Classification of PD from HC | Diagnosis | UCI machine learning repository | 31; 8 HC + 23 PD | SVM-RBF, SVM-linear with 10-fold cross validation | accuracy = 99%<br>specificity = 99%<br>sensitivity = 100% | 2019 | Haq et al |
| Classification of PD from HC | Diagnosis | UCI machine learning repository | 31; 8 HC + 23 PD | LS-SVM, PNN, GRNN with conventional (train-test ratio of 50:50) and 10-fold cross validation | LS-SVM or PNN or GRNN:<br>accuracy = 100%<br>precision = 100%<br>sensitivity = 100%<br>specificity = 100%<br>AUC = 100 | 2014 | Hariharan et al |
| Classification of PD from HC | Diagnosis | UCI machine learning repository | 31; 8 HC + 23 PD | random tree, SVM-linear, FBANN with 10-fold cross validation | FBANN:<br>accuracy = 97.37%<br>sensitivity = 98.60%<br>specificity = 93.62%<br>FPR = 6.38%<br>precision = 0.979<br>MSE = 0.027 | 2014 | Islam et al |
| Classification of PD from HC | Diagnosis | UCI machine learning repository | 31; 8 HC + 23 PD | SVM-linear with 5-fold cross validation | error rate ~0.13 | 2012 | Ji and Li |
| Classification of PD from HC | Diagnosis | UCI machine learning repository | 40; 20 HC + 20 PD | decision tree, random forest, SVM, GBM, XGBoost | SVM-linear:<br>FNR = 10%<br>accuracy = 0.725 | 2018 | Junior et al |
| Classification of PD from HC | Diagnosis | UCI machine learning repository | 31; 8 HC + 23 PD | CART, SVM, ANN | SVM accuracy = 93.84% | 2020 | Karapinar Senturk |
| Classification of PD from HC | Diagnosis | UCI machine learning repository | dataset 1: 31; 8 HC + 23 PD dataset 2: 40; 20 HC + 20 PD | EWNN with a train-test ratio of 90:10 and cross validation | dataset 1: accuracy = 92.9% ensemble classification accuracy = 100.0% sensitivity = 100.0% MCC = 100.0% dataset 2: accuracy = 66.3% ensemble classification accuracy = 90.0% sensitivity = 93.0% specificity = 97.0% MCC = 87.0% | 2018 | Khan et al |
| Classification of PD from HC | Diagnosis | UCI machine learning repository | 40; 20 HC + 20 PD | stacked generalization with CMTNN with 10-fold cross validation | accuracy = ~70% | 2015 | Kraipeerapun and Amornsamankul |
| Classification of PD from HC | Diagnosis | UCI machine learning repository | 40; 20 HC + 20 PD | HMM, SVM | HMM: accuracy = 95.16% sensitivity = 93.55% specificity = 91.67% | 2019 | Kuresan et al |
| Classification of PD from HC | Diagnosis | UCI machine learning repository | 31; 8 HC + 23 PD | IGWO-KELM with 10-fold cross validation | iteration number = 100 accuracy = 97.45% sensitivity = 99.38% specificity = 93.48% precision = 97.33% G-mean = 96.38% F-measure = 98.34% | 2017 | Li et al |
| Classification of PD from HC | Diagnosis | UCI machine learning repository | 31; 8 HC + 23 PD | SCFW-KELM with 10-fold cross validation | accuracy = 99.49% sensitivity = 100% specificity = 99.39% AUC = 99.69% f-measure = 0.9966 kappa = 0.9863 | 2014 | Ma et al |
| Classification of PD from HC | Diagnosis | UCI machine learning repository | 31; 8 HC + 23 PD | SVM-RBF with 10-fold cross validation | accuracy = 96.29% sensitivity = 95.00% specificity = 97.50% | 2016 | Ma et al |
| Classification of PD from HC | Diagnosis | UCI machine learning repository | 31; 8 HC + 23 PD | logistic regression, NN, SVM, SMO, Pegasos, AdaBoost, ensemble selection, FURIA, rotation forest Bayesian network with 10-fold cross-validation | average accuracy across all models = 97.06% SMO, Pegasos or AdaBoost accuracy = 98.24% | 2013 | Mandal and Sairam |
| Classification of PD from HC | Diagnosis | UCI machine learning repository | 31; 8 HC + 23 PD | logistic regression, KNN, SVM, naïve Bayes, decision tree, random forest, ANN | ANN: accuracy = 94.87% specificity = 96.55% sensitivity = 90% | 2018 | Marar et al |
| Classification of PD from HC | Diagnosis | UCI machine learning repository | dataset 1: 31; 8 HC + 23 PD dataset 2: 20 HC + 20 PD | KNN | dataset 1 accuracy = 90% dataset 2 accuracy = 65% | 2017 | Moharkan et al |
| Classification of PD from HC | Diagnosis | UCI machine learning repository | 31; 8 HC + 23 PD | rotation forest ensemble with 10-fold cross validation | accuracy = 87.1% kappa error = 0.63 AUC = 0.860 | 2011 | Ozcift and Gulten |
| Classification of PD from HC | Diagnosis | UCI machine learning repository | 31; 8 HC + 23 PD | rotation forest ensemble | accuracy = 96.93%<br>Kappa = 0.92<br>AUC = 0.97 | 2012 | Ozcift |
| Classification of PD from HC | Diagnosis | UCI machine learning repository | 31; 8 HC + 23 PD | SVM-RBF with 10-fold cross validation or a train-test ratio of 50:50 | 10-fold cross validation:<br>accuracy = 98.95%<br>sensitivity = 96.12%<br>specificity = 100%<br>f-measure = 0.9795<br>Kappa = 0.9735<br>AUC = 0.9808 | 2016 | Peker |
| Classification of PD from HC | Diagnosis | UCI machine learning repository | 31; 8 HC + 23 PD | ELM with 10-fold cross validation | accuracy = 88.72%<br>recall = 94.33%<br>precision = 90.48%<br>F-score = 92.36% | 2016 | Shahsavari et al |
| Classification of PD from HC | Diagnosis | UCI machine learning repository | 31; 8 HC + 23 PD | ensemble learning with 10-fold cross validation | accuracy = 90.6%<br>sensitivity = 95.8%<br>specificity = 75% | 2019 | Sheibani et al |
| Classification of PD from HC | Diagnosis | UCI machine learning repository | 31; 8 HC + 23 PD | GLRA, SVM, bagging ensemble with 5-fold cross validation | bagging:<br>sensitivity = 0.9796<br>specificity = 0.6875<br>MCC = 0.6977<br>AUC = 0.9558<br><br>SVM:<br>sensitivity = 0.9252<br>specificity = 0.8542<br>MCC = 0.7592<br>AUC = 0.9349 | 2017 | Wu et al |
| Classification of PD from HC | Diagnosis | UCI machine learning repository | 31; 8 HC + 23 PD | decision tree classifier, logistic regression, SVM with 10-fold cross validation | SVM:<br>accuracy = 0.76<br>sensitivity = 0.9745<br>specificity = 0.13 | 2011 | Yadav et al |
| Classification of PD from HC | Diagnosis | UCI machine learning repository | 80; 40 HC + 40 PD | KNN, SVM with 10-fold cross validation | SVM:<br>accuracy = 91.25%<br>precision = 0.9125<br>recall = 0.9125<br>F-Measure = 0.9125 | 2019 | Yaman et al |
| Classification of PD from HC | Diagnosis | UCI machine learning repository | 31; 8 HC + 23 PD | MAP, SVM-RBF, FLDA with 5-fold cross validation | MAP:<br>accuracy = 91.8%<br>sensitivity = 0.986<br>specificity = 0.708<br>AUC = 0.94 | 2014 | Yang et al |
| Classification of PD from other disorders | Differential diagnosis | collected from participants | 50; 30 PD + 9 MSA + 5 FND + 1 somatization + 1 dystonia + 2 CD + 1 ET + 1 GPD | SVM, KNN, DA, naïve Bayes, classification tree with LOSO | SVM-linear:<br>accuracy = 90%<br>sensitivity = 90%<br>specificity = 90%<br>MCC = 0.794067<br>PE = 0.788177 | 2016 | Benba et al |
| Classification of PD from other disorders | Differential diagnosis | collected from participants | 40; 20 PD + 9 MSA + 5 FND + 1 somatization + 1 dystonia + 2 CD + 1ET + 1 GPD | SVM (RBF, linear, polynomial and MLP kernels) with LOSO | SVM-linear accuracy = 85% | 2016 | Benba et al |
| Classification of PD from HC and assess the severity of PD | Diagnosis | collected from participants | 52; 9 HC + 43 PD | SVM-RBF with cross validation | accuracy = 81.8% | 2014 | Frid et al |
| Classification of PD from HC | Diagnosis | collected from participants | 54; 27 HC + 27 PD | SVM with stratified 10-fold cross validation or leave-one-out cross validation | accuracy = 94.4%<br>specificity = 100%<br>sensitivity = 88.9% | 2018 | Montaña et al |
| Classification of PD from HC | Diagnosis | collected from participants | 40; 20 HC + 20 PD | KNN, SVM-linear, SVM-RBF with leave-one-subject-out or summarized leave-one-out | SVM-linear:<br>accuracy = 77.50%<br>MCC = 0.5507<br>sensitivity = 80.00%<br>specificity = 75.00% | 2013 | Sakar et al |
| Classification of PD from HC | Diagnosis | collected from participants | 78; 27 HC + 51 PD | KNN, SVM-linear, SVM-RBF, ANN, DNN with leave-one-out cross validation | SVM-RBF:<br>accuracy = 84.62%<br>precision = 88.04%<br>recall = 78.65% | 2017 | Sztahó et al |
| Classification of PD from HC and assess the severity of PD | Diagnosis | collected from participants | 88; 33 HC + 55 PD | KNN, SVM-linear, SVM-RBF, ANN, DNN with leave-one-subject-out cross validation | SVM-RBF:<br>accuracy = 89.3%<br>sensitivity = 90.2%<br>specificity = 87.9% | 2019 | Sztahó et al |
| Classification of PD from HC | Diagnosis | collected from participants | 43; 10 HC + 33 PD | random forests, SVM with 10-fold cross validation and a train-test ratio of 90:10 | SVM accuracy = 98.6% | 2012 | Tsanas et al |
| Classification of PD from HC | Diagnosis | collected from participants | 99; 35 HC + 64 PD | random forest with internal out-of-bag (OOB) validation | EER = 19.27% | 2017 | Vaiciukynas et al |
| Classification of PD from HC | Diagnosis | UCI machine learning repository and participants | 40 and 28; 20 HC + 20 PD and 28 PD, respectively | ELM | training data:<br>accuracy = 90.76%<br>MCC = 0.815<br><br>test data:<br>accuracy = 81.55% | 2016 | Agrawal et al |
| Classification of PD from HC | Diagnosis | the Neurovoz corpus | 108; 56 HC + 52 PD | Siamese LSTM-based NN with 10-fold cross-validation | EER = 1.9% | 2019 | Bhati et al |
| Classification of PD from HC | Diagnosis | mPower database | 2,289; 2,023 HC + 246 PD | L2-regularized logistic regression, random forest, gradient boosted decision trees with 5-fold cross validation | gradient boosted decision trees:<br>recall = 0.797<br>precision = 0.901<br>F1-score = 0.836 | 2019 | Tracy et al |
| Classification of PD from HC | Diagnosis | PC-GITA database | 100; 50 HC + 50 PD | ResNet with train-validation ratio of 90:10 | precision = 0.92<br>recall = 0.92<br>F1-score = 0.92<br>accuracy = 91.7% | 2019 | Wodzinski et al |

**Table 4.** Studies that applied machine learning models to voice recordings to diagnose PD (n = 55).

### *3.6.2 Movement/gait data (n = 51)*

The 43 out of 51 studies used accuracy to assess model performance achieved an average accuracy of 89.1 (8.3) %, ranging from 62.1%[65] to 100.0%[17, 66-68] (Figure 4, a, b). One study reported three machine learning methods (SVM, nearest neighbor and decision tree) achieving the highest accuracy individually[69] . Out of the 51 studies, the per-study highest accuracy was achieved with SVM in 22 studies (41.5%), with ensemble learning in 13 studies (24.5%), with neural network in 9 studies (17.0%), with nearest neighbor in 4 studies (7.5%), with discriminant analysis in 1 study (1.9%), with naïve Bayes in 1 study (1.9%) and with decision tree in 1 study (1.9%). Models that do not belong to any given categories were associated with the highest per-study accuracy in two studies (3.8%; Figure 4, c, d).

Among the 33 studies that collected movement data from recruited participants, 25 used accuracy in model evaluation, leading to an average accuracy of 87.0 (7.3) % (Table 5). The lowest and highest accuracies were 64.1% and 100.0%, respectively. Fifteen studies used data from the PhysioNet database (Table 5) and had an average accuracy of 94.4 (4.6) %, a lowest accuracy of 86.4% and a highest accuracy of 100%. Three studies used data from the mPower database (n = 2) or data sourced from another study (n = 1), and the average accuracy of these studies was 80.6 (16.2) %.

| **Objectives** | **Type of diagnosis** | **Source of data** | **Number of subjects (n)** | **Machine learning method(s)** | **Outcomes** | **Year** | **Reference** |
|---|---|---|---|---|---|---|---|
| Classification of PD from HC | Diagnosis | collected from participants | 103; 71 HC + 32 PD | ensemble method of 8 models (SVM, MLP, LRM, RFC, NSVC, DTC, KNN, QDA) | sensitivity = 96%<br>specificity = 97%<br>AUC = 0.98 | 2017 | Adams |
| Classification of PD, HC and other neurological stance disorders | Diagnosis and differential diagnosis | collected from participants | 293; 57 HC + 27 PD + 49 AVS + 12 PNP + 48 CA + 16 DN + 25 OT + 59 PPV | ensemble method of 7 models (logistic regression, KNN, shallow and deep ANNs, SVM, random forest, extra-randomized trees) with 90% training and 10% testing data in stratified k-fold cross-validation | 8-class classification accuracy = 82.7% | 2019 | Ahmadi et al |
| Classification of PD from HC | Diagnosis | collected from participants | 137; 38 HC + 99 PD | SVM with leave-one-out-cross validation | PD vs HC accuracy = 92.3%<br>mild vs severe accuracy = 89.8%<br>mild vs HC accuracy = 85.9% | 2016 | Bernad-Elazari et al |
| Classification of PD from HC | Diagnosis | collected from participants | 30; 14 HC + 16 PD | SVM (linear, quadratic, cubic, Gaussian kernels), ANN, with 5-fold cross-validation | classification with ANN:<br>accuracy = 89.4%<br>sensitivity = 87.0%<br>specificity = 91.8%<br><br>severity assessment with ANN:<br>accuracy = 95.0%<br>sensitivity = 90.0%<br>specificity = 99.0% | 2019 | Boungiorno et al |
| Classification of PD from HC | Diagnosis | collected from participants | 28; 12 HC + 16 PD | NN with a train-validation-test ratio of 70:15:15, SVM with leave-one-out cross-validation, logistic regression with 10-fold cross validation | SVM:<br>accuracy = 85.71%<br>sensitivity = 83.5%<br>specificity = 87.5% | 2017 | Butt et al |
| Classification of PD from HC | Diagnosis | collected from participants | 28; 12 HC + 16 PD | logistic regression, naïve Bayes, SVM with 10-fold cross validation | naïve Bayes:<br>accuracy = 81.45%<br>sensitivity = 76%<br>specificity = 86.5%<br>AUC = 0.811 | 2018 | Butt et al |
| Classification of PD from HC | Diagnosis | collected from participants | 54; 27 HC + 27 PD | naïve Bayes, LDA, KNN, decision tree, SVM-linear, SVM-rbf, majority of votes with 5-fold cross validation | majority of votes (weighted) accuracy = 96% | 2018 | Caramia et al |
| Classification of PD, HC and PD, HC, IH | Diagnosis | collected from participants | 90; 30 PD + 30 HC + 30 IH | SVM, random forest, naïve Bayes with 10-fold cross validation | random forest:<br>HC vs PD:<br>accuracy = 0.950<br>F-measure = 0.947<br><br>HC + IH vs PD:<br>accuracy = 0.917<br>F-measure = 0.912<br><br>HC vs IH vs PD:<br>accuracy = 0.789<br>F-measure = 0.796 | 2019 | Cavallo et al |
| Classification of PD from HC and classification of HC, MCI, PDNOMCI, and PDMCI | Diagnosis, differential diagnosis and subtyping | collected from participants | PD vs HC:<br>75; 50 HC + 25 PD<br><br>subtyping:<br>52; 18 HC + 16 PDNOMCI + 9 PDMCI + 9 MCI | decision tree, naïve Bayes, random forest, SVM, adaptive boosting (with decision tree or random forest) with 10-fold cross validation | adaptive boosting with decision tree:<br>PD vs HC:<br>accuracy = 0.79<br>AUC = 0.82<br><br>subtyping (HOA vs MCI vs PDNOMCI vs PDMCI):<br>accuracy = 0.85<br>AUC = 0.96 | 2015 | Cook et al |
| Classification of PD from HC | Diagnosis | collected from participants | 580; 424 HC + 156 PD | hidden Markov models with nearest neighbor classifier with cross validation and train-test ratio of 66.6:33.3 | accuracy = 85.51% | 2017 | Cuzzolin et al |
| Classification of PD from HC | Diagnosis | collected from participants | 80; 40 HC + 40 PD | random forest, SVM with 10-fold cross validation | SVM-RBF:<br>accuracy = 85%<br>sensitivity = 85%<br>specificity = 82%<br>PPV = 86%<br>NPV = 83% | 2017 | Djurić-Jovičić et al |
| Classification of PD from HC | Diagnosis | collected from participants | 13; 5 HC + 8 PD | SVM-RBF with leave-one-out cross validation | 100% HC and PD classified correctly (confusion matrix) | 2014 | Dror et al |
| Classification of PD from HC | Diagnosis | collected from participants | 75; 38 HC + 37 PD | SVM with leave-one-out cross validation | accuracy = 85.61%<br>sensitivity = 85.95%<br>specificity = 85.26% | 2014 | Drotar et al |
| Classification of PD from ET | Differential diagnosis | collected from participants | 24; 13 PD + 11 ET | SVM-linear, SVM-RBF with leave-one-out cross validation | accuracy = 83% | 2016 | Ghassemi et al |
| Classification of PD from HC | Diagnosis | collected from participants | 41; 22 HC + 19 PD | SVM, decision tree, random forest, linear regression with 10-fold and leave-one-individual out (L1O) cross validation | SVM accuracy = 0.89 | 2018 | Javed et al |
| Classification of PD from HC | Diagnosis | collected from participants | 74; 33 young HC + 14 elderly HC + 27 PD | SVM with 10-fold cross validation | sensitivity = ~90% | 2017 | Klein et al |
| Classification of PD from HC and assess the severity of PD | Diagnosis | collected from participants | 55; 20 HC + 35 PD | SVM with leave-one-out cross validation | PD diagnosis:<br>accuracy = 89%<br>precision = 0.91<br>recall = 0.94<br><br>severity assessment:<br>HYS 1 accuracy = 72%<br>HYS 2 accuracy = 77%<br>HYS 3 accuracy = 75%<br>HYS 4 accuracy = 33% | 2016 | Kocer and Oktay |
| Classification of PD from HC | Diagnosis | collected from participants | 45; 20 HC + 25 PD | naïve Bayes, logistic regression, SVM, AdaBoost, C4.5, BagDT with 10-fold stratified cross-validation apart from BagDT | BagDT:<br>sensitivity = 82%<br>specificity = 90%<br>AUC = 0.94 | 2015 | Kostikis et al |
| Classification of PD from HC | Diagnosis | collected from participants | 40; 26 HC + 14 PD | random forest with leave-one-subject-out cross-validation | accuracy = 94.6%<br>sensitivity = 91.5%<br>specificity = 97.2% | 2017 | Kuhner et al |
| Classification of PD from HC | Diagnosis | collected from participants | 177; 70 HC + 107 PD | ESN with 10-fold cross validation | AUC = 0.852 | 2018 | Lacy et al |
| Classification of PD from HC | Diagnosis | collected from participants | 39; 16 young HC + 12 elderly HC + 11 PD | LDA with leave-one-out cross validation | multiclass classification (young HC vs age-matched HC vs PD):<br>accuracy = 64.1%<br>sensitivity = 47.1%<br>specificity = 77.3% | 2018 | Martinez et al |
| Classification of PD from HC | Diagnosis | collected from participants | 38; 10 HC + 28 PD | SVM-Gaussian with leave-one-out cross validation | training accuracy = 96.9%<br>test accuracy = 76.6% | 2018 | Oliveira et al |
| Classification of PD from HC | Diagnosis | collected from participants | 30; 15 HC + 15 PD | SVM-RBF, PNN with 10-fold cross validation | SVM-RBF:<br>accuracy = 88.80%<br>sensitivity = 88.70%<br>specificity = 88.15%<br>AUC = 88.48 | 2015 | Oung et al |
| Classification of PD from HC | Diagnosis | collected from participants | 45; 14 HC + 31 PD | Deep-MIL-CNN with LOSO or RkF | with LOSO:<br>precision = 0.987<br>sensitivity = 0.9<br>specificity = 0.993<br>F1-score = 0.943<br><br>with RkF:<br>precision = 0.955<br>sensitivity = 0.828<br>specificity = 0.979<br>F1-score = 0.897 | 2019 | Papadopouloos et al |
| Classification of PD, HC and post-stroke | Diagnosis and differential diagnosis | collected from participants | 11; 3 HC + 5 PD + 3 post-stroke | MTFL with 10-fold cross validation | PD vs HC AUC = 0.983 | 2017 | Papavasileiou et al |
| Classification of PD from HC | Diagnosis | collected from participants | 182; 94 HC + 88 PD | LSTM, CNN-1D, CNN-LSTM with 5-fold cross-validation and a training-test ratio of 90:10 | CNN-LSTM:<br>accuracy = 83.1%<br>precision = 83.5%<br>recall = 83.4%<br>F1-score = 81%<br>Kappa = 64% | 2019 | Reyes et al |
| Classification of PD from HC | Diagnosis | collected from participants | 60; 30 HC + 30 PD | naïve Bayes, KNN, SVM with leave-one-out cross validation | SVM:<br>accuracy = 95%<br>precision = 0.951<br>AUC = 0.950 | 2019 | Ricci et al |
| Classification of PD, HC and IH | Diagnosis and differential diagnosis | collected from participants | 90; 30 HC + 30 PD + 30 IH | SVM-polynomial, random forest, naïve Bayes with 10-fold cross validation | HC vs PD, naïve Bayes or random forest:<br>precision = 0.967<br>recall = 0.967<br>specificity = 0.967<br>accuracy = 0.967<br>F-Measure = 0.967<br><br>multiclass classification, random forest:<br>precision = 0.784<br>recall = 0.778<br>specificity = 0.889<br>accuracy = 0.778<br>F-Measure = 0.781 | 2018 | Rovini et al |
| Classification of PD, HC and IH | Diagnosis and differential diagnosis | collected from participants | 45; 15 HC + 15 PD + 15 IH | SVM-polynomial, random forest with 5-fold cross validation | HC vs PD, random forest:<br>precision = 1.000<br>recall = 1.000<br>specificity = 1.000<br>accuracy = 1.000<br>F-Measure = 1.000<br><br>multiclass classification (HC vs IH vs PD), random forest:<br>precision = 0.930<br>recall = 0.911<br>specificity = 0.956<br>accuracy = 0.911<br>F-Measure = 0.920 | 2019 | Rovini et al |
| Classification of PD from ET | Differential diagnosis | collected from participants | 52; 32 PD + 20 ET | SVM-linear with 10-fold cross validation | accuracy = 1<br>sensitivity = 1<br>specificity = 1 | 2016 | Surangsrirat et al |
| Classification of PD from HC | Diagnosis | collected from participants | 12; 10 HC + 2 PD | naive Bayes, LogitBoost, random forest, SVM with 10-fold cross-validation | random forest:<br>accuracy = 92.29%<br>precision = 0.99<br>recall = 0.99 | 2017 | Tahavori et al |
| Classification of PD from HC | Diagnosis | collected from participants | 39; 16 HC + 23 PD | SVM-RBF with 10-fold stratified cross validation | sensitivity = 88.9%<br>specificity = 100%<br>precision = 100%<br>FPR = 0.0% | 2010 | Tien et al |
| Classification of PD from HC | Diagnosis | collected from participants | 60; 30 HC + 30 PD | logistic regression, naïve Bayes, random forest, decision tree with 10-fold cross validation | random forest:<br>accuracy = 82%<br>false negative rate = 23% | 2018 | Urcuqui et al |
| | | | | | false positive rate = 12% | | |
| Classification of PD from HC | Diagnosis | PhysioNet | 47; 18 HC + 29 PD | SVM, KNN, random forest, decision tree | SVM with cubic kernel:<br>accuracy = 93.6%<br>sensitivity = 93.1%<br>specificity = 94.1% | 2017 | Alam et al |
| Classification of PD from HC | Diagnosis | PhysioNet | 34; 17 HC + 17 PD | MLP, SVM, decision tree | MLP:<br>accuracy = 91.18%<br>sensitivity = 1<br>specificity = 0.83<br>error = 0.09<br>AUC = 0.92 | 2018 | Alaskar and Hussain |
| Classification of PD from HC and assess the severity of PD | Diagnosis | PhysioNet | 166; 73 HC + 93 PD | 1D-CNN, 2D-CNN, LSTM, decision tree, logisitic regression, SVM, MLP | 2D-CNN and LSTM accuracy = 96.0% | 2019 | Alharthi and Ozanyan |
| Classification of PD from HC | Diagnosis | PhysioNet | 146; 60 HC + 86 PD | SVM-Gaussian with 3- or 5-fold cross validation | accuracy = 100%, 88.88% and 100% in three test groups | 2019 | Andrei et al |
| Classification of PD from HC | Diagnosis | PhysioNet | 166; 73 HC + 93 PD | ANN, SVM, naïve Bayes with cross validation | ANN accuracy = 86.75% | 2017 | Baby et al |
| Classification of PD from HC | Diagnosis | PhysioNet | 31; 16 HC + 15 PD | SVM-linear, KNN, naïve Bayes, LDA, decision tree with leave-one-out cross validation | SVM, KNN and decision tree accuracy = 96.8% | 2019 | Felix et al |
| Classification of PD from HC | Diagnosis | PhysioNet | 31; 16 HC + 15 PD | SVM-linear with leave-one-out cross validation | accuracy = 100% | 2017 | Joshi et al |
| Classification of PD from HC | Diagnosis | PhysioNet | 165; 72 HC + 93 PD | KNN, CART, decision tree, random forest, naïve Bayes, SVM-polynomial, SVM-linear, K-means, GMM with leave-one-out cross validation | SVM:<br>accuracy = 90.32%<br>precision = 90.55%<br>recall = 90.21%<br>F-measure = 90.38% | 2019 | Khoury et al |
| Classification of ALS, HD, PD from HC | Diagnosis | PhysioNet | 64; 16 HC + 15 PD + 13 ALS + 20 HD | string grammar unsupervised possibilistic fuzzy C-medians with FKNN, with 4-fold cross validation | PD vs HC accuracy = 96.43% | 2018 | Klomsae et al |
| Classification of PD from HC | Diagnosis | PhysioNet | 166; 73 HC + 93 PD | logistic regression, decision trees, random forest, SVM-Linear, SVM-RBF, SVM-Poly, KNN with cross validation | KNN:<br>accuracy = 93.08%<br>precision = 89.58%<br>recall = 84.31%<br>F1-score = 86.86% | 2018 | Mittra and Rustagi |
| Classification of PD from HC | Diagnosis | PhysioNet | 85; 43 HC + 42 PD | LS-SVM with leave-one-out, 2- or 10-fold cross validation | leave-one-out cross validation: AUC = 1 sensitivity = 100% specificity = 100% accuracy = 100%<br><br>10-fold cross validation: AUC = 0.89 sensitivity = 85.00% specificity = 73.21% accuracy = 79.31% | 2018 | Pham |
| Classification of PD from HC | Diagnosis | PhysioNet | 165; 72 HC + 93 PD | LS-SVM with leave-one-out, 2- or 5- or 10-fold cross validation | accuracy = 100% sensitivity = 100% specificity = 100% AUC = 1 | 2018 | Pham and Yan |
| Classification of PD from HC | Diagnosis | PhysioNet | 166; 73 HC + 93 PD | DCALSTM with stratified 5-fold cross validation | sensitivity = 99.10% specificity = 99.01% accuracy = 99.07% | 2019 | Xia et al |
| Classification of HC, PD, ALS and HD | Diagnosis and differential diagnosis | PhysioNet | 64; 16 HC + 15 PD + 13 ALS + 20 HD | SVM-RBF with 10-fold cross validation | PD vs HC: accuracy = 86.43% AUC = 0.92 | 2009 | Yang et al |
| Classification of PD, HD, ALS and ND from HC | Diagnosis | PhysioNet | 64; 16 HC + 15 PD + 13 ALS + 20 HD | adaptive neuro-fuzzy inference system with leave-one-out cross validation | PD vs HC: accuracy = 90.32% sensitivity = 86.67% specificity = 93.75% | 2018 | Ye et al |
| Classification of PD from HC and assess the severity of PD | Diagnosis | mPower database | 50; 22 HC + 28 PD | Random forest, bagged trees, SVM, KNN with 10-fold cross validation | random forest: PD vs HC accuracy = 87.03%<br><br>PD severity assessment accuracy = 85.8% | 2017 | Abujrida et al |
| Classification of PD from HC | Diagnosis | mPower database | 1,815; 866 HC + 949 PD | CNN with 10-fold cross validation | accuracy = 62.1% F1 score = 63.4% AUC = 63.5% | 2018 | Prince and de Vos |
| Classification of PD from HC | Diagnosis | dataset from Fernandez et al., 2013 | 49; 26 HC + 23 PD | KFD-RBF, naïve Bayes, KNN, SVM-RBF, random forest with 10-fold cross validation | random forest accuracy = 92.6% | 2015 | Wahid et al |

**Table 5.** Studies that applied machine learning models to movement or gait data to diagnose PD (n = 51).

### *3.6.3 Handwriting (n = 16)*

Fifteen out of 16 studies used accuracy in model evaluation and the average accuracy was 87.0 (6.3) % (Supplementary Table 1). Among these studies, the lowest accuracy was 76.44%[70] and the highest accuracy was 99.3%[14] (Figure 4, a, b). The highest accuracy per-study was obtained with neural network in 6 studies (37.5%), with SVM in 5 studies (31.3%), with ensemble learning in 4 studies (25.0%) and with naïve Bayes in 1 study (6.3%; Figure 4, c, d).

### *3.6.4 MRI (n = 36)*

Average accuracy of the 32 studies that used accuracy to evaluate the performance of machine learning models was 87.5 (8.0) %. In these studies, the lowest accuracy was 70.5%[71] and the highest accuracy was 100.0%[72] (Figure 4, a, b). Out of the 36 studies, the per-study highest accuracy was obtained with SVM in 21 studies (58.3%), with neural network in 8 studies (22.2%), with discriminant analysis in 3 studies (8.3%), with regression in 2 studies (5.6%) and with ensemble learning in 1 study (2.8%). One study (2.8%) obtained the highest per-study accuracy using models that do not belong to any of the given categories (Figure 4, c, d). In 8 of 36 studies, neural networks were directly applied to MRI data, while the remaining studies used machine learning models to learn from extracted features, e.g., cortical thickness and volume of brain regions, to diagnose PD.

Out of 17 studies that used MRI data from the PPMI database, 16 used accuracy to evaluate model performance and the average accuracy was 87.9 (8.0) %. The lowest and highest accuracies were 70.5% and 99.9%, respectively (Table 6). In 16 out of 19 studies that acquired MRI data from human participants, accuracy was used to evaluate classification performance and an average accuracy was 87.0 (8.1) % was achieved. The lowest reported accuracy was 76.2% and the highest reported accuracy was 100% (Table 6).

| **Objectives** | **Type of diagnosis** | **Source of data** | **Number of subjects (n)** | **Machine learning method(s)** | **Outcomes** | **Year** | **Reference** |
|---|---|---|---|---|---|---|---|
| Classification of PD from MSA | Differential diagnosis | collected from participants | 150; 54 HC + 65 PD + 31 MSA | SVM with leave-one-out-cross validation | MSA vs PD:<br>accuracy = 0.79<br>sensitivity = 0.71<br>specificity = 0.86<br><br>MSA vs HC:<br>accuracy = 0.79<br>sensitivity = 0.84<br>specificity = 0.74<br><br>MSA vs subsample of PD:<br>accuracy = 0.84<br>sensitivity = 0.77<br>specificity = 0.90 | 2019 | Abos et al |
| Classification of PD from MSA | Differential diagnosis | collected from participants | 151; 59 HC + 62 PD + 30 MSA | SVM with leave-one-out-cross validation | accuracy = 77.17%<br>sensitivity = 83.33%<br>specificity = 74.19% | 2019 | Baggio et al |
| Classification of PD from HC | Diagnosis | collected from participants | 94; 50 HC + 44 PD | CNN with 85 subjects for training and 9 for testing | training accuracy = 95.24%<br>testing accuracy = 88.88% | 2019 | Banerjee et al |
| Classification of PD from HC | Diagnosis | collected from participants | 47; 26 HC + 21 PD | SVM-linear with leave-one-out cross validation | accuracy = 93.62%<br>sensitivity = 90.47%<br>specificity = 96.15% | 2015 | Chen et al |
| Classification of PD from PSP | Differential diagnosis | collected from participants | 78; 57 PD + 21 PSP | SVM with leave-one-out cross validation | accuracy = 100%<br>sensitivity = 1<br>specificity = 1 | 2013 | Cherubini et al |
| Classification of PD, MSA, PSP and HC | Diagnosis and differential diagnosis | collected from participants | 106; 36 HC + 35 PD + 16 MSA + 19 PSP | Elastic Net regularized logistic regression with nested 10-fold cross validation | HC vs PD/MSA-P/PSP:<br>AUC = 0.88<br>sensitivity = 0.80<br>specificity = 0.83<br>PPV = 0.82<br>NPV = 0.81<br><br>HC vs PD:<br>AUC = 0.91<br>sensitivity = 0.86<br>specificity = 0.80<br>PPV = 0.82<br>NPV = 0.89<br><br>PD vs MSA/PSP:<br>AUC = 0.94<br>sensitivity = 0.86<br>specificity = 0.87<br>PPV = 0.88<br>NPV = 0.84<br><br>PD vs MSA:<br>AUC = 0.99<br>sensitivity = 0.97<br>specificity = 1.00<br>PPV = 1.00<br>NPV = 0.93<br><br>PD vs PSP:<br>AUC = 0.99<br>sensitivity = 0.97<br>specificity = 1.00<br>PPV = 1.00<br>NPV = 0.94<br><br>MSA vs PSP:<br>AUC = 0.98<br>sensitivity = 0.94<br>specificity = 1.00<br>PPV = 1.00<br>NPV = 0.93 | 2017 | Du et al |
| Classification of HC, PD, MSA and PSP | Diagnosis and differential diagnosis | collected from participants | 64; 22 HC + 21 PD + 11 MSA + 10 PSP | SVM-linear with leave-one-out cross validation | PD vs HC:<br>accuracy = 41.86%<br>sensitivity = 38.10%<br>specificity = 45.45%<br><br>PD vs MSA:<br>accuracy = 71.87%<br>sensitivity = 36.36%<br>specificity = 90.48%<br><br>PD vs PSP:<br>accuracy = 96.77%<br>sensitivity = 90%<br>specificity = 100%<br><br>MSA vs PSP:<br>accuracy = 76.19%<br><br>MSA vs HC:<br>accuracy = 78.78%<br>sensitivity = 54.55%<br>specificity = 90.91%<br><br>PSP vs HC:<br>accuracy = 93.75%<br>sensitivity = 90.00%<br>specificity = 95.45% | 2011 | Focke et al |
| Classification of PD and atypical PD | Differential diagnosis | collected from participants | 40; 17 PD + 23 atypical PD | SVM-RBF with 10-fold cross-validation | accuracy = 97.50%<br>TPR = 0.94<br>FPR = 0.00<br>TNR = 1.00<br>FNR = 0.06 | 2012 | Haller et al |
| Classification of PD and other forms of Parkinsonism | Differential diagnosis | collected from participants | 36; 16 PD + 20 other Parkinsonism | SVM-RBF with 10-fold cross validation | accuracy = 86.92%<br>TP = 0.87<br>FP = 0.14<br>TN = 0.87<br>FN = 0.13 | 2012 | Haller et al |
| Classification of HC, PD, PSP, MSA-C and MSA-P | Diagnosis and differential diagnosis | collected from participants | 464; 73 HC + 204 PD + 106 PSP + 21 MSA-C + 60 MSA-P | SVM-RBF with 10-fold cross validation | PD vs HC:<br>sensitivity = 65.2%<br>specificity = 67.1%<br>accuracy = 65.7%<br><br>PD vs PSP:<br>sensitivity = 82.5%<br>specificity = 86.8%<br>accuracy = 85.3%<br><br>PD vs MSA-C:<br>sensitivity = 76.2%<br>specificity = 96.1%<br>accuracy = 94.2%<br><br>PD vs MSA-P:<br>sensitivity = 86.7%<br>specificity = 92.2%<br>accuracy = 90.5% | 2016 | Huppertz et al |
| Classification of PD from HC | Diagnosis | collected from participants | 42; 21 HC + 21 PD | SVM-linear with stratified 10-fold cross validation | accuracy = 78.33%<br>precision = 85.00%<br>recall = 81.67%<br>AUC = 85.28% | 2017 | Kamagata et al |
| Classification of PD, PSP, MSA-P and HC | Diagnosis and differential diagnosis | collected from participants | 419; 142 HC + 125 PD + 98 PSP + 54 MSA-P | CNN with train-validation ratio of 85:15 | PD:<br>sensitivity = 94.4%<br>specificity = 97.8%<br>accuracy = 96.8%<br>AUC = 0.995<br><br>PSP:<br>sensitivity = 84.6%<br>specificity = 96.0%<br>accuracy = 93.7%<br>AUC = 0.982<br><br>MSA-P:<br>sensitivity = 77.8%<br>specificity = 98.1%<br>accuracy = 95.2%<br>AUC = 0.990<br><br>HC:<br>sensitivity = 100.0%<br>specificity = 97.5%<br>accuracy = 98.4%<br>AUC = 1.000 | 2019 | Kiryu et al |
| Classification of PD from HC | Diagnosis | collected from participants | 65; 31 HC + 34 PD | FCP with 36 out of the 65 subjects as the training set | AUC = 0.997 | 2016 | Liu et al |
| Classification of PD, PSP, MSA-C and MSA-P | Differential diagnosis | collected from participants | 85; 47 PD + 22 PSP + 9 MSA-C + 7 MSA-P | SVM-linear with leave-one-out cross validation | 4-class classification (MSA-C vs MSA-P vs PSP vs PD) accuracy = 88% | 2017 | Morisi et al |
| Classification of PD from HC | Diagnosis | collected from participants | 89; 47 HC + 42 PD | boosted logistic regression with nested cross-validation | accuracy = 76.2%<br>sensitivity = 81%<br>specificity = 72.7% | 2019 | Rubbert et al |
| Classification of PD, PSP and HC | Diagnosis and differential diagnosis | collected from participants | 84; 28 HC + 28 PSP + 28 PD | SVM-linear with leave-one-out cross validation | PD vs HC:<br>accuracy = 85.8%<br>specificity = 86.0%<br>sensitivity = 86.0%<br><br>PSP vs HC:<br>accuracy = 89.1%<br>specificity = 89.1%<br>sensitivity = 89.5%<br><br>PSP vs PD:<br>accuracy = 88.9%<br>specificity = 88.5%<br>sensitivity = 89.5% | 2014 | Salvatore et al |
| Classification of PD, APS (MSA, PSP) and HC | Diagnosis and differential diagnosis | collected from participants | 100; 35 HC + 45 PD + 20 APS | CNN-DL, CR-ML, RA-ML with 5-fold cross-validation | PD vs HC with CNN-DL:<br>test accuracy = 80.0%<br>test sensitivity = 0.86<br>test specificity = 0.70<br>test AUC = 0.913<br><br>PD vs APS with CNN-DL:<br>test accuracy = | 2019 | Shinde et al |
| | | | | | 85.7%<br>test sensitivity = 1.00<br>test specificity = 0.50<br>test AUC = 0.911 | | |
| Classification of PD from HC | Diagnosis | collected from participants | 101; 50 HC + 51 PD | SVM-RBF with leave-one-out cross validation | sensitivity = 92%<br>specificity = 87% | 2017 | Tang et al |
| Classification of PD from HC | Diagnosis | collected from participants | 85; 40 HC + 45 PD | SVM-linear with leave-one-out, 5-fold, .632-fold (1-1/e), 2-fold cross validation | accuracy = 97.7% | 2016 | Zeng et al |
| Classification of PD from HC | Diagnosis | PPMI database | 543; 169 HC + 374 PD | RLDA with JFSS with 10-fold cross validation | accuracy = 81.9% | 2016 | Adeli et al |
| Classification of PD from HC | Diagnosis | PPMI database | 543; 169 HC + 374 PD | RFS-LDA with 10-fold cross validation | accuracy = 79.8% | 2019 | Adeli et al |
| Classification of PD from HC | Diagnosis | PPMI database | 543; 169 HC + 374 PD | random forest (for feature selection and clinical score); SVM with 10-fold stratified cross validation | accuracy = 0.93<br>AUC = 0.97<br>sensitivity = 0.93<br>specificity = 0.92 | 2018 | Amoroso et al |
| Classification of PD, HC and prodromal | Diagnosis | PPMI database | 906; 203 HC + 66 prodromal + 637 PD | MLP, XgBoost, random forest, SVM with 5-fold cross validation | MLP:<br>accuracy = 95.3%<br>recall = 95.41%<br>precision = 97.28%<br>f1-score = 94% | 2020 | Chakraborty et al |
| Classification of PD from HC | Diagnosis | PPMI database | dataset 1:<br>15; 6 HC + 9 PD<br><br>dataset 2:<br>39; 21 HC + 18 PD | SVM with leave-one-out cross validation | dataset 1:<br>EER = 87%<br>accuracy = 80%<br>AUC = 0.907<br><br>dataset 2:<br>EER = 73%<br>accuracy = 68%<br>AUC = 0.780 | 2014 | Chen et al |
| Classification of PD from HC | Diagnosis | PPMI database | 80; 40 HC + 40 PD | naïve Bayes, SVM-RBF with 10-fold cross validation | SVM:<br>accuracy = 87.50%<br>sensitivity = 85.00%<br>specificity = 90.00%<br>AUC = 90.00% | 2019 | Cigdem et al |
| Classification of PD from HC | Diagnosis | PPMI database | 37; 18 HC + 19 PD | SVM-linear with leave-one-out cross validation | accuracy = 94.59% | 2017 | Kazeminejad et al |
| Classification of PD, HC and SWEDD | Diagnosis and subtyping | PPMI database | 238; 62 HC + 142 PD + 34 SWEDD | joint learning with 10-fold cross validation | HC vs PD:<br>accuracy = 91.12%<br>AUC = 94.88%<br><br>HC vs SWEDD:<br>accuracy = 94.89%<br>AUC = 97.80%<br><br>PD vs SWEDD:<br>accuracy = 92.12%<br>AUC = 93.82% | 2018 | Lei et al |
| Classification of PD and SWEDD from HC | Diagnosis | PPMI database | baseline: 238; 62 HC + 142 PD + 34 SWEDD<br><br>12 months: 186; 54 HC + 123 PD + 9 SWEDD<br><br>24 months: 127; 7 HC + 88 PD + 22 SWEDD | SSAE with 10-fold cross validation | HC vs PD: accuracy = 85.24%, 88.14% and 96.19% for baseline, 12 m, and 24 m<br><br>HC vs SWEDD: accuracy = 89.67%, 95.24% and 93.10% for baseline, 12 m, and 24 m | 2019 | Li et al |
| Classification of PD from HC | Diagnosis | PPMI database | 112; 56 HC + 56 PD | RLDA with 8-fold cross validation | accuracy = 70.5%<br>AUC = 71.1 | 2016 | Liu et al |
| Classification of PD from HC | Diagnosis | PPMI database | 60; 30 HC + 30 PD | SVM, ELM with train-test ratio of 80:20 | ELM:<br>training accuracy = 94.87%<br>testing accuracy = 90.97%<br>sensitivity = 0.9245<br>specificity = 0.9730 | 2016 | Pahuja and Nagabhushan |
| Classification of PD from HC | Diagnosis | PPMI database | 172; 103 HC + 69 PD | multi-kernel SVM with 10-fold cross validation | accuracy = 85.78%<br>specificity = 87.79%<br>sensitivity = 87.64%<br>AUC = 0.8363 | 2017 | Peng et al |
| Classification of PD from HC | Diagnosis and subtyping | PPMI database | 109; 32 HC + 77 PD (55 PD-NC + 22 PD-MCI) | SVM with 2-fold cross validation | PD vs. HC:<br>accuracy = 92.35%<br>sensitivity = 0.9035<br>specificity = 0.9431<br>AUC = 0.9744<br><br>PD-MCI vs. HC:<br>accuracy = 83.91%<br>sensitivity = 0.8355<br>specificity = 0.8587<br>AUC = 0.9184<br><br>PD-MCI vs. PD-NC:<br>accuracy = 80.84%<br>sensitivity = 0.7705<br>specificity = 0.8457<br>AUC = 0.8677 | 2016 | Peng et al |
| Classification of PD, HC and SWEDD | Diagnosis and subtyping | PPMI database | 831; 245 HC + 518 PD + 68 SWEDD | LSSVM-RBF with cross validation | accuracy = 99.9%<br>specificity = 100%<br>sensitivity = 99.4% | 2015 | Singh and Samavedham |
| Classification of PD, HC and SWEDD | Diagnosis and differential diagnosis | PPMI database | 741; 262 HC + 408 PD + 71 SWEDD | LSSVM-RBF with 10-fold cross validation | PD vs HC accuracy = 95.37%<br>PD vs SWEDD accuracy = 96.04%<br>SWEDD vs HC accuracy = 93.03% | 2018 | Singh et al |
| Classification of PD from HC | Diagnosis | PPMI database | 408; 204 HC + 204 PD | CNN (VGG and ResNet) | ResNet50 accuracy = 88.6% | 2019 | Yagis et al |
| Classification of PD from HC | Diagnosis | PPMI database | 754; 158 HC + 596 PD | FCN, GCN with 5-fold cross validation | AUC = 95.37% | 2018 | Zhang et al |

**Table 6.** Studies that applied machine learning models to MRI data to diagnose PD (n = 36).

#### *3.6.5 SPECT (n = 14)*

Average accuracy of 12 out of 14 studies that used accuracy to measure the performance of machine learning models was 94.4 (4.2) % (Supplementary Table 1). The lowest reported accuracy was 83.2%[42] and 97.9%[73] (Figure 4, a, b). SVM led to the highest per-study accuracy in 10 out of 14 studies (71.4%). The highest per-study accuracy was obtained with neural networks in 3 studies (21.4%) and with regression in 1 study (7.1%; Figure 4, c, d).

#### *3.6.6 PET (n = 4)*

All 4 studies used sensitivity and specificity (Supplementary Table 1) in model evaluation while 3 used accuracy. Average accuracy of the 3 studies was 85.6 (6.6) %, with a lowest accuracy of 78.16%[74] and a highest accuracy of 90.72%[75] (Figure 4, a, b). Half of the 4 studies (50.0%) obtained the highest per-study accuracy with SVM[74, 75] and the other half (50.0%) with neural networks (Figure 4, c, d).

#### *3.6.7 CSF (n = 5)*

All 5 studies used AUC, instead of accuracy, to evaluate machine learning models (Supplementary Table 1). The average AUC was 0.8 (0.1), the lowest AUC was 0.6825[24] and the highest AUC was 0.839[76], respectively. Two studies obtained the highest per-study AUC with ensemble learning, 2 studies with SVM and 1 study with regression (Figure 4, c, d).

#### *3.6.8 Other types of data (n = 10)*

Only 5 studies used accuracy to measure the performance of machine learning models (Supplementary Table 1). An average accuracy of 91.9 (6.4) % was obtained, with a lowest accuracy of 84.85%[55] and a highest accuracy of 100%[25] (Figure 4, a, b). Out of the 10 studies, 5 (50%) used SVM to achieve the per-study highest accuracy, 3 (30%) used ensemble learning, 1 (10%) used decision trees and 1 (10%) used machine learning models that do not belong to any given categories (Figure 4, c, d).

#### *3.6.9 Combination of more than one data type (n = 18)*

Out of the 18 studies that used more than one type of data, 15 used accuracy in model evaluation (Supplementary Table 1). An average accuracy of 92.6 (6.1) % was obtained, and the lowest and highest accuracy among the 15 studies was 82.0%[77] and 100.0%[28], respectively (Figure 4, a, b). The per-study highest accuracy was achieved with ensemble learning in 6 studies (33.3%), with neural network in 5 studies (27.8%), with SVM in 4 studies (22.2%), with regression in 1 (5.6%) study and with nearest neighbor (5.6%) in 1 study. One study (5.6%) used machine learning models that do not belong to any given categories to obtain the highest per-study accuracy (Figure 4, c, d).

# 4 Discussion

## 4.1 Principal findings

In this review, we present results from published studies that applied machine learning to the diagnosis and differential diagnosis of PD. Since the number of included papers was relatively large, we focused on a high-level summary rather than a detailed description of methodology and outcomes of individual studies. We also provide an overview of sample size, data source and data type, for a more in-depth understanding of methodological differences across studies. Furthermore, we assessed (a) how large the participant pool/dataset was, (b) to what extend original data (i.e., data acquired from local human participants) were collected and used, (c) the feasibility of machine learning and the possibility of adapting new biomarkers in the diagnosis of PD.

As a result, a per-study diagnostic accuracy above chance levels was achieved in all studies that used accuracy in model evaluation (Figure 4, a). Apart from studies using CSF data that measured model performance with AUC, classification accuracy associated with 8 other data types ranged between 85.6% (PET) and 94.4% (SPECT), with an average of 89.9 (3.0) %. Therefore, although the small number of studies of some data types may not allow for a generalizable prediction of how well these data types can help us differentiate PD from HC or atypical Parkinsonian disorders, the application of machine learning to a variety of data types led to high accuracy in the diagnosis of PD. In addition, an accuracy significantly above chance level was achieved in all machine learning models (Supplementary Table 2), while SVM, neural networks and ensemble learning were among the most popular model choices, all yielding great applicability to a variety of modalities. In the meantime, when compared with other models, they led to the per-study highest classification accuracy in > 50% of all cases (50.7%, 51.9% and 52.3%; Supplementary Table 2).

As previously discussed[31], although satisfactory diagnostic outcomes could be achieved, sample size in few studies was extremely small (< 15 subjects). The application of some machine learning models, especially neural networks, typically rely on a large dataset. Nevertheless, collecting data from a large pool of participants remains challenging in clinical studies, and data generated are commonly of high dimensionality and small sample size[78]. To address this challenge, one solution is to combine data from a local cohort with public repositories including PPMI, UCI machine learning repository, PhysioNet and many others, depending on the type of data that have been collected from the local cohort. Furthermore, when a great difference in group size is observed (i.e., class imbalance problem), labeling all samples after the majority class may lead to an undesired high accuracy. In this case, evaluating machine learning models with other metrics including precision, recall and F-1 score is recommended[79].

Even though high diagnostic accuracy of PD has been achieved in clinical settings, machine learning approaches have also reached high accuracy as shown in the present study, while models including SVM and neural networks are particularly useful in (a) diagnosis of PD using data modalities that have been overlooked in clinical decision making (e.g., voice), and (b) identification of features of high relevance from these data. For example, the use of machine learning models with feature selection techniques allows for assessing the relative importance of features of a large feature space in order to select the most differentiating ones, which is conventionally challenging using manual approaches. In the meantime, diagnosing PD using more than one data modality has led to promising results. Accordingly, supplying clinicians with non-motor data and machine learning approaches may support clinical decision making in patients with ambiguous symptom presentations, and/or improve diagnosis at an earlier stage.

### 4.2 Limitations

In the present study, we have excluded research articles in languages other than English and results published in the form of conference abstracts, posters and talks. Despite the ongoing discussion of advantages and importance of including conference abstracts in systematic reviews[80], conference abstracts often do not report sufficient key information which is why we had to exclude them. However, this may lead to a publication and result bias. Moreover, due to the high inter-study variance in the presentation of results, it was challenging to compare outcomes associated with each type of model across studies, as some studies failed to indicate whether model performance was evaluated using a test set, and/or results given by models that did not yield the best per-study performance. Results of published studies were discussed and summarized based on data and machine learning models used, and for data modalities such as PET (n = 4) or CSF (n = 5), the number of studies were too small despite the high total number of studies included. Therefore, it was improbable to assess the general performance of machine learning techniques when PET or CSF data are used.

Another issue was insufficient or inaccurate description of methods or results, as some studies failed to provide accurate information of the number and type of subjects used, or how machine learning models were implemented, trained and tested. Infrequently, authors skipped basic information such as number of subjects and their medical conditions and referred to another publication. Although we attempted to list hyperparameters and cross-validation strategies in the data extraction table, many included studies did not make this information available in the main text, leading to potential difficulties in replication of results. Apart from these, rounding errors or inconsistent reporting of results also exist. Furthermore, although we treated the differentiation of PD from SWEDD as subtyping, there is ongoing controversy regarding whether it should be considered as differential diagnosis or subtyping[35, 81-83]. Given these limitations, clinicians interested in adapting machine learning models or implementing diagnostic systems based on novel biomarkers are advised to interpret results with care. Further, this strengthens the need for uniform reporting standards in studies using machine learning.

# 5 Conclusions

To the best of our knowledge, the present study is the first exhaustive review which included results from all studies that applied machine learning methods to the diagnosis of PD. Here, we presented included studies in a high-level summary, providing access to information including (a) machine learning methods that have been used in the diagnosis of PD and associated outcomes, (b) types of clinical, behavioral and biometric data that could be used for rendering more accurate diagnoses, (c) potential biomarkers for assisting clinical decision making and (d) other highly relevant information, including databases that could be used to enlarge and enrich the dataset. In summary, realization of machine learning-assisted diagnosis of PD yields high potential for a more systematic clinical decision-making system, while adaptation of novel biomarkers gives rise to easier access to PD diagnosis at an earlier stage. Machine learning approaches therefore have the potential to provide clinicians with additional tools to screen, detect or diagnose PD.

# Acknowledgements:

We thank Dr. Antje Haehner for her comments on the manuscript. This work was supported by the Natural Sciences and Engineering Research Council of Canada (NSERC) and Québec Bio-Imaging Network. J.M. is supported by the Québec Bio-Imaging Network Postdoctoral Fellowship (FRSQ – Réseaux de recherche thématiques; Dossier: 35450). J.F. holds a Research Chair in Chemosensory Neuroanatomy at UQTR.

## References

[1] O.-B. Tysnes, A. Storstein, Epidemiology of Parkinson’s disease, Journal of Neural Transmission 124(8) (2017) 901-905.
[2] E.R. Dorsey, A. Elbaz, E. Nichols, F. Abd-Allah, A. Abdelalim, J.C. Adsuar, M.G. Ansha, C. Brayne, J.-Y.J. Choi, D. Collado-Mateo, Global, regional, and national burden of Parkinson's disease, 1990–2016: a systematic analysis for the Global Burden of Disease Study 2016, The Lancet Neurology 17(11) (2018) 939-953.
[3] J. Jankovic, Parkinson’s disease: clinical features and diagnosis, Journal of neurology, neurosurgery & psychiatry 79(4) (2008) 368-376.
[4] J. Contreras-Vidal, G.E. Stelmach, Effects of Parkinsonism on motor control, Life Sciences 58(3) (1995) 165-176.
[5] J. Opara, W. Brola, M. Leonardi, B. Błaszczyk, Quality of life in Parkinsons Disease, Journal of medicine and life 5(4) (2012) 375.
[6] S.J. Johnson, M.D. Diener, A. Kaltenboeck, H.G. Birnbaum, A.D. Siderowf, An economic model of P arkinson's disease: Implications for slowing progression in the United States, Movement Disorders 28(3) (2013) 319-326.
[7] S.L. Kowal, T.M. Dall, R. Chakrabarti, M.V. Storm, A. Jain, The current and projected economic burden of Parkinson's disease in the United States, Movement Disorders 28(3) (2013) 311-318.
[8] J.-X. Yang, L. Chen, Economic burden analysis of Parkinson’s disease patients in China, Parkinson’s Disease 2017 (2017).
[9] C. Tremblay, P.D. Martel, J. Frasnelli, Trigeminal system in Parkinson's disease: A potential avenue to detect Parkinson-specific olfactory dysfunction, Parkinsonism Relat Disord 44 (2017) 85-90.
[10] T.A. Zesiewicz, K.L. Sullivan, R.A. Hauser, Nonmotor symptoms of Parkinson’s disease, Expert review of neurotherapeutics 6(12) (2006) 1811-1822.
[11] H. Braak, K. Del Tredici, U. Rüb, R.A. De Vos, E.N.J. Steur, E. Braak, Staging of brain pathology related to sporadic Parkinson’s disease, Neurobiology of aging 24(2) (2003) 197-211.
[12] R.B. Postuma, D. Berg, M. Stern, W. Poewe, C.W. Olanow, W. Oertel, J. Obeso, K. Marek, I. Litvan, A.E. Lang, MDS clinical diagnostic criteria for Parkinson's disease, Movement Disorders 30(12) (2015) 1591-1601.
[13] P. Drotár, J. Mekyska, I. Rektorová, L. Masarová, Z. Smékal, M. Faundez-Zanuy, Decision support framework for Parkinson's disease based on novel handwriting markers, IEEE Trans Neural Syst Rehabil Eng 23(3) (2015) 508-516.
[14] C.R. Pereira, D.R. Pereira, G.H. Rosa, V.H.C. Albuquerque, S.A.T. Weber, C. Hook, J.P. Papa, Handwritten dynamics assessment through convolutional neural networks: An application to Parkinson's disease identification, Artif Intell Med 87 (2018) 67-77.
[15] M. Yang, H. Zheng, H. Wang, S. McClean, Feature selection and construction for the discrimination of neurodegenerative diseases based on gait analysis, 2009 3rd International Conference on Pervasive Computing Technologies for Healthcare, 2009, pp. 1-7.
[16] F. Wahid, R.K. Begg, C.J. Hass, S. Halgamuge, D.C. Ackland, Classification of Parkinson's Disease Gait Using Spatial-Temporal Gait Features, IEEE J Biomed Health Inform 19(6) (2015) 1794-1802.
[17] T.D. Pham, H. Yan, Tensor Decomposition of Gait Dynamics in Parkinson's Disease, IEEE Trans Biomed Eng 65(8) (2018) 1820-1827.
[18] A. Cherubini, M. Morelli, R. Nisticó, M. Salsone, G. Arabia, R. Vasta, A. Augimeri, M.E. Caligiuri, A. Quattrone, Magnetic resonance support vector machine discriminates between Parkinson disease and progressive supranuclear palsy, Mov Disord 29(2) (2014) 266-269.

[19] H. Choi, S. Ha, H.J. Im, S.H. Paek, D.S. Lee, Refining diagnosis of Parkinson's disease with deep learning-based interpretation of dopamine transporter imaging, Neuroimage Clin 16 (2017) 586-594.
[20] F. Segovia, J.M. Górriz, J. Ramírez, F.J. Martínez-Murcia, D. Castillo-Barnes, Assisted Diagnosis of Parkinsonism Based on the Striatal Morphology, Int J Neural Syst 29(9) (2019) 1950011-1950011.
[21] C. Ma, J. Ouyang, H.-L. Chen, X.-H. Zhao, An efficient diagnosis system for Parkinson's disease using kernel-based extreme learning machine with subtractive clustering features weighting approach, Comput Math Methods Med 2014 (2014) 985789-985789.
[22] B.E. Sakar, M.E. Isenkul, C.O. Sakar, A. Sertbas, F. Gurgen, S. Delil, H. Apaydin, O. Kursun, Collection and analysis of a Parkinson speech dataset with multiple types of sound recordings, IEEE J Biomed Health Inform 17(4) (2013) 828-834.
[23] P.A. Lewitt, J. Li, M. Lu, T.G. Beach, C.H. Adler, L. Guo, C. Arizona Parkinson's Disease, 3-hydroxykynurenine and other Parkinson's disease biomarkers discovered by metabolomic analysis, Mov Disord 28(12) (2013) 1653-1660.
[24] F. Maass, B. Michalke, D. Willkommen, A. Leha, C. Schulte, L. Tönges, B. Mollenhauer, C. Trenkwalder, D. Rückamp, M. Börger, I. Zerr, M. Bähr, P. Lingor, Elemental fingerprint: Reassessment of a cerebrospinal fluid biomarker for Parkinson's disease, Neurobiol Dis 134 (2020) 104677-104677.
[25] S. Nuvoli, A. Spanu, M.L. Fravolini, F. Bianconi, S. Cascianelli, G. Madeddu, B. Palumbo, [(123)I]Metaiodobenzylguanidine (MIBG) Cardiac Scintigraphy and Automated Classification Techniques in Parkinsonian Disorders, Mol Imaging Biol (2019) 10.1007/s11307-019-01406-6.
[26] C. Váradi, K. Nehéz, O. Hornyák, B. Viskolcz, J. Bones, Serum N-Glycosylation in Parkinson's Disease: A Novel Approach for Potential Alterations, Molecules 24(12) (2019) 2220.
[27] A. Nunes, G. Silva, C. Duque, C. Januário, I. Santana, A.F. Ambrósio, M. Castelo-Branco, R. Bernardes, Retinal texture biomarkers may help to discriminate between Alzheimer's, Parkinson's, and healthy controls, PLoS One 14(6) (2019) e0218826-e0218826.
[28] A. Cherubini, R. Nisticó, F. Novellino, M. Salsone, S. Nigro, G. Donzuso, A. Quattrone, Magnetic resonance support vector machine discriminates essential tremor with rest tremor from tremor-dominant Parkinson disease, Mov Disord 29(9) (2014) 1216-1219.
[29] Z. Wang, X. Zhu, E. Adeli, Y. Zhu, F. Nie, B. Munsell, G. Wu, Adni, Ppmi, Multi-modal classification of neurodegenerative disease by progressive graph-based transductive learning, Med Image Anal 39 (2017) 218-230.
[30] C. Ahlrichs, M. Lawo, Parkinson's disease motor symptoms in machine learning: A review, arXiv preprint arXiv:1312.3825 (2013).
[31] M. Belić, V. Bobić, M. Badža, N. Šolaja, M. Đurić-Jovičić, V.S. Kostić, Artificial intelligence for assisting diagnostics and assessment of Parkinson's disease–A review, Clin Neurol Neurosurg (2019) 105442.
[32] R.A. Ramdhani, A. Khojandi, O. Shylo, B.H. Kopell, Optimizing clinical assessments in Parkinson's disease through the use of wearable sensors and data driven modeling, Frontiers in computational neuroscience 12 (2018) 72.
[33] C.R. Pereira, D.R. Pereira, S.A. Weber, C. Hook, V.H.C. de Albuquerque, J.P. Papa, A survey on computer-assisted Parkinson's disease diagnosis, Artif Intell Med 95 (2019) 48-63.
[34] D. Moher, A. Liberati, J. Tetzlaff, D.G. Altman, Preferred reporting items for systematic reviews and meta-analyses: the PRISMA statement, Annals of internal medicine 151(4) (2009) 264-269.
[35] R. Erro, S.A. Schneider, M. Stamelou, N.P. Quinn, K.P. Bhatia, What do patients with scans without evidence of dopaminergic deficit (SWEDD) have? New evidence and continuing controversies, Journal of Neurology, Neurosurgery & Psychiatry 87(3) (2016) 319-323.
[36] D. Dua, C. Graff, UCI Machine Learning Repository. University of California, School of Information and Computer Science, Irvine, CA (2017), 2018.

[37] K. Marek, D. Jennings, S. Lasch, A. Siderowf, C. Tanner, T. Simuni, C. Coffey, K. Kieburtz, E. Flagg, S. Chowdhury, The parkinson progression marker initiative (PPMI), Progress in neurobiology 95(4) (2011) 629-635.
[38] A.L. Goldberger, L.A. Amaral, L. Glass, J.M. Hausdorff, P.C. Ivanov, R.G. Mark, J.E. Mietus, G.B. Moody, C.-K. Peng, H.E. Stanley, PhysioBank, PhysioToolkit, and PhysioNet: components of a new research resource for complex physiologic signals, circulation 101(23) (2000) e215-e220.
[39] C.R. Pereira, D.R. Pereira, F.A. da Silva, C. Hook, S.A. Weber, L.A. Pereira, J.P. Papa, A step towards the automated diagnosis of parkinson's disease: Analyzing handwriting movements, 2015 IEEE 28th international symposium on computer-based medical systems, IEEE, 2015, pp. 171-176.
[40] B.M. Bot, C. Suver, E.C. Neto, M. Kellen, A. Klein, C. Bare, M. Doerr, A. Pratap, J. Wilbanks, E.R. Dorsey, The mPower study, Parkinson disease mobile data collected using ResearchKit, Scientific data 3(1) (2016) 1-9.
[41] S. Bhati, L.M. Velazquez, J. Villalba, N. Dehak, LSTM Siamese Network for Parkinson’s Disease Detection from Speech, 2019 IEEE Global Conference on Signal and Information Processing (GlobalSIP), 2019, pp. 1-5.
[42] S.-Y. Hsu, H.-C. Lin, T.-B. Chen, W.-C. Du, Y.-H. Hsu, Y.-C. Wu, P.-W. Tu, Y.-H. Huang, H.-Y. Chen, Feasible Classified Models for Parkinson Disease from (99m)Tc-TRODAT-1 SPECT Imaging, Sensors (Basel) 19(7) (2019) 1740.
[43] J. Mucha, J. Mekyska, M. Faundez-Zanuy, K. Lopez-De-Ipina, V. Zvoncak, Z. Galaz, T. Kiska, Z. Smekal, L. Brabenec, I. Rektorova, Advanced Parkinson's Disease Dysgraphia Analysis Based on Fractional Derivatives of Online Handwriting, 2018 10th International Congress on Ultra Modern Telecommunications and Control Systems and Workshops (ICUMT), 2018, pp. 1-6.
[44] C. Taleb, M. Khachab, C. Mokbel, L. Likforman-Sulem, Visual Representation of Online Handwriting Time Series for Deep Learning Parkinson's Disease Detection, 2019 International Conference on Document Analysis and Recognition Workshops (ICDARW), 2019, pp. 25-30.
[45] A. Vlachostergiou, A. Tagaris, A. Stafylopatis, S. Kollias, Multi-Task Learning for Predicting Parkinson's Disease Based on Medical Imaging Information, 2018 25th IEEE International Conference on Image Processing (ICIP), 2018, pp. 2052-2056.
[46] M. Wodzinski, A. Skalski, D. Hemmerling, J.R. Orozco-Arroyave, E. Nöth, Deep Learning Approach to Parkinson’s Disease Detection Using Voice Recordings and Convolutional Neural Network Dedicated to Image Classification, 2019 41st Annual International Conference of the IEEE Engineering in Medicine and Biology Society (EMBC), 2019, pp. 717-720.
[47] A. Agarwal, S. Chandrayan, S.S. Sahu, Prediction of Parkinson's disease using speech signal with Extreme Learning Machine, 2016 International Conference on Electrical, Electronics, and Optimization Techniques (ICEEOT), 2016, pp. 3776-3779.
[48] J.C. Taylor, J.W. Fenner, Comparison of machine learning and semi-quantification algorithms for (I123)FP-CIT classification: the beginning of the end for semi-quantification?, EJNMMI Phys 4(1) (2017) 29-29.
[49] K.M. Fernandez, R.T. Roemmich, E.L. Stegemöller, S. Amano, A. Thompson, M.S. Okun, C.J. Hass, Gait initiation impairments in both Essential Tremor and Parkinson's disease, Gait Posture 38(4) (2013) 956-961.
[50] P. Kugler, C. Jaremenko, J. Schlachetzki, J. Winkler, J. Klucken, B. Eskofier, Automatic recognition of Parkinson's disease using surface electromyography during standardized gait tests, Conf Proc IEEE Eng Med Biol Soc 2013 (2013) 5781-5784.
[51] J.M. Tracy, Y. Özkanca, D.C. Atkins, R. Hosseini Ghomi, Investigating voice as a biomarker: Deep phenotyping methods for early detection of Parkinson's disease, J Biomed Inform (2019) 103362-103362.

[52] M. Cibulka, M. Brodnanova, M. Grendar, M. Grofik, E. Kurca, I. Pilchova, O. Osina, Z. Tatarkova, D. Dobrota, M. Kolisek, SNPs rs11240569, rs708727, and rs823156 in SLC41A1 Do Not Discriminate Between Slovak Patients with Idiopathic Parkinson's Disease and Healthy Controls: Statistics and Machine-Learning Evidence, Int J Mol Sci 20(19) (2019) 4688.
[53] R. Prashanth, S. Dutta Roy, Early detection of Parkinson's disease through patient questionnaire and predictive modelling, Int J Med Inform 119 (2018) 75-87.
[54] R. Shamir, C. Klein, D. Amar, E.-J. Vollstedt, M. Bonin, M. Usenovic, Y.C. Wong, A. Maver, S. Poths, H. Safer, J.-C. Corvol, S. Lesage, O. Lavi, G. Deuschl, G. Kuhlenbaeumer, H. Pawlack, I. Ulitsky, M. Kasten, O. Riess, A. Brice, B. Peterlin, D. Krainc, Analysis of blood-based gene expression in idiopathic Parkinson disease, Neurology 89(16) (2017) 1676-1683.
[55] J. Shi, M. Yan, Y. Dong, X. Zheng, Q. Zhang, H. An, Multiple Kernel Learning Based Classification of Parkinson’s Disease With Multi-Modal Transcranial Sonography, 2018 40th Annual International Conference of the IEEE Engineering in Medicine and Biology Society (EMBC), 2018, pp. 61-64.
[56] P.-H. Tseng, I.G.M. Cameron, G. Pari, J.N. Reynolds, D.P. Munoz, L. Itti, High-throughput classification of clinical populations from natural viewing eye movements, J Neurol 260(1) (2013) 275-284.
[57] M.I. Vanegas, M.F. Ghilardi, S.P. Kelly, A. Blangero, Machine learning for EEG-based biomarkers in Parkinson’s disease, 2018 IEEE International Conference on Bioinformatics and Biomedicine (BIBM), 2018, pp. 2661-2665.
[58] L. Ali, S.U. Khan, M. Arshad, S. Ali, M. Anwar, A Multi-model Framework for Evaluating Type of Speech Samples having Complementary Information about Parkinson's Disease, 2019 International Conference on Electrical, Communication, and Computer Engineering (ICECCE), 2019, pp. 1-5.
[59] P. Kraipeerapun, S. Amornsamankul, Using stacked generalization and complementary neural networks to predict Parkinson's disease, 2015 11th International Conference on Natural Computation (ICNC), 2015, pp. 1290-1294.
[60] R.H. Abiyev, S. Abizade, Diagnosing Parkinson's Diseases Using Fuzzy Neural System, Comput Math Methods Med 2016 (2016) 1267919-1267919.
[61] L. Ali, C. Zhu, Z. Zhang, Y. Liu, Automated Detection of Parkinson’s Disease Based on Multiple Types of Sustained Phonations Using Linear Discriminant Analysis and Genetically Optimized Neural Network, IEEE Journal of Translational Engineering in Health and Medicine 7 (2019) 1-10.
[62] N.K. Dastjerd, O.C. Sert, T. Ozyer, R. Alhajj, Fuzzy Classification Methods Based Diagnosis of Parkinson's disease from Speech Test Cases, Curr Aging Sci 12(2) (2019) 100-120.
[63] M. Hariharan, K. Polat, R. Sindhu, A new hybrid intelligent system for accurate detection of Parkinson's disease, Comput Methods Programs Biomed 113(3) (2014) 904-913.
[64] I. Mandal, N. Sairam, Accurate telemonitoring of Parkinson's disease diagnosis using robust inference system, Int J Med Inform 82(5) (2013) 359-377.
[65] J. Prince, M. de Vos, A Deep Learning Framework for the Remote Detection of Parkinson'S Disease Using Smart-Phone Sensor Data, Conf Proc IEEE Eng Med Biol Soc 2018 (2018) 3144-3147.
[66] T.D. Pham, Pattern analysis of computer keystroke time series in healthy control and early-stage Parkinson's disease subjects using fuzzy recurrence and scalable recurrence network features, J Neurosci Methods 307 (2018) 194-202.
[67] D. Surangsrirat, C. Thanawattano, R. Pongthornseri, S. Dumnin, C. Anan, R. Bhidayasiri, Support vector machine classification of Parkinson's disease and essential tremor subjects based on temporal fluctuation, Conf Proc IEEE Eng Med Biol Soc 2016 (2016) 6389-6392.
[68] D. Joshi, A. Khajuria, P. Joshi, An automatic non-invasive method for Parkinson's disease classification, Comput Methods Programs Biomed 145 (2017) 135-145.

[69] J.P. Félix, F.H.T. Vieira, Á.A. Cardoso, M.V.G. Ferreira, R.A.P. Franco, M.A. Ribeiro, S.G. Araújo, H.P. Corrêa, M.L. Carneiro, A Parkinson's Disease Classification Method: An Approach Using Gait Dynamics and Detrended Fluctuation Analysis, 2019 IEEE Canadian Conference of Electrical and Computer Engineering (CCECE), 2019, pp. 1-4.
[70] L. Ali, C. Zhu, N.A. Golilarz, A. Javeed, M. Zhou, Y. Liu, Reliable Parkinson's Disease Detection by Analyzing Handwritten Drawings: Construction of an Unbiased Cascaded Learning System Based on Feature Selection and Adaptive Boosting Model, IEEE Access 7 (2019) 116480-116489.
[71] L. Liu, Q. Wang, E. Adeli, L. Zhang, H. Zhang, D. Shen, Feature Selection Based on Iterative Canonical Correlation Analysis for Automatic Diagnosis of Parkinson's Disease, Med Image Comput Comput Assist Interv 9901 (2016) 1-8.
[72] O. Cigdem, H. Demirel, D. Unay, The Performance of Local-Learning Based Clustering Feature Selection Method on the Diagnosis of Parkinson's Disease Using Structural MRI, 2019 IEEE International Conference on Systems, Man and Cybernetics (SMC), 2019, pp. 1286-1291.
[73] F.P.M. Oliveira, D.B. Faria, D.C. Costa, M. Castelo-Branco, J.M.R.S. Tavares, Extraction, selection and comparison of features for an effective automated computer-aided diagnosis of Parkinson's disease based on [(123)I]FP-CIT SPECT images, Eur J Nucl Med Mol Imaging 45(6) (2018) 1052-1062.
[74] F. Segovia, J.M. Górriz, J. Ramírez, J. Levin, M. Schuberth, M. Brendel, A. Rominger, G. Garraux, C. Phillips, Analysis of 18F-DMFP PET data using multikernel classification in order to assist the diagnosis of Parkinsonism, 2015 IEEE Nuclear Science Symposium and Medical Imaging Conference (NSS/MIC), 2015, pp. 1-4.
[75] Y. Wu, J.-H. Jiang, L. Chen, J.-Y. Lu, J.-J. Ge, F.-T. Liu, J.-T. Yu, W. Lin, C.-T. Zuo, J. Wang, Use of radiomic features and support vector machine to distinguish Parkinson's disease cases from normal controls, Ann Transl Med 7(23) (2019) 773-773.
[76] F. Maass, B. Michalke, A. Leha, M. Boerger, I. Zerr, J.-C. Koch, L. Tönges, M. Bähr, P. Lingor, Elemental fingerprint as a cerebrospinal fluid biomarker for the diagnosis of Parkinson's disease, J Neurochem 145(4) (2018) 342-351.
[77] J. Prince, F. Andreotti, M.D. Vos, Multi-Source Ensemble Learning for the Remote Prediction of Parkinson's Disease in the Presence of Source-Wise Missing Data, IEEE Transactions on Biomedical Engineering 66(5) (2019) 1402-1411.
[78] A. Vabalas, E. Gowen, E. Poliakoff, A.J. Casson, Machine learning algorithm validation with a limited sample size, PLoS One 14(11) (2019).
[79] L.A. Jeni, J.F. Cohn, F. De La Torre, Facing imbalanced data--recommendations for the use of performance metrics, 2013 Humaine association conference on affective computing and intelligent interaction, IEEE, 2013, pp. 245-251.
[80] R.W. Scherer, I.J. Saldanha, How should systematic reviewers handle conference abstracts? A view from the trenches, Systematic reviews 8(1) (2019) 264.
[81] K.L. Chou, Diagnosis and differential diagnosis of Parkinson disease, Waltham (MA): UpToDate (2017).
[82] D.-Y. Kwon, Y. Kwon, J.-W. Kim, Quantitative analysis of finger and forearm movements in patients with off state early stage Parkinson's disease and scans without evidence of dopaminergic deficit (SWEDD), Parkinsonism Relat Disord 57 (2018) 33-38.
[83] M.J. Lee, S.L. Kim, C.H. Lyoo, M.S. Lee, Kinematic analysis in patients with Parkinson's disease and SWEDD, J Parkinsons Dis 4(3) (2014) 421-430.

## Supplementary materials for:

# Machine learning for the diagnosis of Parkinson's disease: A systematic review

Jie Mei[a*], Christian Desrosiers[b] and Johannes Frasnelli[a,c]

a. Department of Anatomy, Université du Québec à Trois-Rivières (UQTR), Trois-Rivières, QC, Canada
b. Department of Software and IT Engineering, École de technologie supérieure, 1100 Notre-Dame St W, Montreal, QC H3C 1K3, Canada
c. Centre de Recherche de l'Hôpital du Sacré-Coeur de Montréal, Centre intégré universitaire de santé et de services sociaux du Nord-de-l'Île-de-Montréal (CIUSSS du Nord-de-l'Île-de-Montréal), Canada

***Correspondence to:** Jie Mei, Department of Anatomy, Université du Québec à Trois-Rivières, 3351, boul. des Forges, C.P. 500, Trois-Rivières, QC G9A 5H7 Canada; Email: jie.mei@uqtr.ca

## Machine learning methods and associated outcomes

***SVM (n = 130)***

SVM was used to analyze movement or gait data (n = 37), voice recordings (n = 32), MRI (n = 23), SPECT (n = 10), handwriting (n = 10), CSF (n = 2) and PET (n = 2). Other data such as EMG (Kugler et al., 2013), OCT imaging (Nunes et al., 2019), cardiac scintigraphy (Nuvoli et al., 2019), Patient Questionnaire of MDS-UPDRS (Prashanth and Dutta Roy, 2018), whole-blood gene expression profiles (Shamir et al., 2017) and eye movements during natural viewing (Tseng et al., 2013) have also been used to train SVM. Eight studies used more than one data types (Supplementary Figure 1).

In the 57 studies that did not compare SVM with other models, 48 used accuracy in model evaluation, leading to an average of 90.5 (7.6)% and a range between 74.0% (Chen et al., 2014) to 100% (Cherubini et al., 2014; Cherubini et al., 2014; Joshi et al., 2017; Pham, 2018; Pham and Yan, 2018; Surangsrirat et al., 2016). Nine studies (15.8%) evaluated the performance of SVM using other metrics. In 37/73 studies that compared SVM with other machine learning models, SVM achieved the highest performance. The average accuracy of the 33 studies that used accuracy in model evaluation was 89.8 (8.0) %, with a lowest and highest per-study accuracy of 70.0% (Ali et al., 2019) and 100.0% (Nuvoli et al., 2019; Hariharan et al., 2014), respectively.

***Neural networks (n = 62)***

Neural networks were applied to voice recordings (n = 24), movement or gait data (n = 11), MRI (n = 8), handwriting (n = 6), SPECT (n = 3), PET (n = 2). Eight studies used a combination of data types (Supplementary Figure 1). In 12 studies, multiple neural networks were tested.

In the 35 studies that applied neural networks to the diagnosis of PD without experimenting with other machine learning models, 28 studies used classification accuracy as the measure of model performance and achieved an average accuracy of 89.7 (8.5) %. The highest accuracy was 100% (Ali et al., 2019) and the lowest accuracy was 62.1% (Prince and de Vos, 2018). Eight out of 35 studies used metrics other than accuracy (Supplementary Table 1). In 14 out of 27 studies that compared the performance of neural networks with other classifiers, the highest per-study accuracy was achieved by neural networks (92.9 (5.5) %), with a lowest accuracy of 82.9% (Shinde et al., 2019) and a highest accuracy of 100% (Hariharan et al., 2014).

***Ensemble learning (n = 57)***

Ensemble learning models have been trained with movement or gait data (n = 18), voice recordings (n = 14), handwriting (n = 6), MRI (n = 2), CSF (n = 2) and SPECT (n = 1). Four studies used other types of data, such as SNPs (Cibulka et al., 2019), cardiac scintigraphy (Nuvoli et al., 2019), Patient Questionnaire of MDS-UPDRS (Prashanth and Dutta Roy, 2018) and EEG (Vanegas et al., 2018). In 10 studies, ensemble learning was applied to combinations of more than one data type (Supplementary Figure 1).

In 13 studies, ensemble learning was the only method used. Nine out of 13 studies used accuracy to measure the performance of ensemble learning models, with an average accuracy of 89.2 (6.5) %. In these studies, the lowest accuracy was 76.2% (Rubbert et al., 2019) and the highest accuracy was 96.93% (Ozcift, 2012). Ensemble learning achieved the highest per-study performance in 23 out of the 44 studies that compared ensemble learning with other machine learning models. Among 19 out of the 23 studies that used accuracy in model evaluation, the lowest and highest accuracy was 76.44% (Ali et al., 2019) and 98.61% (Dinov et al., 2016), and the average accuracy was 90.4 (6.6) %.

***Nearest Neighbors (n = 33)***

Nearest neighbors was applied to analyze voice recordings (n = 12), movement or gait data (n = 10), handwriting (n = 5), SPECT data (n = 2). In 4 studies, more than one data type was used (Supplementary Figure 1).

An average accuracy of 86.5 (9.5) % was observed in the 3 studies that have examined the performance of k-nearest neighbors algorithms without testing other machine learning models (Moharkan et al., 2017; Klomsae et al., 2018; Cuzzolin et al., 2017). In the 3 studies, the highest accuracy was 96.43% (Klomsae et al., 2018) and the lowest accuracy was 77.5% (Moharkan et al., 2017). Compared with other machine learning models, nearest neighbors displayed the highest accuracy in 5 out of 30 studies, leading to an average accuracy of 93.1 (6.3) %, with a lowest accuracy of 82.2% (Mabrouk et al., 2019) and a highest accuracy of 97.89% (Cai et al., 2018).

***Regression (n = 31)***

Regression has been applied to voice recordings (n = 10), movement or gait data (n = 7), SPECT (n = 4), MRI data (n = 2), CSF (n = 2), handwriting (n = 1), EEG data (n = 1) and Patient Questionnaire of MDS-UPDRS (n = 1). In 3 studies, more than one data type was used (Supplementary Figure 1).

Four out of 31 studies did not compare the performance of regression methods with other machine learning models (Du et al., 2017; Liu et al., 2016; Tagare et al., 2017; Trezzi et al., 2017). Of these studies, 3 have reported the AUC as the performance metric (Du et al., 2017; Liu et al., 2016; Trezzi et al., 2017), leading to an average of 0.93 (0.08) and a range between 0.833 (Trezzi et al., 2017) and 0.997 (Liu et al., 2016). Among the 27 studies that compared regression with other machine learning models, regression achieved the highest performance in 3 studies, yielding an accuracy of 70% (Ali et al., 2019) or 76.03% (Celik and Omurca, 2019), and an averaged AUC of 0.835 (Stoessel et al., 2018).

***Decision tree (n = 27)***

In the 27 studies, 11 applied decision tree to movement or gait data, 9 to voice recordings, 2 to handwriting, 1 to EEG and 1 to serum samples. In 3 studies, decision tree was applied to analyze combinations of data of more than one type (Supplementary Figure 1).

In the only study that used decision tree to classify HC and PD without evaluating other models, a cross validation score of 0.86 in male subjects and 0.63 in female subjects was achieved (Váradi et al., 2019). Decision tree achieved the highest per-study accuracy of 96.8% in 1 out of 26 studies that tested multiple machine learning models (Félix et al., 2019). However, the same accuracy was achieved with SVM or nearest neighbors.

***Naïve Bayes (n = 26)***

Naïve Bayes was applied to movement or gait data (n = 13), handwriting (n = 5), voice recordings (n = 4), MRI (n = 1), SPECT (n = 1). In 2 studies, combinations of different data types were used (Supplementary Figure 1). In all 26 studies, naïve Bayes was compared with other methods and achieved the highest performance in 2 studies, reaching an accuracy of 78.9% (Pereira et al., 2015) or 81.45% (Butt et al., 2018).

***Discriminant analysis (n = 12)***

Discriminant analysis has been applied to movement or gait data (n = 4), MRI (n = 3), voice recordings (n = 3), SPECT (n = 1) and handwriting (n = 1; Supplementary Figure 1). In the 4 studies that did not compare discriminant analysis with other models, the average accuracy was 74.1 (8.3) %, with a lowest accuracy of 64.1% (Martinez et al., 2018) and a highest accuracy of 81.9% (Adeli et al., 2016). In all 8 studies that compared discriminant analysis with other methods, the highest reported performance was achieved by other machine learning models.

***Other models (n = 24)***

A small percentage of studies (n = 24, 11.5%) used machine learning models that did not belong to any given categories (Supplementary Table 1). These models have been applied to voice recordings (n = 10), handwriting (n = 4), movement or gait data (n = 3), MRI data (n = 2), SPECT (n = 1) and TCS image (n = 1). Three studies used more than one type of data (Supplementary Figure 1). In the 24 studies, 10 used only one model, and 14 assessed performance of multiple models.

In the 10 studies that used one machine learning model, 9 measured model performance with accuracy and the average accuracy was 92.8 (6.9) %. The lowest accuracy was 79.6% (Khan et al., 2018) and the highest accuracy was 100.0% (Abiyev and Abizade, 2016; Dastjerd et al., 2019). Models that do not belong to any given categories were associated with the highest accuracy in 2 out of 14 studies that assessed multiple models (Kuresan et al., 2019; Yang et al., 2014). The highest accuracy in the two studies was 95.16% and 91.8%, respectively.

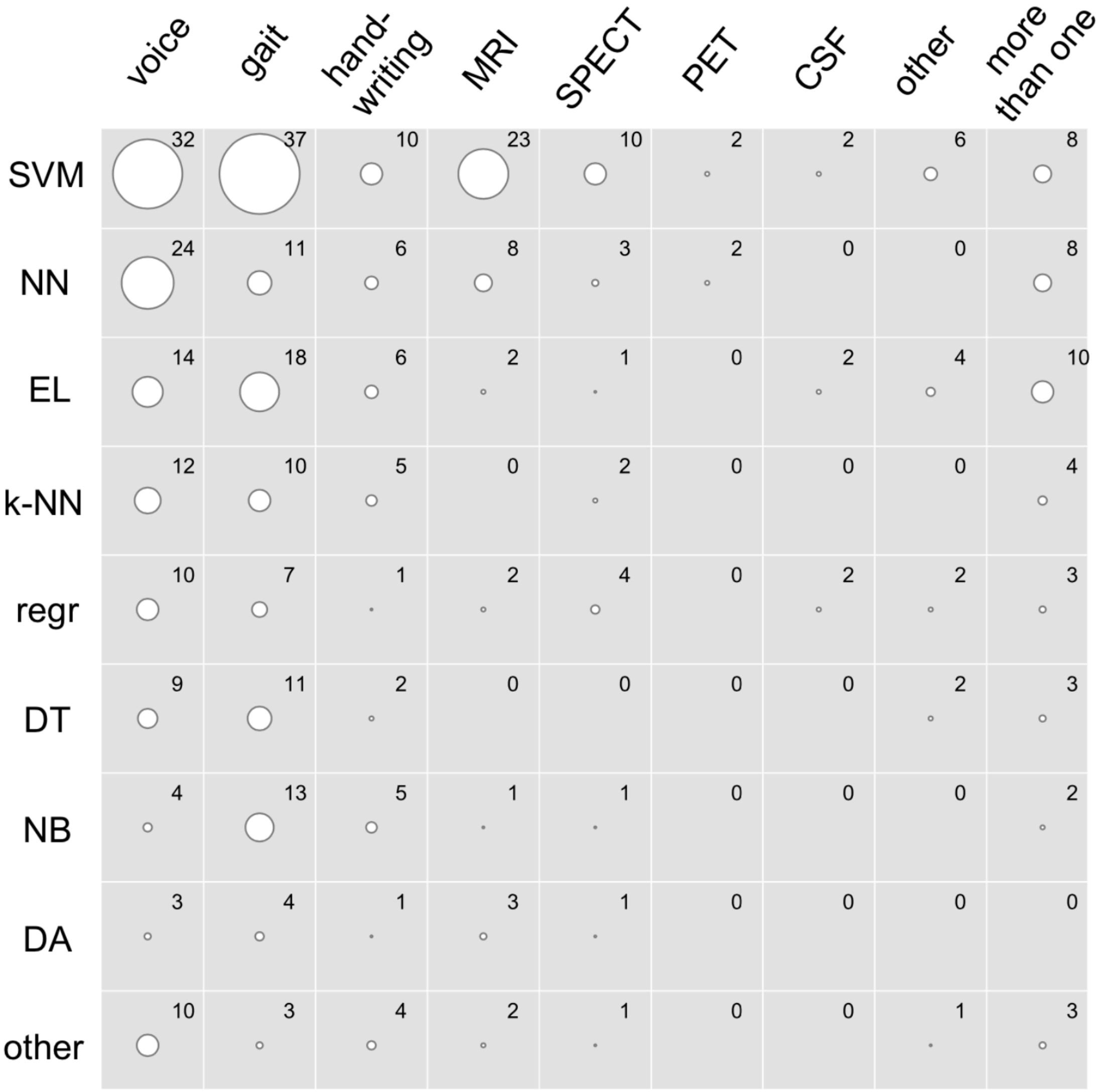

**Supplementary Figure 1.** Bubble chart showing the number of studies that applied each machine learning model, arranged according to the type of data analyzed. Studies using multiple machine learning models have been counted more than once. MRI: magnetic resonance imaging; SPECT: single-photon emission computed tomography; PET: positron emission tomography; CSF: cerebrospinal fluid; SVM: support vector machine; NN: neural network; EL: ensemble learning; k-NN: nearest neighbor; regr: regression; DT: decision tree; NB: naïve Bayes; DA: discriminant analysis; other: data/models that do not belong to any of the given categories.

| Objectives | Type of diagnosis | Source of data | Type of data | Number of subjects (n) | Machine learning method(s) | Outcomes | Year | Reference |
|---|---|---|---|---|---|---|---|---|
| Classification of PD from HC | Diagnosis | HandPD | handwritten patterns | 92; 18 HC + 74 PD | LDA, KNN, Gaussian naïve Bayes, decision tree, Chi2 with Adaboost with 5- or 4-fold stratified cross validation | Chi-2 with Adaboost: accuracy = 76.44% sensitivity = 70.94% specificity = 81.94% | 2019 | Ali et al |
| Classification of PD (PD + SWEDD) from HC | Diagnosis | PPMI database | more than one | 388; 194 HC + 168 PD + 26 SWEDD | ensemble method of several SVM with linear kernel with leave-one-out cross validation | accuracy = 94.38% | 2018 | Castillo-Barnes et al |
| Classification of PD from HC | Diagnosis | PPMI database | more than one | 586; 184 HC + 402 PD | MLP, BayesNet, random forest, boosted logistic regression with a train-test ratio of 70:30 | boosted logistic regression: accuracy = 97.159% AUC curve = 98.9% | 2016 | Challa et al |
| Classification of tPD from rET | Differential diagnosis | collected from participants | more than one | 30; 15 tPD + 15rET | multi-kernel SVM with leave-one-out cross validation | accuracy = 100% | 2014 | Cherubini et al |
| Classfication of PD, HC and atypical PD | Diagnosis, differential diagnosis and subtyping | PPMI database and SNUH cohort | SPECT imaging data | PPMI: 701; 193 HC + 431 PD + 77 SWEDD snuh: 82 PD | CNN with train-validation ratio of 90:10 | PPMI: accuracy = 96.0% sensitivity = 94.2% specificity = 100% SNUH: accuracy = 98.8% sensitivity = 98.6% specificity = 100% | 2017 | Choi et al |
| Classification of PD from HC | Diagnosis | collected from participants | other | 270; 120 HC + 150 PD | random forest | classification error = 49.6% (rs11240569) classification error = 44.8% (rs708727) classification error = 49.3% (rs823156) | 2019 | Cibulka et al |
| Classification of PD from HC | Diagnosis | HandPD | handwritten patterns | 92; 18 HC + 74 PD | naïve Bayes, OPF, SVM with cross-validation | SVM-RBF accuracy = 85.54% | 2018 | de Souza et al |
| Classification of PD from HC | Diagnosis | PPMI database | more than one | 1194; 816 HC + 378 PD | BoostPark | accuracy = 0.901 AUC-ROC = 0.977 AUC-PR = 0.947 F1-score = 0.851 | 2017 | Dhami et al |
| Classification of PD and HC, and PD + SWEDD and HC | Diagnosis | PPMI database | more than one | 430; 127 HC + 263 PD + 40 SWEDD | AdaBoost, SVM, naïve Bayes, decision tree, KNN, K-Means with 5-fold cross validation | PD vs HC (adaboost): accuracy = 0.98954704 sensitivity = 0.97831978 specificity = 0.99796748 PPV = 0.99723757 NPV = 0.98396794 LOR = 10.0058805 PD + SWEDD vs HC (adaboost): accuracy = 0.9825784 sensitivity = 0.97560976 specificity = 0.98780488 PPV = 0.98360656 NPV = 0.98181818 LOR = 8.08332861 | 2016 | Dinov et al |
| Classification of PD from HC | Diagnosis | collected from participants | CSF | cohort 1: 160; 80 HC + 80 PD<br><br>cohort 2: 60; 30 HC + 30 PD | Elastic Net and gradient boosted regression with 10-fold cross validation | ensemble of 60 decision trees identified with gradient boosted model:<br>sensitivity = 85%<br>specificity = 75%<br>PPV = 77%<br>NPV = 83%<br>AUC = 0.77 | 2018 | Dos Santos et al |
| Classification of PD from HC | Diagnosis | collected from participants | handwritten patterns | 75; 38 HC + 37 PD | SVM-RBF with stratified 10-fold cross-validation | accuracy = 88.13%<br>sensitivity = 89.47%<br>specificity = 91.89% | 2015 | Drotar et al |
| Classification of PD from HC | Diagnosis | collected from participants | handwritten patterns | 75; 38 HC + 37 PD | KNN, ensemble AdaBoost, SVM with | SVM:<br>accuracy = 81.3%<br>sensitivity = 87.4%<br>specificity = 80.9% | 2016 | Drotar et al |
| Classification of IPD, VaP and HC | Differential diagnosis | collected from participants | more than one | 45; 15 HC + 15 IPD + 15 VaP | MLP, DBN with 10-fold cross validation | IPD + VaP vs HC with MLP:<br>accuracy = 95.68%<br>specificity = 98.08%<br>sensitivity = 92.44%<br><br>VaP vs. IPD with DBN:<br>accuracy = 75.33%<br>specificity = 72.31%<br>sensitivity = 79.18% | 2018 | Fernandes et al |
| Classification of PD from HC | Diagnosis | collected from participants | more than one | 75; 15 HC + 60 PD<br>blood: 75; 15 HC + 60 PD<br>FDOPA PET: 58; 14 HC + 44 PD<br>FDG PET: 67; 16 HC + 51 PD | SVM-linear, random forest with leave-one-out cross validation | SVM AUC for FDOPA + metabolomics: 0.98<br>SVM AUC for FDG + metabolomics: 0.91 | 2019 | Glaab et al |
| Classification of PD, HC and SWEDD | Diagnosis and subtyping | PPMI database | more than one | 666; 415 HC + 189 PD + 62 SWEDD | EPNN, PNN, SVM, KNN, classification tree with train-test ratio of 90:10 | EPNN:<br>PD vs SWEDD vs HC accuracy = 92.5%<br>PD vs HC accuracy = 98.6%<br>SWEDD vs HC accuracy = 92.0%<br>PD vs SWEDD accuracy = 95.3% | 2015 | Hirschauer et al |
| Classification of PD from HC and assess the severity of PD | Diagnosis | Picture Archiving and Communication System (PACS) | SPECT imaging data | 202; 6 HC + 102 mild PD + 94 severe PD | linear regression, SVM-RBF with a train-test ratio of 50:50 | SVM-RBF:<br>sensitivity = 0.828<br>specificity = 1.000<br>PPV = 0.837<br>NPV = 0.667<br>accuracy = 0.832<br>AUC = 0.845<br>Kappa = 0.680 | 2019 | Hsu et al |
| Classification of PD from VP | Differential diagnosis | collected from participants | SPECT imaging data | 244; 164 PD + 80 VP | logistic regression, LDA, SVM with 10-fold cross-validation | SVM:<br>accuracy = 0.904<br>sensitivity = 0.954<br>specificity = 0.801<br>AUC = 0.954 | 2014 | Huertas-Fernández et al |
| Classification of PD from HC | Diagnosis | collected from participants | SPECT imaging data | 208; 108 HC + 100 PD | SVM, KNN, NM with 3-fold cross validation | SVM:<br>sensitivity = 89.02%<br>specificity = 93.21%<br>AUC = 0.9681 | 2012 | Illan et al |
| Classification of PD from HC | Diagnosis | collected from participants | handwritten patterns | 72; 15 HC + 57 PD | CNN with 10-fold cross validation or leave-one-out cross validation | accuracy = 88.89% | 2018 | Khatamino et al |
| Classification of PD from HC | Diagnosis | collected from participants | other | 10; 5 HC + 5 PD | SVM with leave-one-subject-out cross validation | sensitivity = 0.90<br>specificity = 0.90 | 2013 | Kugler et al |
| Classification of PD from HC | Diagnosis | UCI machine learning repository | handwritten patterns | 72; 15 HC + 57 PD | SVM-linear, SVM-RBF, KNN with leave-one-subject- | SVM-linear:<br>accuracy = 97.52%<br>MCC = 0.9150<br>F-score = 0.9828 | 2019 | Kurt et al |
| | | | | | out cross validation | | | |
| Classification of PD from HC | Diagnosis | collected postmortem | CSF | 105; 57 HC + 48 PD | SVM with 10-fold cross validation | sensitivity = 65%<br>specificity = 79%<br>AUC = 0.79 | 2013 | LeWitt et al |
| Classification of PD from HC | Diagnosis | collected from participants | CSF | 78; 42 HC + 36 PD | random forest and extreme gradient tree boosting with 10-fold cross validation | extreme gradient tree boosting:<br>specificity = 78.6%<br>sensitivity = 83.3%<br>AUC = 83.9% | 2018 | Maass et al |
| Classification of PD from HC or NPH | Diagnosis and differential diagnosis | collected from participants | CSF | 157; 68 HC + 82 PD + 7 NPH | SVM with 10-fold cross validation or leave-one-out cross validation | cohort 1, PD vs HC:<br>AUC = 0.76<br><br>cohort 2, PD vs HC:<br>AUC = 0.78<br><br>cohort 3, PD vs HC:<br>AUC = 0.31<br><br>cohort 4, PD vs NPH:<br>AUC = 0.88 | 2020 | Maass et al |
| Classification of PD from HC | Diagnosis | PPMI database | more than one | 550; 157 HC + 342 PD + 51 SWEDD | SVM, random forest, MLP, logistic regression, KNN with nested cross-validation | motor features, SVM:<br>accuracy = 78.4%<br>AUC = 84.7%<br><br>non-motor features, KNN:<br>accuracy = 82.2%<br>AUC = 88% | 2018 | Mabrouk et al |
| Classification of PD from HC | Diagnosis | PPMI database | SPECT imaging data | 642; 194 HC + 448 PD | CNN (LENET53D, ALEXNET3D) with 10-fold stratified cross-validation | ALEXNET3D:<br>accuracy = 94.1%<br>AUC = 0.984 | 2018 | Martinez-Murcia et al |
| Classification of PD from HC | Diagnosis | collected from participants | handwritten patterns | 75; 10 HC + 65 PD | MLP, non-linear SVM, random forest, logistic regression with stratified 10-fold cross-validation | MLP:<br>accuracy = 84%<br>sensitivity = 75.7%<br>specificity = 88.9%<br>weighted Kappa = 0.65<br>AUC = 0.86 | 2015 | Memedi et al |
| Classification of PD from HC | Diagnosis | Parkinson's Disease Handwriting Database (PaHaW) | handwritten patterns | 69; 36 HC + 33 PD | random forest with stratified 7-fold cross-validation | accuracy = 89.81%<br>sensitivity = 88.63%<br>specificity = 90.87%<br>MCC = 0.8039 | 2018 | Mucha et al |
| Classification of PD, MSA, PSP, CBS and HC | Differential diagnosis | collected from participants | SPECT imaging data | 578; 208 HC + 280 PD + 21 MSA + 41 PSP + 28 CBS | SVM with 5-fold cross-validation | accuracy = 58.4 - 92.9% | 2019 | Nicastro et al |
| Classification of PD from HC | Diagnosis | collected from participants | handwritten patterns | 30; 15 HC + 15 PD | KNN, decision tree, random forest, SVM, AdaBoost with 3-fold cross validation | random forest accuracy = 0.91 | 2018 | Nomm et al |
| Classification of HC, AD and PD | Diagnosis and differential diagnosis | the authors' institutional OCT database | other | 75; 27 HC + 28 PD + 20 AD | SVM-RBF with 2-, 5- and 10-fold cross validation | accuracy = 87.7%<br>HC sensitivity = 96.2%<br>HC specificity = 88.2%<br>PD sensitivity = 87.0%<br>PD specificity = 100.0% | 2019 | Nunes et al |
| Classification of idiopathic PD, atypical Parkinsonian and ET | Differential diagnosis | collected from participants | other | 85; 50 idiopathic PD + 26 atypical PD + 9 ET | SVM, random forest with leave-one-out cross validation | SVM accuracy = 100%<br>random forest accuracy = 98.5% | 2019 | Nuvoli et al |
| Classification of PD from HC | Diagnosis | PPMI database | SPECT imaging data | 654; 209 HC + 445 PD | SVM-linear with leave-one-out cross validation | accuracy = 97.86%<br>sensitivity = 97.75%<br>specificity = 98.09% | 2015 | Oliveira and Castelo-Branco |
| Classification of PD from HC | Diagnosis | PPMI database | SPECT imaging data | 652; 209 HC + 443 PD | SVM-linear, KNN, logistic regression with leave-one-out cross validation | SVM-linear: accuracy = 97.9% sensitivity = 98.0% specificity = 97.6% | 2017 | Oliveira et al |
| Classification of PD and non-PD (ET, drug-induced Parkinsonism) | Differential diagnosis | collected from participants | SPECT imaging data | 90; 56 PD + 34 non-PD | SVM-RBF with leave-one-out or 5-fold cross validation | accuracy = 95.6% | 2014 | Palumbo et al |
| Classification of PD from HC | Diagnosis | collected from participants | handwritten patterns | 55; 18 HC + 37 PD | naïve Bayes, OPF, SVM-rbf with 10-fold cross validation | naïve Bayes accuracy = 78.9% | 2015 | Pereira et al |
| Classification of PD from HC | Diagnosis | HandPD | handwritten patterns | 92; 18 HC + 74 PD | naïve Bayes, OPF, SVM-RBF with cross-validation | SVM-RBF recognition rate (sensitivity) = 66.72% | 2016 | Pereira et al |
| Classification of PD from HC | Diagnosis | extended HandPD dataset with signals extracted from a smart pen | handwritten patterns | 35; 21 HC + 14 PD | CNN with cross validation with a train:test ratio of 75:25 or 50:50 | accuracy = 87.14% | 2016 | Pereira et al |
| Classification of PD from HC | Diagnosis | HandPD | handwritten patterns | 92; 18 HC + 74 PD | CNN, OPF, SVM, naïve Bayes with train-test split = 50:50 | CNN-Cifar10 accuracy = 99.30% early stage accuracy with CNN-ImageNet = 96.35% or 94.01% for Exam 3 or Exam 4 | 2018 | Pereira et al |
| Classification of PD from HC | Diagnosis | UCI machine learning repository | more than one | dataset 1: 40; 20 HC + 20 PD<br><br>dataset 2: 77; 15 HC + 62 PD | random forest, KNN, SVM-RBF, ensemble method with 5-fold cross validation | ensemble method: accuracy = 95.89% specificity = 100% sensitivity = 91.43% | 2019 | Pham et al |
| Classification of PD from HC | Diagnosis | PPMI database | more than one | 618; 195 HC + 423 PD | SVM-linear, SVM-RBF, classification tree with a train-test ratio of 70:30 | SVM-RBF, test set: accuracy = 85.48% sensitivity = 90.55% specificity = 74.58% AUC = 88.22% | 2014 | Prashanth et al |
| Classification of PD from HC | Diagnosis and subtyping | PPMI database | SPECT imaging data | 715; 208 HC + 427 PD + 80 SWEDD | SVM, naïve Bayes, boosted trees, random forest with 10-fold cross validation | SVM: accuracy = 97.29% sensitivity = 97.37% specificity = 97.18% AUC = 99.26 | 2016 | Prashanth et al |
| Classification of PD from HC | Diagnosis | PPMI database | more than one | 584; 183 HC + 401 PD | naïve Bayes, SVM-RBF, boosted trees, random forest with 10-fold cross validation | SVM: accuracy = 96.40% sensitivity = 97.03% specificity = 95.01% AUC = 98.88% | 2016 | Prashanth et al |
| Classification of PD from HC | Diagnosis | PPMI database | other | 626; 180 HC + 446 PD | logistic regression, random forests, boosted trees, SVM with cross validation | accuracy > 95% AUC > 95%<br><br>random forests: accuracy = 96.20%-97.14% (95% CI) | 2018 | Prashanth and Roy |
| Classification of PD from HC | Diagnosis | mPower database | more than one | 133 out of 1,513 with complete source data; 46 HC + 87 PD | logistic regression, random forests, DNN, CNN, Classifier Ensemble, Multi-Source Ensemble learning with stratified 10-fold cross validation | ensemble learning: accuracy = 82.0% F1-score = 87.1% | 2019 | Prince et al |
| Classification of PD from HC | Diagnosis | HandPD | handwritten patterns | 35; 21 HC + 14 PD | Bidirectional Gated Recurrent Units with a train-validation-test ratio of 40:10:50 or 65:10:25 | the Spiral dataset:<br>accuracy = 89.48%<br>precision = 0.848<br>recall = 0.955<br>F1-score = 0.897<br><br>the Meander dataset:<br>accuracy = 92.24%<br>precision = 0.952<br>recall = 0.883<br>F1-score = 0.924 | 2019 | Ribeiro |
| Classification of PD from HC | Diagnosis | collected from participants | handwritten patterns | 130; 39 elderly HC + 40 young HC + 39 PD + 6 PD (validation set) + 6 HC (validation set) | KNN, SVM-Gaussian, random forest with leave-one-out cross validation | SVM for PD vs young HC:<br>accuracy = 94.0%<br>sensitivity = 0.94<br>specificity = 0.94<br>F1-score = 0.94<br><br>SVM for PD vs elderly HC:<br>accuracy = 89.3%<br>sensitivity = 0.89<br>specificity = 0.89<br>F1-score = 0.89<br><br>random forest for validation set:<br>accuracy = 83.3%<br>sensitivity = 0.92<br>specificity = 0.93<br>F1-score = 0.92 | 2019 | Rios-Urrego et al |
| Classification of IPD from non-IPD | Differential diagnosis | collected from participants | PET imaging | 87; 39 IPD + 48 non-IPD (24 MSA + 24 PSP) | SVM with leave-one-out cross validation | accuracy = 78.16%<br>sensitivity = 69.29%<br>specificity = 85.42% | 2015 | Segovia et al |
| Classification of PD from HC | Diagnosis | dataset from “Virgen de la Victoria” hospital | SPECT imaging data | 189; 94 HC + 95 PD | SVM with 10-fold cross validation | accuracy = 94.25%<br>sensitivity = 91.26%<br>specificity = 96.17% | 2019 | Segovia et al |
| Classification of PD from HC | Diagnosis | collected from participants | other | 486; 233 HC + 205 PD + 48 NDD | SVM-linear with leave-batch-out cross validation | validation AUC = 0.79<br>test AUC = 0.74 | 2017 | Shamir et al |
| Classification of PD from HC | Diagnosis | collected from participants | PET imaging | 350; 225 HC + 125 PD | GLS-DBN with a train-validation ratio of 80:20 | test dataset 1:<br>accuracy = 90%<br>sensitivity = 0.96<br>specificity = 0.84<br>AUC = 0.9120<br><br>test dataset 2:<br>accuracy = 86%<br>sensitivity = 0.92<br>specificity = 0.80<br>AUC = 0.8992 | 2019 | Shen et al |
| Classification of PD from HC | Diagnosis | collected from participants | other | 33; 18 HC + 15 PD | SMMKL-linear with leave-one-out cross validation | accuracy = 84.85%<br>sensitivity = 80.00%<br>specificity = 88.89%<br>YI = 68.89%<br>PPV = 85.71%<br>NPV = 84.21%<br>F1 score = 82.76% | 2018 | Shi et al |
| Classification of PD from HC | Diagnosis | collected from participants | more than one | plasma samples: 156; 76 HC + 80 PD;<br><br>CSF samples: 77; 37 HC + 40 PD | PLS, random forest with 10-fold cross validation with train-test ratio of 70:30 | PLS:<br>AUC(plasma) = 0.77<br>AUC(CSF) = 0.90 | 2018 | Stoessel et al |
| Classification of PD from HC | Diagnosis | PPMI database | SPECT imaging data | 658; 210 HC + 448 PD | logistic Lasso with 10-fold cross validation | test errors:<br>FP = 2.83%<br>FN = 3.78%<br>ERR = 3.47% | 2017 | Tagare et al |
| Classification of PD from HC | Diagnosis | PDMultiMC | handwritten patterns | 42; 21 HC + 21 PD | CNN, CNN-BLSTM with stratified 3-fold cross validation | CNN:<br>accuracy = 83.33%<br>sensitivity = 85.71%<br>specificity = 80.95%<br><br>CNN-BLSTM:<br>accuracy = 83.33%<br>sensitivity = 71.43%<br>specificity = 95.24% | 2019 | Taleb et al |
| Classification of PD from HC | Diagnosis | PPMI database and local database | SPECT imaging data | local: 304; 113 Non-PDD + 191 PD<br>PPMI: 657; 209 HC + 448 PD | SVM with stratified, nested 10-fold cross-validation | local data:<br>accuracy = 0.88 to 0.92<br><br>PPMI:<br>accuracy = 0.95 to 0.97 | 2017 | Taylor and Fenner |
| Classification of PD from HC | Diagnosis | collected from participants | CSF | 87; 43 HC + 44 PD | logistic regression | sensitivity = 0.797<br>specificity = 0.800<br>AUC = 0.833 | 2017 | Trezzi et al |
| Classification of PD from HC | Diagnosis | collected from participants | other | 38; 24 HC + 14 PD | SVM-RFE with repeated leave-one-out bootstrap validation | accuracy = 89.6% | 2013 | Tseng et al |
| Classification of MSA and PD | Differential diagnosis | collected from participants | more than one | 85; 25 HC + 30 PD + 30 MSA-P | NN | AUC = 0.775 | 2019 | Tsuda et al |
| Classification of PD from HC | Diagnosis | collected from participants | other | 59; 30 HC + 29 PD | logistic regression, decision tree, extra tree | extra tree AUC = 0.99422 | 2018 | Vanegas et al |
| Classification of PD from HC | Diagnosis | commercially sourced | other | 30; 15 HC + 15 PD | decision tree | cross validation score = 0.86 (male)<br>cross validation score = 0.63 (female) | 2019 | Varadi et al |
| Classification of PD from HC | Diagnosis | collected from participants | more than one | 84; 40 HC + 44 PD | CNN with train-validation-test ratio of 80:10:10 | accuracy = 97.6%<br>AUC = 0.988 | 2018 | Vasquez-Correa et al |
| Classification of PD and Parkinsonism | Differential diagnosis | The NTUA Parkinson Dataset | more than one | 78; 55 PD + 23 Parkinsonism | MTL with DNN | accuracy = 0.91<br>precision = 0.83<br>sensitivity = 1.0<br>specificity = 0.83<br>AUC = 0.92 | 2018 | Vlachostergiou et al |
| Classification of PD from HC | Diagnosis | PPMI database | more than one | 534; 165 HC + 369 PD | pGTL with 10-fold cross validation | accuracy = 97.4% | 2017 | Wang et al |
| Classification of PD from HC | Diagnosis | PPMI database | SPECT imaging data | 645; 207 HC + 438 PD | CNN with train-validation-test ratio of 60:20:20 | accuracy = 0.972<br>sensitivity = 0.983<br>specificity = 0.962 | 2019 | Wenzel et al |
| Classification of PD from HC | Diagnosis | collected from participants | PET imaging | cohort 1: 182; 91 HC + 91 PD<br><br>cohort 2: 48; 26 HC + 22 PD | SVM-linear, SVM-sigmoid, SVM-RBF with 5-fold cross validation | cohort 1:<br>accuracy = 91.26%<br>sensitivity = 89.43%<br>specificity = 93.27%<br><br>cohort 2:<br>accuracy = 90.18%<br>sensitivity = 82.05%<br>specificity = 92.05% | 2019 | Wu et al |
| Classification of PD, MSA and PSP | Differential diagnosis | collected from participants | PET imaging | 920; 502 PD + 239 MSA + 179 PSP | 3D residual CNN with 6-fold cross validation | classification of PD:<br>sensitivity = 97.7%<br>specificity = 94.1%<br>PPV = 95.5%<br>NPV = 97.0%<br><br>classification of MSA:<br>sensitivity = 96.8%<br>specificity = 99.5%<br>PPV = 98.7%<br>NPV = 98.7%<br><br>classification of PSP:<br>sensitivity = 83.3%<br>specificity = 98.3%<br>PPV = 90.0%<br>NPV = 97.8% | 2019 | Zhao et al |

**Supplementary Table 1**. Studies that applied machine learning models to handwriting, SPECT, PET, CSF, other data types and combinations of data to diagnose PD (n = 67).

| ML model | Number of studies | Type of data | Only model tested | Performance | Compared with other models | Performance |
|---|---|---|---|---|---|---|
| **SVM** | 130 | • Movement (37)<br>• Voice (32)<br>• MRI (23)<br>• SPECT (10)<br>• Handwriting (10)<br>• CSF (2)<br>• PET (2)<br>• Other (6)<br>• Combination (8) | In 57 studies | accuracy:<br>90.52 (7.65) %<br><br>min: 74%<br>max: 100% | In 73 studies; SVM reached highest performance in 37 studies (50.7%) | accuracy:<br>89.78 (8.01) %<br><br>min: 70.0%<br>max: 100.0% |
| **NN** | 62 | • Voice (24)<br>• Movement (11)<br>• MRI (8)<br>• Handwriting (6)<br>• SPECT (3)<br>• PET (2)<br>• Combination (8) | In 35 studies | accuracy:<br>89.7 (8.5) %<br><br>min: 62.1%<br>max: 100% | In 27 studies; NN reached highest performance in 14 studies (51.9%) | Accuracy: 92.9 (5.5) %<br><br>min: 82.9%<br>max: 100% |
| **EL** | 57 | • Movement (18)<br>• Voice (14)<br>• Handwriting (6)<br>• MRI (2)<br>• CSF (2)<br>• SPECT (1)<br>• Other (4)<br>• Combination (10) | In 13 studies | accuracy:<br>89.16 (6.47) %<br><br>min: 76.20%<br>max: 96.93% | In 44 studies; EL reached highest performance in 23 studies (52.3%) | accuracy: 90.56 (6.73) %<br><br>min: 76.44%<br>max: 98.61% |
| **k-NN** | 33 | • Voice (12)<br>• Movement (10)<br>• Handwriting (5)<br>• SPECT (2)<br>• Combination (4) | In 3 studies | accuracy:<br>86.48 (9.50) %<br><br>min: 77.50%<br>max: 96.43% | In 30 studies, k-NN reached highest performance in 5 studies (16.7%) | accuracy: 93.10 (6.35) %<br><br>min: 82.20%<br>max: 97.89% |
| **regression** | 31 | • Voice (10)<br>• Movement (7)<br>• SPECT (4)<br>• MRI (2)<br>• CSF (2)<br>• Handwriting (1)<br>• Other (2)<br>• Combination (3) | In 4 studies | AUC:<br>0.93 (0.08)<br><br>min: 0.833<br>max: 0.997 | In 27 studies, regression reached highest performance in 3 studies (11.1%) | accuracy: 70% (Ali et al., 2019) or 76.03% (Celik and Omurca, 2019)<br><br>AUC = 0.835 (Stoessel et al., 2018) |
| **DT** | 27 | • Movement (11)<br>• Voice (9)<br>• Handwriting (2)<br>• Other (2)<br>• Combination (3) | In 1 study | Cross validation score = 0.86 (male) or 0.63 (female) | In 26 studies, DT reached highest performance in 1 study (3.8%) | accuracy: 96.8% |
| **NB** | 26 | • Movement (13)<br>• Handwriting (5)<br>• Voice (4)<br>• Other (4) | In 0 study | N/A | In 26 studies, NB reached highest performance in 2 studies (7.7 %) | accuracy: 80.18 (1.80) %<br><br>min: 78.90%<br>max: 81.45% |
| **DA** | 12 | • Movement (4)<br>• MRI (3)<br>• Voice (3)<br>• SPECT (1)<br>• Handwriting (1) | In 4 studies | accuracy:<br>74.08 (8.29) %<br><br>min: 64.10%<br>max: 81.90% | In 8 studies, DA reached highest performance in 0 study (0.0%) | N/A |
| **other** | 24 | • Voice (10)<br>• Handwriting (4)<br>• Movement (3)<br>• MRI (3)<br>• SPECT (1)<br>• Other (1)<br>• Combination (2) | In 10 studies | accuracy:<br>92.82 (6.92) %<br><br>min: 79.60%<br>max: 100.00% | In 14 studies, these models reached highest performance in 2 studies (14.3%) | accuracy: 93.48 (2.38) %<br><br>min: 91.80%<br>max: 95.16% |

**Supplementary Table 2**. Summary of machine learning models used in the included studies by category. SVM: support vector machine; NN: neural network; EL: ensemble learning; k-NN: nearest neighbor; DT: decision tree; NB: naïve Bayes; DA: discriminant analysis; other: data/models that do not belong to any of the given categories.

## References

1 R.H. Abiyev, S. Abizade, Diagnosing Parkinson's Diseases Using Fuzzy Neural System, Comput Math Methods Med 2016 (2016) 1267919-1267919.

2 A. Abos, H.C. Baggio, B. Segura, A. Campabadal, C. Uribe, D.M. Giraldo, A. Perez-Soriano, E. Muñoz, Y. Compta, C. Junque, M.J. Marti, Differentiation of multiple system atrophy from Parkinson's disease by structural connectivity derived from probabilistic tractography, Sci Rep 9(1) (2019) 16488-16488.

3 H. Abujrida, E. Agu, K. Pahlavan, Smartphone-based gait assessment to infer Parkinson's disease severity using crowdsourced data, 2017 IEEE Healthcare Innovations and Point of Care Technologies (HI-POCT), 2017, pp. 208-211.

4 W.R. Adams, High-accuracy detection of early Parkinson's Disease using multiple characteristics of finger movement while typing, PLoS One 12(11) (2017) e0188226-e0188226.

5 E. Adeli, F. Shi, L. An, C.-Y. Wee, G. Wu, T. Wang, D. Shen, Joint feature-sample selection and robust diagnosis of Parkinson's disease from MRI data, Neuroimage 141 (2016) 206-219.

6 E. Adeli, K.-H. Thung, L. An, G. Wu, F. Shi, T. Wang, D. Shen, Semi-Supervised Discriminative Classification Robust to Sample-Outliers and Feature-Noises, IEEE Trans Pattern Anal Mach Intell 41(2) (2019) 515-522.

7 A. Agarwal, S. Chandrayan, S.S. Sahu, Prediction of Parkinson's disease using speech signal with Extreme Learning Machine, 2016 International Conference on Electrical, Electronics, and Optimization Techniques (ICEEOT), 2016, pp. 3776-3779.

8 S.-A. Ahmadi, G. Vivar, J. Frei, S. Nowoshilow, S. Bardins, T. Brandt, S. Krafczyk, Towards computerized diagnosis of neurological stance disorders: data mining and machine learning of posturography and sway, J Neurol 266(Suppl 1) (2019) 108-117.

9 S. Aich, H. Kim, K. younga, K.L. Hui, A.A. Al-Absi, M. Sain, A Supervised Machine Learning Approach using Different Feature Selection Techniques on Voice Datasets for Prediction of Parkinson’s Disease, 2019 21st International Conference on Advanced Communication Technology (ICACT), 2019, pp. 1116-1121.

10 A.H. Al-Fatlawi, M.H. Jabardi, S.H. Ling, Efficient diagnosis system for Parkinson's disease using deep belief network, 2016 IEEE Congress on Evolutionary Computation (CEC), 2016, pp. 1324-1330.

11 M.N. Alam, A. Garg, T.T.K. Munia, R. Fazel-Rezai, K. Tavakolian, Vertical ground reaction force marker for Parkinson's disease, PLoS One 12(5) (2017) e0175951-e0175951.

12 H. Alaskar, A. Hussain, Prediction of Parkinson Disease Using Gait Signals, 2018 11th International Conference on Developments in eSystems Engineering (DeSE), 2018, pp. 23-26.

13 A.S. Alharthi, K.B. Ozanyan, Deep Learning for Ground Reaction Force Data Analysis: Application to Wide-Area Floor Sensing, 2019 IEEE 28th International Symposium on Industrial Electronics (ISIE), 2019, pp. 1401-1406.

14 L. Ali, S.U. Khan, M. Arshad, S. Ali, M. Anwar, A Multi-model Framework for Evaluating Type of Speech Samples having Complementary Information about Parkinson's Disease, 2019 International Conference on Electrical, Communication, and Computer Engineering (ICECCE), 2019, pp. 1-5.

15 L. Ali, C. Zhu, N.A. Golilarz, A. Javeed, M. Zhou, Y. Liu, Reliable Parkinson’s Disease Detection by Analyzing Handwritten Drawings: Construction of an Unbiased Cascaded

Learning System Based on Feature Selection and Adaptive Boosting Model, IEEE Access 7 (2019) 116480-116489.
16 L. Ali, C. Zhu, Z. Zhang, Y. Liu, Automated Detection of Parkinson’s Disease Based on Multiple Types of Sustained Phonations Using Linear Discriminant Analysis and Genetically Optimized Neural Network, IEEE Journal of Translational Engineering in Health and Medicine 7 (2019) 1-10.
17 E.J. Alqahtani, F.H. Alshamrani, H.F. Syed, S.O. Olatunji, Classification of Parkinson’s Disease Using NNge Classification Algorithm, 2018 21st Saudi Computer Society National Computer Conference (NCC), 2018, pp. 1-7.
18 N. Amoroso, M. La Rocca, A. Monaco, R. Bellotti, S. Tangaro, Complex networks reveal early MRI markers of Parkinson's disease, Med Image Anal 48 (2018) 12-24.
19 A. Anand, M.A. Haque, J.S.R. Alex, N. Venkatesan, Evaluation of Machine learning and Deep learning algorithms combined with dimentionality reduction techniques for classification of Parkinson’s Disease, 2018 IEEE International Symposium on Signal Processing and Information Technology (ISSPIT), 2018, pp. 342-347.
20 A. Andrei, A. Tăuțan, B. Ionescu, Parkinson’s Disease Detection from Gait Patterns, 2019 E-Health and Bioengineering Conference (EHB), 2019, pp. 1-4.
21 M.S. Baby, A.J. Saji, C.S. Kumar, Parkinsons disease classification using wavelet transform based feature extraction of gait data, 2017 International Conference on Circuit ,Power and Computing Technologies (ICCPCT), 2017, pp. 1-6.
22 H.C. Baggio, A. Abos, B. Segura, A. Campabadal, C. Uribe, D.M. Giraldo, A. Perez-Soriano, E. Muñoz, Y. Compta, C. Junque, M.J. Marti, Cerebellar resting-state functional connectivity in Parkinson's disease and multiple system atrophy: Characterization of abnormalities and potential for differential diagnosis at the single-patient level, Neuroimage Clin 22 (2019) 101720-101720.
23 Z.A. Bakar, D.I. Ispawi, N.F. Ibrahim, N.M. Tahir, Classification of Parkinson's disease based on Multilayer Perceptrons (MLPs) Neural Network and ANOVA as a feature extraction, 2012 IEEE 8th International Colloquium on Signal Processing and its Applications, 2012, pp. 63-67.
24 M. Banerjee, R. Chakraborty, D. Archer, D. Vaillancourt, B.C. Vemuri, DMR-CNN: A CNN Tailored For DMR Scans With Applications To PD Classification, 2019 IEEE 16th International Symposium on Biomedical Imaging (ISBI 2019), 2019, pp. 388-391.
25 A. Benba, A. Jilbab, A. Hammouch, Discriminating Between Patients With Parkinson’s and Neurological Diseases Using Cepstral Analysis, IEEE Transactions on Neural Systems and Rehabilitation Engineering 24(10) (2016) 1100-1108.
26 A. Benba, A. Jilbab, A. Hammouch, S. Sandabad, Using RASTA-PLP for discriminating between different Neurological diseases, 2016 International Conference on Electrical and Information Technologies (ICEIT), 2016, pp. 406-409.
27 H. Bernad-Elazari, T. Herman, A. Mirelman, E. Gazit, N. Giladi, J.M. Hausdorff, Objective characterization of daily living transitions in patients with Parkinson's disease using a single body-fixed sensor, J Neurol 263(8) (2016) 1544-1551.
28 S. Bhati, L.M. Velazquez, J. Villalba, N. Dehak, LSTM Siamese Network for Parkinson’s Disease Detection from Speech, 2019 IEEE Global Conference on Signal and Information Processing (GlobalSIP), 2019, pp. 1-5.
29 D. Buongiorno, I. Bortone, G.D. Cascarano, G.F. Trotta, A. Brunetti, V. Bevilacqua, A low-cost vision system based on the analysis of motor features for recognition and severity rating of Parkinson's Disease, BMC Med Inform Decis Mak 19(Suppl 9) (2019) 243-243.

30 A.H. Butt, E. Rovini, C. Dolciotti, P. Bongioanni, G. De Petris, F. Cavallo, Leap motion evaluation for assessment of upper limb motor skills in Parkinson's disease, IEEE Int Conf Rehabil Robot 2017 (2017) 116-121.
31 A.H. Butt, E. Rovini, C. Dolciotti, G. De Petris, P. Bongioanni, M.C. Carboncini, F. Cavallo, Objective and automatic classification of Parkinson disease with Leap Motion controller, Biomed Eng Online 17(1) (2018) 168-168.
32 Z. Cai, J. Gu, C. Wen, D. Zhao, C. Huang, H. Huang, C. Tong, J. Li, H. Chen, An Intelligent Parkinson's Disease Diagnostic System Based on a Chaotic Bacterial Foraging Optimization Enhanced Fuzzy KNN Approach, Comput Math Methods Med 2018 (2018) 2396952-2396952.
33 C. Caramia, D. Torricelli, M. Schmid, A. Munoz-Gonzalez, J. Gonzalez-Vargas, F. Grandas, J.L. Pons, IMU-Based Classification of Parkinson's Disease From Gait: A Sensitivity Analysis on Sensor Location and Feature Selection, IEEE J Biomed Health Inform 22(6) (2018) 1765-1774.
34 D. Castillo-Barnes, J. Ramírez, F. Segovia, F.J. Martínez-Murcia, D. Salas-Gonzalez, J.M. Górriz, Robust Ensemble Classification Methodology for I123-Ioflupane SPECT Images and Multiple Heterogeneous Biomarkers in the Diagnosis of Parkinson's Disease, Front Neuroinform 12 (2018) 53-53.
35 F. Cavallo, A. Moschetti, D. Esposito, C. Maremmani, E. Rovini, Upper limb motor pre-clinical assessment in Parkinson's disease using machine learning, Parkinsonism Relat Disord 63 (2019) 111-116.
36 E. Celik, S.I. Omurca, Improving Parkinson's Disease Diagnosis with Machine Learning Methods, 2019 Scientific Meeting on Electrical-Electronics & Biomedical Engineering and Computer Science (EBBT), 2019, pp. 1-4.
37 S. Chakraborty, S. Aich, H.-C. Kim, 3D Textural, Morphological and Statistical Analysis of Voxel of Interests in 3T MRI Scans for the Detection of Parkinson's Disease Using Artificial Neural Networks, Healthcare (Basel) 8(1) (2020) E34.
38 K.N.R. Challa, V.S. Pagolu, G. Panda, B. Majhi, An improved approach for prediction of Parkinson's disease using machine learning techniques, 2016 International Conference on Signal Processing, Communication, Power and Embedded System (SCOPES), 2016, pp. 1446-1451.
39 Y. Chen, J. Storrs, L. Tan, L.J. Mazlack, J.-H. Lee, L.J. Lu, Detecting brain structural changes as biomarker from magnetic resonance images using a local feature based SVM approach, J Neurosci Methods 221 (2014) 22-31.
40 Y. Chen, W. Yang, J. Long, Y. Zhang, J. Feng, Y. Li, B. Huang, Discriminative analysis of Parkinson's disease based on whole-brain functional connectivity, PLoS One 10(4) (2015) e0124153-e0124153.
41 A. Cherubini, M. Morelli, R. Nisticó, M. Salsone, G. Arabia, R. Vasta, A. Augimeri, M.E. Caligiuri, A. Quattrone, Magnetic resonance support vector machine discriminates between Parkinson disease and progressive supranuclear palsy, Mov Disord 29(2) (2014) 266-269.
42 A. Cherubini, R. Nisticó, F. Novellino, M. Salsone, S. Nigro, G. Donzuso, A. Quattrone, Magnetic resonance support vector machine discriminates essential tremor with rest tremor from tremor-dominant Parkinson disease, Mov Disord 29(9) (2014) 1216-1219.
43 H. Choi, S. Ha, H.J. Im, S.H. Paek, D.S. Lee, Refining diagnosis of Parkinson's disease with deep learning-based interpretation of dopamine transporter imaging, Neuroimage Clin 16 (2017) 586-594.
44 M. Cibulka, M. Brodnanova, M. Grendar, M. Grofik, E. Kurca, I. Pilchova, O. Osina, Z. Tatarkova, D. Dobrota, M. Kolisek, SNPs rs11240569, rs708727, and rs823156 in SLC41A1

Do Not Discriminate Between Slovak Patients with Idiopathic Parkinson's Disease and Healthy Controls: Statistics and Machine-Learning Evidence, Int J Mol Sci 20(19) (2019) 4688.
45 O. Cigdem, H. Demirel, D. Unay, The Performance of Local-Learning Based Clustering Feature Selection Method on the Diagnosis of Parkinson’s Disease Using Structural MRI, 2019 IEEE International Conference on Systems, Man and Cybernetics (SMC), 2019, pp. 1286-1291.
46 S. Çimen, B. Bolat, Diagnosis of Parkinson's disease by using ANN, 2016 International Conference on Global Trends in Signal Processing, Information Computing and Communication (ICGTSPICC), 2016, pp. 119-121.
47 D.J. Cook, M. Schmitter-Edgecombe, P. Dawadi, Analyzing Activity Behavior and Movement in a Naturalistic Environment Using Smart Home Techniques, IEEE J Biomed Health Inform 19(6) (2015) 1882-1892.
48 F. Cuzzolin, M. Sapienza, P. Esser, S. Saha, M.M. Franssen, J. Collett, H. Dawes, Metric learning for Parkinsonian identification from IMU gait measurements, Gait Posture 54 (2017) 127-132.
49 S. Dash, R. Thulasiram, P. Thulasiraman, An Enhanced Chaos-Based Firefly Model for Parkinson's Disease Diagnosis and Classification, 2017 International Conference on Information Technology (ICIT), 2017, pp. 159-164.
50 N.K. Dastjerd, O.C. Sert, T. Ozyer, R. Alhajj, Fuzzy Classification Methods Based Diagnosis of Parkinson's disease from Speech Test Cases, Curr Aging Sci 12(2) (2019) 100-120.
51 J.W.M. de Souza, S.S.A. Alves, E.d.S. Rebouças, J.S. Almeida, P.P. Rebouças Filho, A New Approach to Diagnose Parkinson's Disease Using a Structural Cooccurrence Matrix for a Similarity Analysis, Comput Intell Neurosci 2018 (2018) 7613282-7613282.
52 D.S. Dhami, A. Soni, D. Page, S. Natarajan, Identifying Parkinson's Patients: A Functional Gradient Boosting Approach, Artif Intell Med Conf Artif Intell Med (2005-) 10259 (2017) 332-337.
53 A. Dinesh, J. He, Using machine learning to diagnose Parkinson's disease from voice recordings, 2017 IEEE MIT Undergraduate Research Technology Conference (URTC), 2017, pp. 1-4.
54 I.D. Dinov, B. Heavner, M. Tang, G. Glusman, K. Chard, M. Darcy, R. Madduri, J. Pa, C. Spino, C. Kesselman, I. Foster, E.W. Deutsch, N.D. Price, J.D. Van Horn, J. Ames, K. Clark, L. Hood, B.M. Hampstead, W. Dauer, A.W. Toga, Predictive Big Data Analytics: A Study of Parkinson's Disease Using Large, Complex, Heterogeneous, Incongruent, Multi-Source and Incomplete Observations, PLoS One 11(8) (2016) e0157077-e0157077.
55 M. Djurić-Jovičić, M. Belić, I. Stanković, S. Radovanović, V.S. Kostić, Selection of gait parameters for differential diagnostics of patients with de novo Parkinson's disease, Neurol Res 39(10) (2017) 853-861.
56 M.C.T. Dos Santos, D. Scheller, C. Schulte, I.R. Mesa, P. Colman, S.R. Bujac, R. Bell, C. Berteau, L.T. Perez, I. Lachmann, D. Berg, W. Maetzler, A. Nogueira da Costa, Evaluation of cerebrospinal fluid proteins as potential biomarkers for early stage Parkinson's disease diagnosis, PLoS One 13(11) (2018) e0206536-e0206536.
57 B. Dror, E. Yanai, A. Frid, N. Peleg, N. Goldenthal, I. Schlesinger, H. Hel-Or, S. Raz, Automatic assessment of Parkinson's Disease from natural hands movements using 3D depth sensor, 2014 IEEE 28th Convention of Electrical & Electronics Engineers in Israel (IEEEI), 2014, pp. 1-5.

58 P. Drotár, J. Mekyska, I. Rektorová, L. Masarová, Z. Smékal, M. Faundez-Zanuy, Analysis of in-air movement in handwriting: A novel marker for Parkinson's disease, Comput Methods Programs Biomed 117(3) (2014) 405-411.
59 P. Drotár, J. Mekyska, I. Rektorová, L. Masarová, Z. Smékal, M. Faundez-Zanuy, Decision support framework for Parkinson's disease based on novel handwriting markers, IEEE Trans Neural Syst Rehabil Eng 23(3) (2015) 508-516.
60 P. Drotár, J. Mekyska, I. Rektorová, L. Masarová, Z. Smékal, M. Faundez-Zanuy, Evaluation of handwriting kinematics and pressure for differential diagnosis of Parkinson's disease, Artif Intell Med 67 (2016) 39-46.
61 G. Du, M.M. Lewis, S. Kanekar, N.W. Sterling, L. He, L. Kong, R. Li, X. Huang, Combined Diffusion Tensor Imaging and Apparent Transverse Relaxation Rate Differentiate Parkinson Disease and Atypical Parkinsonism, AJNR Am J Neuroradiol 38(5) (2017) 966-972.
62 B. Erdogdu Sakar, G. Serbes, C.O. Sakar, Analyzing the effectiveness of vocal features in early telediagnosis of Parkinson's disease, PLoS One 12(8) (2017) e0182428-e0182428.
63 J.P. Félix, F.H.T. Vieira, Á.A. Cardoso, M.V.G. Ferreira, R.A.P. Franco, M.A. Ribeiro, S.G. Araújo, H.P. Corrêa, M.L. Carneiro, A Parkinson's Disease Classification Method: An Approach Using Gait Dynamics and Detrended Fluctuation Analysis, 2019 IEEE Canadian Conference of Electrical and Computer Engineering (CCECE), 2019, pp. 1-4.
64 C. Fernandes, L. Fonseca, F. Ferreira, M. Gago, L. Costa, N. Sousa, C. Ferreira, J. Gama, W. Erlhagen, E. Bicho, Artificial Neural Networks Classification of Patients with Parkinsonism based on Gait, 2018 IEEE International Conference on Bioinformatics and Biomedicine (BIBM), 2018, pp. 2024-2030.
65 N.K. Focke, G. Helms, S. Scheewe, P.M. Pantel, C.G. Bachmann, P. Dechent, J. Ebentheuer, A. Mohr, W. Paulus, C. Trenkwalder, Individual voxel-based subtype prediction can differentiate progressive supranuclear palsy from idiopathic Parkinson syndrome and healthy controls, Hum Brain Mapp 32(11) (2011) 1905-1915.
66 A. Frid, E.J. Safra, H. Hazan, L.L. Lokey, D. Hilu, L. Manevitz, L.O. Ramig, S. Sapir, Computational Diagnosis of Parkinson's Disease Directly from Natural Speech Using Machine Learning Techniques, 2014 IEEE International Conference on Software Science, Technology and Engineering, 2014, pp. 50-53.
67 E. Glaab, J.-P. Trezzi, A. Greuel, C. Jäger, Z. Hodak, A. Drzezga, L. Timmermann, M. Tittgemeyer, N.J. Diederich, C. Eggers, Integrative analysis of blood metabolomics and PET brain neuroimaging data for Parkinson's disease, Neurobiol Dis 124 (2019) 555-562.
68 H. Gunduz, Deep Learning-Based Parkinson's Disease Classification Using Vocal Feature Sets, IEEE Access 7 (2019) 115540-115551.
69 N. Haji Ghassemi, J. Hannink, N. Roth, H. Gaßner, F. Marxreiter, J. Klucken, B.M. Eskofier, Turning Analysis during Standardized Test Using On-Shoe Wearable Sensors in Parkinson's Disease, Sensors (Basel) 19(14) (2019) 3103.
70 S. Haller, S. Badoud, D. Nguyen, I. Barnaure, M.L. Montandon, K.O. Lovblad, P.R. Burkhard, Differentiation between Parkinson disease and other forms of Parkinsonism using support vector machine analysis of susceptibility-weighted imaging (SWI): initial results, Eur Radiol 23(1) (2013) 12-19.
71 S. Haller, S. Badoud, D. Nguyen, V. Garibotto, K.O. Lovblad, P.R. Burkhard, Individual detection of patients with Parkinson disease using support vector machine analysis of diffusion tensor imaging data: initial results, AJNR Am J Neuroradiol 33(11) (2012) 2123-2128.
72 A.U. Haq, J. Li, M.H. Memon, J. Khan, S.U. Din, I. Ahad, R. Sun, Z. Lai, Comparative Analysis of the Classification Performance of Machine Learning Classifiers and Deep Neural

Network Classifier for Prediction of Parkinson Disease, 2018 15th International Computer Conference on Wavelet Active Media Technology and Information Processing (ICCWAMTIP), 2018, pp. 101-106.

73 A.U. Haq, J.P. Li, M.H. Memon, J. khan, A. Malik, T. Ahmad, A. Ali, S. Nazir, I. Ahad, M. Shahid, Feature Selection Based on L1-Norm Support Vector Machine and Effective Recognition System for Parkinson’s Disease Using Voice Recordings, IEEE Access 7 (2019) 37718-37734.

74 M. Hariharan, K. Polat, R. Sindhu, A new hybrid intelligent system for accurate detection of Parkinson's disease, Comput Methods Programs Biomed 113(3) (2014) 904-913.

75 T.J. Hirschauer, H. Adeli, J.A. Buford, Computer-Aided Diagnosis of Parkinson's Disease Using Enhanced Probabilistic Neural Network, J Med Syst 39(11) (2015) 179-179.

76 S.-Y. Hsu, H.-C. Lin, T.-B. Chen, W.-C. Du, Y.-H. Hsu, Y.-C. Wu, P.-W. Tu, Y.-H. Huang, H.-Y. Chen, Feasible Classified Models for Parkinson Disease from (99m)Tc-TRODAT-1 SPECT Imaging, Sensors (Basel) 19(7) (2019) 1740.

77 I. Huertas-Fernández, F.J. García-Gómez, D. García-Solís, S. Benítez-Rivero, V.A. Marín-Oyaga, S. Jesús, M.T. Cáceres-Redondo, J.A. Lojo, J.F. Martín-Rodríguez, F. Carrillo, P. Mir, Machine learning models for the differential diagnosis of vascular parkinsonism and Parkinson's disease using [(123)I]FP-CIT SPECT, Eur J Nucl Med Mol Imaging 42(1) (2015) 112-119.

78 H.-J. Huppertz, L. Möller, M. Südmeyer, R. Hilker, E. Hattingen, K. Egger, F. Amtage, G. Respondek, M. Stamelou, A. Schnitzler, E.H. Pinkhardt, W.H. Oertel, S. Knake, J. Kassubek, G.U. Höglinger, Differentiation of neurodegenerative parkinsonian syndromes by volumetric magnetic resonance imaging analysis and support vector machine classification, Mov Disord 31(10) (2016) 1506-1517.

79 K. İ, S. Ulukaya, O. Erdem, Classification of Parkinson’s Disease Using Dynamic Time Warping, 2019 27th Telecommunications Forum (TELFOR), 2019, pp. 1-4.

80 I.A. Illan, J.M. Gorrz, J. Ramirez, F. Segovia, J.M. Jimenez-Hoyuela, S.J. Ortega Lozano, Automatic assistance to Parkinson's disease diagnosis in DaTSCAN SPECT imaging, Med Phys 39(10) (2012) 5971-5980.

81 M.S. Islam, I. Parvez, D. Hai, P. Goswami, Performance comparison of heterogeneous classifiers for detection of Parkinson's disease using voice disorder (dysphonia), 2014 International Conference on Informatics, Electronics & Vision (ICIEV), 2014, pp. 1-7.

82 F. Javed, I. Thomas, M. Memedi, A comparison of feature selection methods when using motion sensors data: a case study in Parkinson's disease, Conf Proc IEEE Eng Med Biol Soc 2018 (2018) 5426-5429.

83 W. Ji, Y. Li, Energy-based feature ranking for assessing the dysphonia measurements in Parkinson detection, IET Signal Processing 6(4) (2012) 300-305.

84 D. Joshi, A. Khajuria, P. Joshi, An automatic non-invasive method for Parkinson's disease classification, Comput Methods Programs Biomed 145 (2017) 135-145.

85 S.B. Junior, V.G.T. Costa, S. Chen, R.C. Guido, U-Healthcare System for Pre-Diagnosis of Parkinson's Disease from Voice Signal, 2018 IEEE International Symposium on Multimedia (ISM), 2018, pp. 271-274.

86 K. Kamagata, A. Zalesky, T. Hatano, M.A. Di Biase, O. El Samad, S. Saiki, K. Shimoji, K.K. Kumamaru, K. Kamiya, M. Hori, N. Hattori, S. Aoki, C. Pantelis, Connectome analysis with diffusion MRI in idiopathic Parkinson's disease: Evaluation using multi-shell, multi-tissue, constrained spherical deconvolution, Neuroimage Clin 17 (2017) 518-529.

87 Z. Karapinar Senturk, Early diagnosis of Parkinson's disease using machine learning algorithms, Med Hypotheses 138 (2020) 109603-109603.

88 A. Kazeminejad, S. Golbabaei, H. Soltanian-Zadeh, Graph theoretical metrics and machine learning for diagnosis of Parkinson's disease using rs-fMRI, 2017 Artificial Intelligence and Signal Processing Conference (AISP), 2017, pp. 134-139.
89 M.M. Khan, A. Mendes, S.K. Chalup, Evolutionary Wavelet Neural Network ensembles for breast cancer and Parkinson's disease prediction, PLoS One 13(2) (2018) e0192192-e0192192.
90 P. Khatamino, C. İ, L. Özyılmaz, A Deep Learning-CNN Based System for Medical Diagnosis: An Application on Parkinson's Disease Handwriting Drawings, 2018 6th International Conference on Control Engineering & Information Technology (CEIT), 2018, pp. 1-6.
91 N. Khoury, F. Attal, Y. Amirat, L. Oukhellou, S. Mohammed, Data-Driven Based Approach to Aid Parkinson's Disease Diagnosis, Sensors (Basel) 19(2) (2019) 242.
92 S. Kiryu, K. Yasaka, H. Akai, Y. Nakata, Y. Sugomori, S. Hara, M. Seo, O. Abe, K. Ohtomo, Deep learning to differentiate parkinsonian disorders separately using single midsagittal MR imaging: a proof of concept study, Eur Radiol 29(12) (2019) 6891-6899.
93 Y. Klein, R. Djaldetti, Y. Keller, I. Bachelet, Motor dysfunction and touch-slang in user interface data, Sci Rep 7(1) (2017) 4702-4702.
94 A. Klomsae, S. Auephanwiriyakul, N. Theera-Umpon, String Grammar Unsupervised Possibilistic Fuzzy C-Medians for Gait Pattern Classification in Patients with Neurodegenerative Diseases, Comput Intell Neurosci 2018 (2018) 1869565-1869565.
95 A. Koçer, A.B. Oktay, Nintendo Wii assessment of Hoehn and Yahr score with Parkinson's disease tremor, Technol Health Care 24(2) (2016) 185-191.
96 N. Kostikis, D. Hristu-Varsakelis, M. Arnaoutoglou, C. Kotsavasiloglou, A Smartphone-Based Tool for Assessing Parkinsonian Hand Tremor, IEEE J Biomed Health Inform 19(6) (2015) 1835-1842.
97 P. Kraipeerapun, S. Amornsamankul, Using stacked generalization and complementary neural networks to predict Parkinson's disease, 2015 11th International Conference on Natural Computation (ICNC), 2015, pp. 1290-1294.
98 P. Kugler, C. Jaremenko, J. Schlachetzki, J. Winkler, J. Klucken, B. Eskofier, Automatic recognition of Parkinson's disease using surface electromyography during standardized gait tests, Conf Proc IEEE Eng Med Biol Soc 2013 (2013) 5781-5784.
99 A. Kuhner, T. Schubert, M. Cenciarini, I.K. Wiesmeier, V.A. Coenen, W. Burgard, C. Weiller, C. Maurer, Correlations between Motor Symptoms across Different Motor Tasks, Quantified via Random Forest Feature Classification in Parkinson's Disease, Front Neurol 8 (2017) 607-607.
100 H. Kuresan, D. Samiappan, S. Masunda, Fusion of WPT and MFCC feature extraction in Parkinson's disease diagnosis, Technol Health Care 27(4) (2019) 363-372.
101 S.E. Lacy, S.L. Smith, M.A. Lones, Using echo state networks for classification: A case study in Parkinson's disease diagnosis, Artif Intell Med 86 (2018) 53-59.
102 H. Lei, Z. Huang, F. Zhou, A. Elazab, E.-L. Tan, H. Li, J. Qin, B. Lei, Parkinson's Disease Diagnosis via Joint Learning From Multiple Modalities and Relations, IEEE J Biomed Health Inform 23(4) (2019) 1437-1449.
103 P.A. Lewitt, J. Li, M. Lu, T.G. Beach, C.H. Adler, L. Guo, C. Arizona Parkinson's Disease, 3-hydroxykynurenine and other Parkinson's disease biomarkers discovered by metabolomic analysis, Mov Disord 28(12) (2013) 1653-1660.
104 Q. Li, H. Chen, H. Huang, X. Zhao, Z. Cai, C. Tong, W. Liu, X. Tian, An Enhanced Grey Wolf Optimization Based Feature Selection Wrapped Kernel Extreme Learning Machine for Medical Diagnosis, Comput Math Methods Med 2017 (2017) 9512741-9512741.

105 S. Li, H. Lei, F. Zhou, J. Gardezi, B. Lei, Longitudinal and Multi-modal Data Learning for Parkinson’s Disease Diagnosis via Stacked Sparse Auto-encoder, 2019 IEEE 16th International Symposium on Biomedical Imaging (ISBI 2019), 2019, pp. 384-387.
106 H. Liu, G. Du, L. Zhang, M.M. Lewis, X. Wang, T. Yao, R. Li, X. Huang, Folded concave penalized learning in identifying multimodal MRI marker for Parkinson's disease, J Neurosci Methods 268 (2016) 1-6.
107 L. Liu, Q. Wang, E. Adeli, L. Zhang, H. Zhang, D. Shen, Feature Selection Based on Iterative Canonical Correlation Analysis for Automatic Diagnosis of Parkinson's Disease, Med Image Comput Comput Assist Interv 9901 (2016) 1-8.
108 C. Ma, J. Ouyang, H.-L. Chen, X.-H. Zhao, An efficient diagnosis system for Parkinson's disease using kernel-based extreme learning machine with subtractive clustering features weighting approach, Comput Math Methods Med 2014 (2014) 985789-985789.
109 H. Ma, T. Tan, H. Zhou, T. Gao, Support Vector Machine-recursive feature elimination for the diagnosis of Parkinson disease based on speech analysis, 2016 Seventh International Conference on Intelligent Control and Information Processing (ICICIP), 2016, pp. 34-40.
110 F. Maass, B. Michalke, A. Leha, M. Boerger, I. Zerr, J.-C. Koch, L. Tönges, M. Bähr, P. Lingor, Elemental fingerprint as a cerebrospinal fluid biomarker for the diagnosis of Parkinson's disease, J Neurochem 145(4) (2018) 342-351.
111 F. Maass, B. Michalke, D. Willkommen, A. Leha, C. Schulte, L. Tönges, B. Mollenhauer, C. Trenkwalder, D. Rückamp, M. Börger, I. Zerr, M. Bähr, P. Lingor, Elemental fingerprint: Reassessment of a cerebrospinal fluid biomarker for Parkinson's disease, Neurobiol Dis 134 (2020) 104677-104677.
112 R. Mabrouk, B. Chikhaoui, L. Bentabet, Machine Learning Based Classification Using Clinical and DaTSCAN SPECT Imaging Features: A Study on Parkinson’s Disease and SWEDD, IEEE Transactions on Radiation and Plasma Medical Sciences 3(2) (2019) 170-177.
113 I. Mandal, N. Sairam, Accurate telemonitoring of Parkinson's disease diagnosis using robust inference system, Int J Med Inform 82(5) (2013) 359-377.
114 S. Marar, D. Swain, V. Hiwarkar, N. Motwani, A. Awari, Predicting the occurrence of Parkinson’s Disease using various Classification Models, 2018 International Conference on Advanced Computation and Telecommunication (ICACAT), 2018, pp. 1-5.
115 M. Martínez, F. Villagra, J.M. Castellote, M.A. Pastor, Kinematic and Kinetic Patterns Related to Free-Walking in Parkinson's Disease, Sensors (Basel) 18(12) (2018) 4224.
116 F.J. Martinez-Murcia, J.M. Górriz, J. Ramírez, A. Ortiz, Convolutional Neural Networks for Neuroimaging in Parkinson's Disease: Is Preprocessing Needed?, Int J Neural Syst 28(10) (2018) 1850035-1850035.
117 M. Memedi, A. Sadikov, V. Groznik, J. Žabkar, M. Možina, F. Bergquist, A. Johansson, D. Haubenberger, D. Nyholm, Automatic Spiral Analysis for Objective Assessment of Motor Symptoms in Parkinson's Disease, Sensors (Basel) 15(9) (2015) 23727-23744.
118 Y. Mittra, V. Rustagi, Classification of Subjects with Parkinson's Disease Using Gait Data Analysis, 2018 International Conference on Automation and Computational Engineering (ICACE), 2018, pp. 84-89.
119 Z.A. Moharkan, H. Garg, T. Chodhury, P. Kumar, A classification based Parkinson detection system, 2017 International Conference On Smart Technologies For Smart Nation (SmartTechCon), 2017, pp. 1509-1513.
120 D. Montaña, Y. Campos-Roca, C.J. Pérez, A Diadochokinesis-based expert system considering articulatory features of plosive consonants for early detection of Parkinson's disease, Comput Methods Programs Biomed 154 (2018) 89-97.

121 R. Morisi, D.N. Manners, G. Gnecco, N. Lanconelli, C. Testa, S. Evangelisti, L. Talozzi, L.L. Gramegna, C. Bianchini, G. Calandra-Buonaura, L. Sambati, G. Giannini, P. Cortelli, C. Tonon, R. Lodi, Multi-class parkinsonian disorders classification with quantitative MR markers and graph-based features using support vector machines, Parkinsonism Relat Disord 47 (2018) 64-70.
122 J. Mucha, J. Mekyska, M. Faundez-Zanuy, K. Lopez-De-Ipina, V. Zvoncak, Z. Galaz, T. Kiska, Z. Smekal, L. Brabenec, I. Rektorova, Advanced Parkinson's Disease Dysgraphia Analysis Based on Fractional Derivatives of Online Handwriting, 2018 10th International Congress on Ultra Modern Telecommunications and Control Systems and Workshops (ICUMT), 2018, pp. 1-6.
123 N. Nicastro, J. Wegrzyk, M.G. Preti, V. Fleury, D. Van de Ville, V. Garibotto, P.R. Burkhard, Classification of degenerative parkinsonism subtypes by support-vector-machine analysis and striatal (123)I-FP-CIT indices, J Neurol 266(7) (2019) 1771-1781.
124 S. Nõmm, K. Bardõš, A. Toomela, K. Medijainen, P. Taba, Detailed Analysis of the Luria's Alternating SeriesTests for Parkinson's Disease Diagnostics, 2018 17th IEEE International Conference on Machine Learning and Applications (ICMLA), 2018, pp. 1347-1352.
125 A. Nunes, G. Silva, C. Duque, C. Januário, I. Santana, A.F. Ambrósio, M. Castelo-Branco, R. Bernardes, Retinal texture biomarkers may help to discriminate between Alzheimer's, Parkinson's, and healthy controls, PLoS One 14(6) (2019) e0218826-e0218826.
126 S. Nuvoli, A. Spanu, M.L. Fravolini, F. Bianconi, S. Cascianelli, G. Madeddu, B. Palumbo, [(123)I]Metaiodobenzylguanidine (MIBG) Cardiac Scintigraphy and Automated Classification Techniques in Parkinsonian Disorders, Mol Imaging Biol (2019) 10.1007/s11307-019-01406-6.
127 F.H.M. Oliveira, A.R.P. Machado, A.O. Andrade, On the Use of t-Distributed Stochastic Neighbor Embedding for Data Visualization and Classification of Individuals with Parkinson's Disease, Comput Math Methods Med 2018 (2018) 8019232-8019232.
128 F.P.M. Oliveira, M. Castelo-Branco, Computer-aided diagnosis of Parkinson's disease based on [(123)I]FP-CIT SPECT binding potential images, using the voxels-as-features approach and support vector machines, J Neural Eng 12(2) (2015) 026008-026008.
129 F.P.M. Oliveira, D.B. Faria, D.C. Costa, M. Castelo-Branco, J.M.R.S. Tavares, Extraction, selection and comparison of features for an effective automated computer-aided diagnosis of Parkinson's disease based on [(123)I]FP-CIT SPECT images, Eur J Nucl Med Mol Imaging 45(6) (2018) 1052-1062.
130 Q.W. Oung, M. Hariharan, H.L. Lee, S.N. Basah, M. Sarillee, C.H. Lee, Wearable multimodal sensors for evaluation of patients with Parkinson disease, 2015 IEEE International Conference on Control System, Computing and Engineering (ICCSCE), 2015, pp. 269-274.
131 A. Ozcift, SVM feature selection based rotation forest ensemble classifiers to improve computer-aided diagnosis of Parkinson disease, J Med Syst 36(4) (2012) 2141-2147.
132 A. Ozcift, A. Gulten, Classifier ensemble construction with rotation forest to improve medical diagnosis performance of machine learning algorithms, Comput Methods Programs Biomed 104(3) (2011) 443-451.
133 G. Pahuja, T.N. Nagabhushan, A novel GA-ELM approach for Parkinson's disease detection using brain structural T1-weighted MRI data, 2016 Second International Conference on Cognitive Computing and Information Processing (CCIP), 2016, pp. 1-6.
134 B. Palumbo, M.L. Fravolini, T. Buresta, F. Pompili, N. Forini, P. Nigro, P. Calabresi, N. Tambasco, Diagnostic accuracy of Parkinson disease by support vector machine (SVM) analysis of 123I-FP-CIT brain SPECT data: implications of putaminal findings and age, Medicine (Baltimore) 93(27) (2014) e228-e228.
135 A. Papadopoulos, K. Kyritsis, L. Klingelhoefer, S. Bostanjopoulou, K.R. Chaudhuri, A. Delopoulos, Detecting Parkinsonian Tremor from IMU Data Collected In-The-Wild using Deep Multiple-Instance Learning, IEEE J Biomed Health Inform (2019) 1-1.
136 I. Papavasileiou, W. Zhang, X. Wang, J. Bi, L. Zhang, S. Han, Classification of Neurological Gait Disorders Using Multi-task Feature Learning, 2017 IEEE/ACM International Conference on Connected Health: Applications, Systems and Engineering Technologies (CHASE), 2017, pp. 195-204.
137 M. Peker, A decision support system to improve medical diagnosis using a combination of k-medoids clustering based attribute weighting and SVM, J Med Syst 40(5) (2016) 116-116.
138 B. Peng, S. Wang, Z. Zhou, Y. Liu, B. Tong, T. Zhang, Y. Dai, A multilevel-ROI-features-based machine learning method for detection of morphometric biomarkers in Parkinson's disease, Neurosci Lett 651 (2017) 88-94.
139 B. Peng, Z. Zhou, C. Geng, B. Tong, Z. Zhou, T. Zhang, Y. Dai, Computer aided analysis of cognitive disorder in patients with Parkinsonism using machine learning method with multilevel ROI-based features, 2016 9th International Congress on Image and Signal Processing, BioMedical Engineering and Informatics (CISP-BMEI), 2016, pp. 1792-1796.
140 C.R. Pereira, D.R. Pereira, G.H. Rosa, V.H.C. Albuquerque, S.A.T. Weber, C. Hook, J.P. Papa, Handwritten dynamics assessment through convolutional neural networks: An application to Parkinson's disease identification, Artif Intell Med 87 (2018) 67-77.
141 C.R. Pereira, D.R. Pereira, F.A. Silva, J.P. Masieiro, S.A.T. Weber, C. Hook, J.P. Papa, A new computer vision-based approach to aid the diagnosis of Parkinson's disease, Comput Methods Programs Biomed 136 (2016) 79-88.
142 C.R. Pereira, D.R. Pereira, F.A.d. Silva, C. Hook, S.A.T. Weber, L.A.M. Pereira, J.P. Papa, A Step Towards the Automated Diagnosis of Parkinson's Disease: Analyzing Handwriting Movements, 2015 IEEE 28th International Symposium on Computer-Based Medical Systems, 2015, pp. 171-176.
143 C.R. Pereira, S.A.T. Weber, C. Hook, G.H. Rosa, J.P. Papa, Deep Learning-Aided Parkinson's Disease Diagnosis from Handwritten Dynamics, 2016 29th SIBGRAPI Conference on Graphics, Patterns and Images (SIBGRAPI), 2016, pp. 340-346.
144 H.N. Pham, T.T.T. Do, K.Y.J. Chan, G. Sen, A.Y.K. Han, P. Lim, T.S.L. Cheng, Q.H. Nguyen, B.P. Nguyen, M.C.H. Chua, Multimodal Detection of Parkinson Disease based on Vocal and Improved Spiral Test, 2019 International Conference on System Science and Engineering (ICSSE), 2019, pp. 279-284.
145 T.D. Pham, Pattern analysis of computer keystroke time series in healthy control and early-stage Parkinson's disease subjects using fuzzy recurrence and scalable recurrence network features, J Neurosci Methods 307 (2018) 194-202.
146 T.D. Pham, H. Yan, Tensor Decomposition of Gait Dynamics in Parkinson's Disease, IEEE Trans Biomed Eng 65(8) (2018) 1820-1827.
147 R. Prashanth, S. Dutta Roy, Early detection of Parkinson's disease through patient questionnaire and predictive modelling, Int J Med Inform 119 (2018) 75-87.
148 R. Prashanth, S. Dutta Roy, P.K. Mandal, S. Ghosh, High-Accuracy Detection of Early Parkinson's Disease through Multimodal Features and Machine Learning, Int J Med Inform 90 (2016) 13-21.

149 R. Prashanth, S.D. Roy, P.K. Mandal, S. Ghosh, Parkinson's disease detection using olfactory loss and REM sleep disorder features, 2014 36th Annual International Conference of the IEEE Engineering in Medicine and Biology Society, 2014, pp. 5764-5767.
150 R. Prashanth, S.D. Roy, P.K. Mandal, S. Ghosh, High-Accuracy Classification of Parkinson's Disease Through Shape Analysis and Surface Fitting in 123I-Ioflupane SPECT Imaging, IEEE J Biomed Health Inform 21(3) (2017) 794-802.
151 J. Prince, F. Andreotti, M.D. Vos, Multi-Source Ensemble Learning for the Remote Prediction of Parkinson's Disease in the Presence of Source-Wise Missing Data, IEEE Transactions on Biomedical Engineering 66(5) (2019) 1402-1411.
152 J. Prince, M. de Vos, A Deep Learning Framework for the Remote Detection of Parkinson'S Disease Using Smart-Phone Sensor Data, Conf Proc IEEE Eng Med Biol Soc 2018 (2018) 3144-3147.
153 J.F. Reyes, J.S. Montealegre, Y.J. Castano, C. Urcuqui, A. Navarro, LSTM and Convolution Networks exploration for Parkinson’s Diagnosis, 2019 IEEE Colombian Conference on Communications and Computing (COLCOM), 2019, pp. 1-4.
154 L.C.F. Ribeiro, L.C.S. Afonso, J.P. Papa, Bag of Samplings for computer-assisted Parkinson's disease diagnosis based on Recurrent Neural Networks, Comput Biol Med 115 (2019) 103477-103477.
155 M. Ricci, G.D. Lazzaro, A. Pisani, N.B. Mercuri, F. Giannini, G. Saggio, Assessment of Motor Impairments in Early Untreated Parkinson's Disease Patients: The Wearable Electronics Impact, IEEE J Biomed Health Inform 24(1) (2020) 120-130.
156 C.D. Rios-Urrego, J.C. Vásquez-Correa, J.F. Vargas-Bonilla, E. Nöth, F. Lopera, J.R. Orozco-Arroyave, Analysis and evaluation of handwriting in patients with Parkinson's disease using kinematic, geometrical, and non-linear features, Comput Methods Programs Biomed 173 (2019) 43-52.
157 E. Rovini, C. Maremmani, A. Moschetti, D. Esposito, F. Cavallo, Comparative Motor Pre-clinical Assessment in Parkinson's Disease Using Supervised Machine Learning Approaches, Ann Biomed Eng 46(12) (2018) 2057-2068.
158 E. Rovini, A. Moschetti, L. Fiorini, D. Esposito, C. Maremmani, F. Cavallo, Wearable Sensors for Prodromal Motor Assessment of Parkinson’s Disease using Supervised Learning*, 2019 41st Annual International Conference of the IEEE Engineering in Medicine and Biology Society (EMBC), 2019, pp. 4318-4321.
159 C. Rubbert, C. Mathys, C. Jockwitz, C.J. Hartmann, S.B. Eickhoff, F. Hoffstaedter, S. Caspers, C.R. Eickhoff, B. Sigl, N.A. Teichert, M. Südmeyer, B. Turowski, A. Schnitzler, J. Caspers, Machine-learning identifies Parkinson's disease patients based on resting-state between-network functional connectivity, Br J Radiol 92(1101) (2019) 20180886-20180886.
160 B.E. Sakar, M.E. Isenkul, C.O. Sakar, A. Sertbas, F. Gurgen, S. Delil, H. Apaydin, O. Kursun, Collection and analysis of a Parkinson speech dataset with multiple types of sound recordings, IEEE J Biomed Health Inform 17(4) (2013) 828-834.
161 C. Salvatore, A. Cerasa, I. Castiglioni, F. Gallivanone, A. Augimeri, M. Lopez, G. Arabia, M. Morelli, M.C. Gilardi, A. Quattrone, Machine learning on brain MRI data for differential diagnosis of Parkinson's disease and Progressive Supranuclear Palsy, J Neurosci Methods 222 (2014) 230-237.
162 O.N.A. Sayaydeha, M.F. Mohammad, Diagnosis of The Parkinson Disease Using Enhanced Fuzzy Min-Max Neural Network and OneR Attribute Evaluation Method, 2019 International Conference on Advanced Science and Engineering (ICOASE), 2019, pp. 64-69.
163 F. Segovia, J.M. Górriz, J. Ramírez, J. Levin, M. Schuberth, M. Brendel, A. Rominger, G. Garraux, C. Phillips, Analysis of 18F-DMFP PET data using multikernel classification in order to assist the diagnosis of Parkinsonism, 2015 IEEE Nuclear Science Symposium and Medical Imaging Conference (NSS/MIC), 2015, pp. 1-4.
164 F. Segovia, J.M. Górriz, J. Ramírez, F.J. Martínez-Murcia, D. Castillo-Barnes, Assisted Diagnosis of Parkinsonism Based on the Striatal Morphology, Int J Neural Syst 29(9) (2019) 1950011-1950011.
165 M.K. Shahsavari, H. Rashidi, H.R. Bakhsh, Efficient classification of Parkinson's disease using extreme learning machine and hybrid particle swarm optimization, 2016 4th International Conference on Control, Instrumentation, and Automation (ICCIA), 2016, pp. 148-154.
166 R. Shamir, C. Klein, D. Amar, E.-J. Vollstedt, M. Bonin, M. Usenovic, Y.C. Wong, A. Maver, S. Poths, H. Safer, J.-C. Corvol, S. Lesage, O. Lavi, G. Deuschl, G. Kuhlenbaeumer, H. Pawlack, I. Ulitsky, M. Kasten, O. Riess, A. Brice, B. Peterlin, D. Krainc, Analysis of blood-based gene expression in idiopathic Parkinson disease, Neurology 89(16) (2017) 1676-1683.
167 R. Sheibani, E. Nikookar, S.E. Alavi, An Ensemble Method for Diagnosis of Parkinson's Disease Based on Voice Measurements, J Med Signals Sens 9(4) (2019) 221-226.
168 T. Shen, J. Jiang, W. Lin, J. Ge, P. Wu, Y. Zhou, C. Zuo, J. Wang, Z. Yan, K. Shi, Use of Overlapping Group LASSO Sparse Deep Belief Network to Discriminate Parkinson's Disease and Normal Control, Front Neurosci 13 (2019) 396-396.
169 J. Shi, M. Yan, Y. Dong, X. Zheng, Q. Zhang, H. An, Multiple Kernel Learning Based Classification of Parkinson’s Disease With Multi-Modal Transcranial Sonography, 2018 40th Annual International Conference of the IEEE Engineering in Medicine and Biology Society (EMBC), 2018, pp. 61-64.
170 S. Shinde, S. Prasad, Y. Saboo, R. Kaushick, J. Saini, P.K. Pal, M. Ingalhalikar, Predictive markers for Parkinson's disease using deep neural nets on neuromelanin sensitive MRI, Neuroimage Clin 22 (2019) 101748-101748.
171 G. Singh, L. Samavedham, Unsupervised learning based feature extraction for differential diagnosis of neurodegenerative diseases: A case study on early-stage diagnosis of Parkinson disease, J Neurosci Methods 256 (2015) 30-40.
172 G. Singh, L. Samavedham, E.C.-H. Lim, I. Alzheimer’s Disease Neuroimaging, I. Parkinson Progression Marker, Determination of Imaging Biomarkers to Decipher Disease Trajectories and Differential Diagnosis of Neurodegenerative Diseases (DIsease TreND), J Neurosci Methods 305 (2018) 105-116.
173 D. Stoessel, C. Schulte, M.C. Teixeira Dos Santos, D. Scheller, I. Rebollo-Mesa, C. Deuschle, D. Walther, N. Schauer, D. Berg, A. Nogueira da Costa, W. Maetzler, Promising Metabolite Profiles in the Plasma and CSF of Early Clinical Parkinson's Disease, Front Aging Neurosci 10 (2018) 51-51.
174 D. Surangsrirat, C. Thanawattano, R. Pongthornseri, S. Dumnin, C. Anan, R. Bhidayasiri, Support vector machine classification of Parkinson's disease and essential tremor subjects based on temporal fluctuation, Conf Proc IEEE Eng Med Biol Soc 2016 (2016) 6389-6392.
175 D. Sztahó, M.G. Tulics, K. Vicsi, I. Valálik, Automatic estimation of severity of Parkinson's disease based on speech rhythm related features, 2017 8th IEEE International Conference on Cognitive Infocommunications (CogInfoCom), 2017, pp. 000011-000016.
176 D. Sztahó, I. Valálik, K. Vicsi, Parkinson’s Disease Severity Estimation on Hungarian Speech Using Various Speech Tasks, 2019 International Conference on Speech Technology and Human-Computer Dialogue (SpeD), 2019, pp. 1-6.

177 H.D. Tagare, C. DeLorenzo, S. Chelikani, L. Saperstein, R.K. Fulbright, Voxel-based logistic analysis of PPMI control and Parkinson's disease DaTscans, Neuroimage 152 (2017) 299-311.
178 F. Tahavori, E. Stack, V. Agarwal, M. Burnett, A. Ashburn, S.A. Hoseinitabatabaei, W. Harwin, Physical activity recognition of elderly people and people with parkinson's (PwP) during standard mobility tests using wearable sensors, 2017 International Smart Cities Conference (ISC2), 2017, pp. 1-4.
179 C. Taleb, M. Khachab, C. Mokbel, L. Likforman-Sulem, Visual Representation of Online Handwriting Time Series for Deep Learning Parkinson's Disease Detection, 2019 International Conference on Document Analysis and Recognition Workshops (ICDARW), 2019, pp. 25-30.
180 Y. Tang, L. Meng, C.-M. Wan, Z.-H. Liu, W.-H. Liao, X.-X. Yan, X.-Y. Wang, B.-S. Tang, J.-F. Guo, Identifying the presence of Parkinson's disease using low-frequency fluctuations in BOLD signals, Neurosci Lett 645 (2017) 1-6.
181 J.C. Taylor, J.W. Fenner, Comparison of machine learning and semi-quantification algorithms for (I123)FP-CIT classification: the beginning of the end for semi-quantification?, EJNMMI Phys 4(1) (2017) 29-29.
182 I. Tien, S.D. Glaser, M.J. Aminoff, Characterization of gait abnormalities in Parkinson's disease using a wireless inertial sensor system, 2010 Annual International Conference of the IEEE Engineering in Medicine and Biology, 2010, pp. 3353-3356.
183 J.M. Tracy, Y. Özkanca, D.C. Atkins, R. Hosseini Ghomi, Investigating voice as a biomarker: Deep phenotyping methods for early detection of Parkinson's disease, J Biomed Inform (2019) 103362-103362.
184 J.-P. Trezzi, S. Galozzi, C. Jaeger, K. Barkovits, K. Brockmann, W. Maetzler, D. Berg, K. Marcus, F. Betsou, K. Hiller, B. Mollenhauer, Distinct metabolomic signature in cerebrospinal fluid in early parkinson's disease, Mov Disord 32(10) (2017) 1401-1408.
185 A. Tsanas, M.A. Little, P.E. McSharry, J. Spielman, L.O. Ramig, Novel speech signal processing algorithms for high-accuracy classification of Parkinson's disease, IEEE Trans Biomed Eng 59(5) (2012) 1264-1271.
186 P.-H. Tseng, I.G.M. Cameron, G. Pari, J.N. Reynolds, D.P. Munoz, L. Itti, High-throughput classification of clinical populations from natural viewing eye movements, J Neurol 260(1) (2013) 275-284.
187 M. Tsuda, S. Asano, Y. Kato, K. Murai, M. Miyazaki, Differential diagnosis of multiple system atrophy with predominant parkinsonism and Parkinson's disease using neural networks, J Neurol Sci 401 (2019) 19-26.
188 C. Urcuqui, Y. Castaño, J. Delgado, A. Navarro, J. Diaz, B. Muñoz, J. Orozco, Exploring Machine Learning to Analyze Parkinson's Disease Patients, 2018 14th International Conference on Semantics, Knowledge and Grids (SKG), 2018, pp. 160-166.
189 E. Vaiciukynas, A. Verikas, A. Gelzinis, M. Bacauskiene, Detecting Parkinson's disease from sustained phonation and speech signals, PLoS One 12(10) (2017) e0185613-e0185613.
190 M.I. Vanegas, M.F. Ghilardi, S.P. Kelly, A. Blangero, Machine learning for EEG-based biomarkers in Parkinson’s disease, 2018 IEEE International Conference on Bioinformatics and Biomedicine (BIBM), 2018, pp. 2661-2665.
191 C. Váradi, K. Nehéz, O. Hornyák, B. Viskolcz, J. Bones, Serum N-Glycosylation in Parkinson's Disease: A Novel Approach for Potential Alterations, Molecules 24(12) (2019) 2220.

192 J.C. Vásquez-Correa, T. Arias-Vergara, J.R. Orozco-Arroyave, B. Eskofier, J. Klucken, E. Nöth, Multimodal Assessment of Parkinson's Disease: A Deep Learning Approach, IEEE J Biomed Health Inform 23(4) (2019) 1618-1630.
193 A. Vlachostergiou, A. Tagaris, A. Stafylopatis, S. Kollias, Multi-Task Learning for Predicting Parkinson's Disease Based on Medical Imaging Information, 2018 25th IEEE International Conference on Image Processing (ICIP), 2018, pp. 2052-2056.
194 F. Wahid, R.K. Begg, C.J. Hass, S. Halgamuge, D.C. Ackland, Classification of Parkinson's Disease Gait Using Spatial-Temporal Gait Features, IEEE J Biomed Health Inform 19(6) (2015) 1794-1802.
195 Z. Wang, X. Zhu, E. Adeli, Y. Zhu, F. Nie, B. Munsell, G. Wu, Adni, Ppmi, Multi-modal classification of neurodegenerative disease by progressive graph-based transductive learning, Med Image Anal 39 (2017) 218-230.
196 M. Wenzel, F. Milletari, J. Krüger, C. Lange, M. Schenk, I. Apostolova, S. Klutmann, M. Ehrenburg, R. Buchert, Automatic classification of dopamine transporter SPECT: deep convolutional neural networks can be trained to be robust with respect to variable image characteristics, Eur J Nucl Med Mol Imaging 46(13) (2019) 2800-2811.
197 M. Wodzinski, A. Skalski, D. Hemmerling, J.R. Orozco-Arroyave, E. Nöth, Deep Learning Approach to Parkinson’s Disease Detection Using Voice Recordings and Convolutional Neural Network Dedicated to Image Classification, 2019 41st Annual International Conference of the IEEE Engineering in Medicine and Biology Society (EMBC), 2019, pp. 717-720.
198 Y. Wu, P. Chen, Y. Yao, X. Ye, Y. Xiao, L. Liao, M. Wu, J. Chen, Dysphonic Voice Pattern Analysis of Patients in Parkinson's Disease Using Minimum Interclass Probability Risk Feature Selection and Bagging Ensemble Learning Methods, Comput Math Methods Med 2017 (2017) 4201984-4201984.
199 Y. Wu, J.-H. Jiang, L. Chen, J.-Y. Lu, J.-J. Ge, F.-T. Liu, J.-T. Yu, W. Lin, C.-T. Zuo, J. Wang, Use of radiomic features and support vector machine to distinguish Parkinson's disease cases from normal controls, Ann Transl Med 7(23) (2019) 773-773.
200 Y. Xia, Z. Yao, Q. Ye, N. Cheng, A Dual-Modal Attention-Enhanced Deep Learning Network for Quantification of Parkinson’s Disease Characteristics, IEEE Transactions on Neural Systems and Rehabilitation Engineering 28(1) (2020) 42-51.
201 G. Yadav, Y. Kumar, G. Sahoo, Predication of Parkinson's disease using data mining methods: a comparative analysis of tree, statistical, and support vector machine classifiers, Indian J Med Sci 65(6) (2011) 231-242.
202 E. Yagis, A.G.S.D. Herrera, L. Citi, Generalization Performance of Deep Learning Models in Neurodegenerative Disease Classification, 2019 IEEE International Conference on Bioinformatics and Biomedicine (BIBM), 2019, pp. 1692-1698.
203 O. Yaman, F. Ertam, T. Tuncer, Automated Parkinson's disease recognition based on statistical pooling method using acoustic features, Med Hypotheses 135 (2020) 109483-109483.
204 M. Yang, H. Zheng, H. Wang, S. McClean, Feature selection and construction for the discrimination of neurodegenerative diseases based on gait analysis, 2009 3rd International Conference on Pervasive Computing Technologies for Healthcare, 2009, pp. 1-7.
205 S. Yang, F. Zheng, X. Luo, S. Cai, Y. Wu, K. Liu, M. Wu, J. Chen, S. Krishnan, Effective dysphonia detection using feature dimension reduction and kernel density estimation for patients with Parkinson's disease, PLoS One 9(2) (2014) e88825-e88825.

206 Q. Ye, Y. Xia, Z. Yao, Classification of Gait Patterns in Patients with Neurodegenerative Disease Using Adaptive Neuro-Fuzzy Inference System, Comput Math Methods Med 2018 (2018) 9831252-9831252.
207 L.-L. Zeng, L. Xie, H. Shen, Z. Luo, P. Fang, Y. Hou, B. Tang, T. Wu, D. Hu, Differentiating Patients with Parkinson's Disease from Normal Controls Using Gray Matter in the Cerebellum, Cerebellum 16(1) (2017) 151-157.
208 X. Zhang, L. He, K. Chen, Y. Luo, J. Zhou, F. Wang, Multi-View Graph Convolutional Network and Its Applications on Neuroimage Analysis for Parkinson's Disease, AMIA Annu Symp Proc 2018 (2018) 1147-1156.
209 Y. Zhao, P. Wu, J. Wang, H. Li, N. Navab, I. Yakushev, W. Weber, M. Schwaiger, S. Huang, P. Cumming, A. Rominger, C. Zuo, K. Shi, A 3D Deep Residual Convolutional Neural Network for Differential Diagnosis of Parkinsonian Syndromes on 18F-FDG PET Images, 2019 41st Annual International Conference of the IEEE Engineering in Medicine and Biology Society (EMBC), 2019, pp. 3531-3534.